%% file: iclr2027_conference.tex
\documentclass{article}
\PassOptionsToPackage{table,dvipsnames}{xcolor}
\usepackage{iclr2027_conference,times}

\usepackage[utf8]{inputenc}
\usepackage[T1]{fontenc}
\usepackage{microtype}

\usepackage{amsmath, amssymb, amsfonts, amsthm, mathtools}
\usepackage{mathrsfs}
\usepackage{nicefrac}

\usepackage{graphicx}
\usepackage{subcaption}
\usepackage{wrapfig}
\usepackage{booktabs}
\usepackage{tabularx}
\usepackage{array}
\usepackage{siunitx}
\usepackage{capt-of}
\usepackage{longtable}

\usepackage{multirow}
\usepackage{booktabs}
\usepackage{graphicx}   % for \resizebox
\usepackage{pdflscape}  % for landscape table
\usepackage[table,dvipsnames]{xcolor}
\usepackage{colortbl}

\definecolor{burgundy}{HTML}{77324c}
\definecolor{tolblue}{HTML}{4477AA}
\definecolor{ours}{HTML}{228833}
\definecolor{baseline}{HTML}{BBBBBB}
\definecolor{oracle}{HTML}{4477AA}

\definecolor{tent}{HTML}{77324c}
\definecolor{eata}{HTML}{9B5774}
\definecolor{read}{HTML}{C07B8E}
\definecolor{sumi}{HTML}{5A2638}

\definecolor{sotta}{HTML}{CCBB44}
\definecolor{aeo}{HTML}{AA3377}
\definecolor{pta}{HTML}{882255}

\definecolor{textbody}{HTML}{333333}
\definecolor{textmuted}{HTML}{666666}

\definecolor{rowours}{HTML}{E3F0E5}
\definecolor{roworacle}{HTML}{DCE6F0}
\definecolor{rowdeg}{HTML}{F0E0E6}
\definecolor{rowgray}{HTML}{F5F5F5}

\usepackage{hyperref}
\hypersetup{colorlinks=true}
\usepackage{url}

\hypersetup{
    linkcolor=burgundy,
    citecolor=tolblue,
    urlcolor=burgundy,
    filecolor=burgundy,
    menucolor=burgundy,
    linktocpage=true,
    unicode=true,
    pdfborder={0 0 0},
}

\usepackage{natbib}

\usepackage[capitalize,noabbrev]{cleveref}

\usepackage{enumitem}
\usepackage{pifont}
\usepackage{gensymb}
\usepackage{adjustbox}
\usepackage[normalem]{ulem}
\useunder{\uline}{\ul}{}
\usepackage{makecell}

\usepackage{lipsum}
\usepackage{rotating}
\theoremstyle{plain}

\theoremstyle{definition}

\theoremstyle{remark}

\usepackage{algorithm}
\usepackage{algpseudocode}

\usepackage{amsmath}      % for \text, \!, equation environments, cases
\usepackage{amssymb}      % for \mathbb (e.g., \mathbb{1}, \mathbb{R})
\usepackage{mathtools}    % nicer-looking math; safe to include

\newcommand{\method}{\texttt{CARAT}}

\title{Source-Learned Reliance for Selective Test-Time Adaptation of Multimodal Time Series}
\author{
Payal Mohapatra$^{1}$\thanks{\small Equal contribution; shared first authorship.} \quad
Yueyuan Sui$^{2}$\footnotemark[1] \quad
Haodong Yang$^{2}$ \quad
Benjamin Lundell$^{1}$ \quad
Stephen Xia$^{2}$ \quad
Qi Zhu$^{2}$ \\
$^{1}$Arm Inc., USA\quad
$^{2}$Northwestern University, Evanston, IL, USA
}

\author{
\textbf{Payal Mohapatra}$^{1}$\thanks{\small Equal contribution; shared first authorship.}\hspace{0.7em}%
\textbf{Yueyuan Sui}$^{2}$\footnotemark[1]\hspace{0.7em}%
\textbf{Haodong Yang}$^{2}$\hspace{0.4em}%
\textbf{Benjamin Lundell}$^{1}$\hspace{0.4em}%
\textbf{Stephen Xia}$^{2}$\hspace{0.4em}%
\textbf{Qi Zhu}$^{2}$ \\[0.6em]
$^{1}$Arm Inc., USA \quad
$^{2}$Northwestern University, USA
}

\iclrfinalcopy % Uncomment only for the camera-ready version.
\begin{document}

\maketitle

\begin{abstract}

Multimodal wearable systems must remain reliable when sensor streams become noisy or unavailable. Existing multimodal test-time adaptation (TTA) methods often assess reliability online, but cross-modal agreement can be misleading when sensors measure different physical processes, and evaluating alternative modality configurations adds inference cost. We propose \method{}, which decouples model reliance from runtime corruption detection to guide omission or attenuation, amortizing reliance estimation through source training. An asymmetric modality-dropout curriculum prepares a missingness-resilient backbone for omission and derives a frozen, backbone-specific reliance proxy from windowed input-projection gradient norms. At deployment, a lightweight one-class detector flags suspect streams, and the proxy guides a joint choice between replacing the suspect set with the backbone's trained missingness symbol and attenuating its representations before fusion, without candidate-subset evaluation. Across four wearable datasets, five corruption types, three backbones, and eight TTA baselines, \method{} achieves the highest overall macro-F1 and best mean rank (2.42), exceeding EATA, the strongest baseline, by 1.58 F1 points across 12 equally weighted dataset--backbone settings. Across five profiled configurations, \method{} uses 9.49\% fewer GFLOPs and updates 47.82\% fewer parameters than EATA. A pattern also emerges across sensing regimes: multimodal TTA methods such as PTA are competitive on IMU-dominated homogeneous datasets, whereas unimodal TTA methods like TENT and EATA match or exceed it on heterogeneous datasets. These results position \method{} as a practical default to wearable TTA, offering competitive robustness with modest computational requirements and benefits that vary across backbones and dataset regimes.

\end{abstract}
\input{sections/intro}

\input{sections/relatedworks}
\input{sections/approach}
\input{sections/experiments}
\input{sections/conclusion}

\bibliographystyle{iclr2027_conference}
\bibliography{iclr2027_conference}

% Appendix numbering is initialized in sections/appendix.tex.
\input{sections/appendix}

\clearpage

% NeurIPS-only checklist is preserved in legacy_neurips/ and excluded from ICLR.

\end{document}

%% file: sections/intro.tex
\section{Introduction} \label{sec:intro}

Multimodal wearable systems combine several sensor streams to support applications such as human activity recognition and stress monitoring~\citep{altun2010comparative, reiss2019deep, schmidt2018introducing, mohapatra2024wearable}. These applications can involve numerous motion and physiological sensing streams, each with distinct physical characteristics and sources of noise. At deployment, missingness, sensor failure, motion artifacts, drift, and nonstationary noise can affect individual streams or several streams simultaneously~\citep{zhang2014troika, sharma2007prevalence}. Maintaining reliable predictions therefore requires handling changes in both the availability and quality of these inputs.

Test-time adaptation (TTA) addresses deployment shifts by updating pretrained models using unlabeled test data~\citep{wang2020tent, niu2022efficient}, with multimodal extensions targeting reliability imbalance and selective adaptation~\citep{yang2024test, guo2025smoothing, chen2025test}. 
These methods are evaluated on two or three modalities of one kind, such as audio, video, and optical flow~\citep{yang2024test, dong2025towards}, and decide which stream is unreliable either by comparing streams against each other, which fails when sensors measure different physical processes (an accelerometer and a skin-temperature sensor disagree by design), or by trying alternative modality configurations at inference, which costs extra forward passes. Wearable systems carry many streams of different kinds (ten in WESAD~\citep{schmidt2018introducing}), which makes both problems worse.
%Representative evaluations involve two or three modalities, including audio--visual recognition and video--audio--flow recognition~\citep{yang2024test, dong2025towards}. For wearable deployment, methods that compare alternative adaptation configurations add forward passes, while cross-modal agreement requires care because disagreement between sensors measuring different physical processes need not indicate corruption. 

% Our benchmark also reveals a structural trend across sensing configurations: multimodal baselines such as PTA are competitive on IMU-dominated datasets, whereas TENT and EATA perform better on the heterogeneous DaliaHAR and WESAD datasets (Section~\ref{sec:effectiveness}).

Missingness-resilient multimodal backbones offer a complementary opportunity~\citep{wang2023multi, NEURIPS2024_b2f2af54, mohapatra2025maestro, mohapatra24_interspeech, mohapatra2023person}. Because they are trained to predict from incomplete inputs, their missing-input operation can be reused to deliberately omit suspect streams during adaptation. \textbf{However, just supporting missing inputs does not establish when omission is beneficial.} Removing a corrupted stream may suppress harmful inputs, but it can also discard useful information that the remaining modalities cannot replace. Choosing between omission and attenuation (keeping a
stream but downweighting it before fusion) therefore requires considering the backbone's reliance on both the suspect and remaining streams, that is, how much its predictions depend on each.

The next challenge is to make this choice without the costly deployment-time probing. With $K$ suspect streams, unrestricted retain/remove choices admit $2^K$ subsets; assessing these alternatives requires repeated predictions, and even leave-one-modality-out tests add inference work. \textbf{Source training can provide an opportunity to estimate modality reliance before deployment.} A source-derived reliance proxy can summarize how the backbone learned to use its modalities, while a detector assesses current input abnormality. Combining these signals offers a way to guide omission and attenuation without evaluating candidate subsets online.

\noindent\textbf{Our Approach.} We propose \textbf{\method{}} (\textbf{C}onditional modality composition \textbf{a}nd \textbf{r}eliance-guided \textbf{at}tenuation for efficient test-time adaptation). The main idea is: during source training, the backbone learns to predict with missing streams and, as a by-product, how much it relies on each one. At test time, a detector flags suspect streams; if the backbone relies mostly on the healthy streams, the suspect ones are dropped, otherwise they are kept but turned down (Figure~\ref{fig:overall_fig}). An asymmetric modality-dropout curriculum uses windowed gradient norms at modality-specific input projections to prepare the backbone for missing inputs and derive a frozen, backbone-specific reliance proxy. %Gradient-flow analysis motivates the local optimization signal~\citep{lubana2020gradient}; its connection to predictive dependence is assessed empirically. 
%At deployment, a lightweight one-class detector flags suspect modalities using compact source-reference statistics. 
For each sample, the share of the reliance profile assigned to unflagged modalities determines whether to replace the entire suspect set with the trained missingness symbol or attenuate its encoder outputs through learnable scalars before fusion. Only normalization and attenuation parameters are updated using a sample-weighted entropy objective. The key contributions are:
\begin{itemize}[leftmargin=10pt, itemsep=0 pt, label=\raisebox{0.1ex}{\tiny$\blacktriangleright$}]
    \item \textbf{Reliance estimated at training time, corruption detected at test time.} \method{} estimates a backbone-specific reliance prior during source training and reuses it alongside runtime abnormality flags to guide omission or attenuation. This amortizes reliance estimation across deployment samples, avoiding explicit cross-modal agreement estimation and candidate-subset evaluation at test time (Section~\ref{sec:adapt}).

    \item \textbf{An asymmetric curriculum for missingness robustness and reliance estimation.} We modulate modality-dropout probabilities using windowed input-projection gradient norms, preparing the backbone for omission while deriving a frozen reliance proxy for adaptation. Section~\ref{sec:ablation} tests the curriculum's robustness benefit and the proxy's usefulness.
    %Source-training comparisons, prior perturbations, and white-noise and projection-width controls assess the curriculum's robustness benefits and the proxy's usefulness and sensitivity (Section~\ref{sec:ablation}).

    % \item \textbf{A multimodal time-series TTA benchmark with empirical insights.} Across three architecturally distinct backbones, four wearable datasets, five corruption types, and eight TTA baselines, \method{} achieves the best mean rank (2.42) across 12 dataset--backbone settings. The benchmark reveals a structural trend: multimodal baselines such as PTA are competitive on IMU-dominated datasets, whereas unimodal TTA baselines TENT and EATA match or exceed them on heterogeneous datasets. Efficiency profiling and design ablations complement these findings (Section~\ref{sec:effectiveness}).
    \item \textbf{A multimodal time-series TTA benchmark with robustness and efficiency insights.} 
    %Across three architecturally distinct backbones, four wearable datasets, five corruption types, and eight TTA baselines, 
    \method{} achieves the best mean rank (2.42) across 12 dataset--backbone settings. The benchmark reveals a pattern across sensing regimes: PTA is competitive on IMU-dominated homogeneous datasets, whereas TENT and EATA match or exceed it on heterogeneous datasets. Across five profiled configurations, \method{} uses 9.49\% fewer GFLOPs and updates 47.82\% fewer parameters than EATA, with substantial latency and CPU energy savings over AeO and PTA. Design ablations further examine the roles of source training, reliance-guided routing, and corruption detection (Sections~\ref{sec:effectiveness} and~\ref{sec:ablation}).
\end{itemize}

%% file: sections/relatedworks.tex
\section{Related Work}
\label{sec:related}

\smallskip
\noindent\textbf{Multimodal test-time adaptation.} TTA updates pretrained models using unlabeled deployment data~\citep{wang2020tent, niu2022efficient, gong2023sotta}. READ~\citep{yang2024test} combines self-adaptive attention with a confidence-aware objective; SuMi~\citep{guo2025smoothing} uses sample selection, unimodal assistance, and information sharing; PTA~\citep{wangpartition} combines prediction-bias reweighting with attention alignment; and AEO~\citep{dong2025towards} optimizes entropy and modality-prediction discrepancies for open-set adaptation. Closest to our intervention design, TSA~\citep{chen2025test} learns modality-specific adapters and a router at test time for single-modality shifts, evaluating alternative adapter configurations at inference. Recently proposed, DASP~\citep{he2026decoupling} uses test-batch feature redundancy to identify shifted modalities and selectively update stable or plastic adapters. \method{} combines a frozen source-derived reliance proxy with current abnormality flags to choose jointly between removing the suspect set and attenuating its representations before fusion, without candidate-subset predictions during routing.

\smallskip
\noindent\textbf{Source training for missingness and modality imbalance.} Missingness-resilient backbones such as ShaSpec~\citep{wang2023multi}, Flex-MoE~\citep{NEURIPS2024_b2f2af54}, and MAESTRO~\citep{mohapatra2025maestro} support prediction with incomplete inputs, providing trained mechanisms for modality omission. Meanwhile, joint training can favor dominant modalities and under-optimize others~\citep{wang2020makes, peng2022balanced}. OGM~\citep{peng2022balanced} modulates gradients according to relative modality contributions, while OPM~\citep{wei2024fly} dynamically drops modality features based on discriminative-performance discrepancies. MIAM~\citep{zbinden2026miam} adjusts token-masking distributions using modality-specific performance and learning dynamics, supporting incomplete inputs within and across modalities. OPM and MIAM thus already connect adaptive masking with missing-input robustness. \method{} builds on these capabilities to address a further decision: whether a currently abnormal stream can be discarded. Its windowed gradient norms guide missingness exposure and produce a frozen, backbone-specific reliance proxy, which combines with deployment-time abnormality flags to select joint removal or pre-fusion attenuation of the suspect set.

\smallskip
\noindent\textbf{Time-series shift detection and adaptation.} Time-series methods detect anomalies through reconstruction-based scoring~\citep{xu2023fits, sunmultivariate} or association discrepancy~\citep{xu2021anomaly}, and adapt through spectral alignment and temporal normalization~\citep{gnassounou2024multi, gnassounou2025psdnorm}. Forecasting methods address nonstationarity through test-time calibration or shift-aware updates~\citep{kim2025battling, grover2025shift}. \method{} combines reconstruction error and power spectral density distance relative to clean source statistics to flag suspect streams, complementing its backbone-dependence proxy with observation-level abnormality information. Detector outputs enter routing as per-modality flags, permitting alternative detectors; Section~\ref{sec:ablation} examines sensitivity to imperfect flags.

% \smallskip
% \noindent\textbf{Learning under noisy multimodal time-series sensing.}
% Time-series sensing is routinely affected by motion artifacts, sensor drift, missing samples, and structured signal degradation~\citep{zhang2014troika, temko2017accurate, sabatini2011estimating, cao2018brits, malhotra2016lstm}. Classical approaches address these issues through domain-specific correction pipelines, while data-driven methods rely on augmentation to enlarge the effective training distribution~\citep{ hendrycks2019augmix}. Both paradigms assume that corruption families are sufficiently characterized during training or deployment. In contrast, \method{} targets multimodal test-time corruptions that are unknown a priori and may affect arbitrary subsets of modalities simultaneously.

% \smallskip
% \noindent\textbf{Time-series OOD detection and filtering.}
% Recent time-series adaptation and OOD methods have explored spectral normalization, reconstruction-based scoring, and sample filtering for stabilizing adaptation under distribution shift. \method{} does not introduce a specialized OOD detector; instead, it uses lightweight one-class statistics only as a coarse routing signal, similar in spirit to the lightweight filtering heuristics employed in methods such as SoTTA~\citep{gong2023sotta} and SuMi~\citep{guo2025smoothing}. 

%% file: sections/approach.tex
\section{Methodology}\label{sec:method}
\begin{figure}[t]
    \centering
    \includegraphics[width=\linewidth]{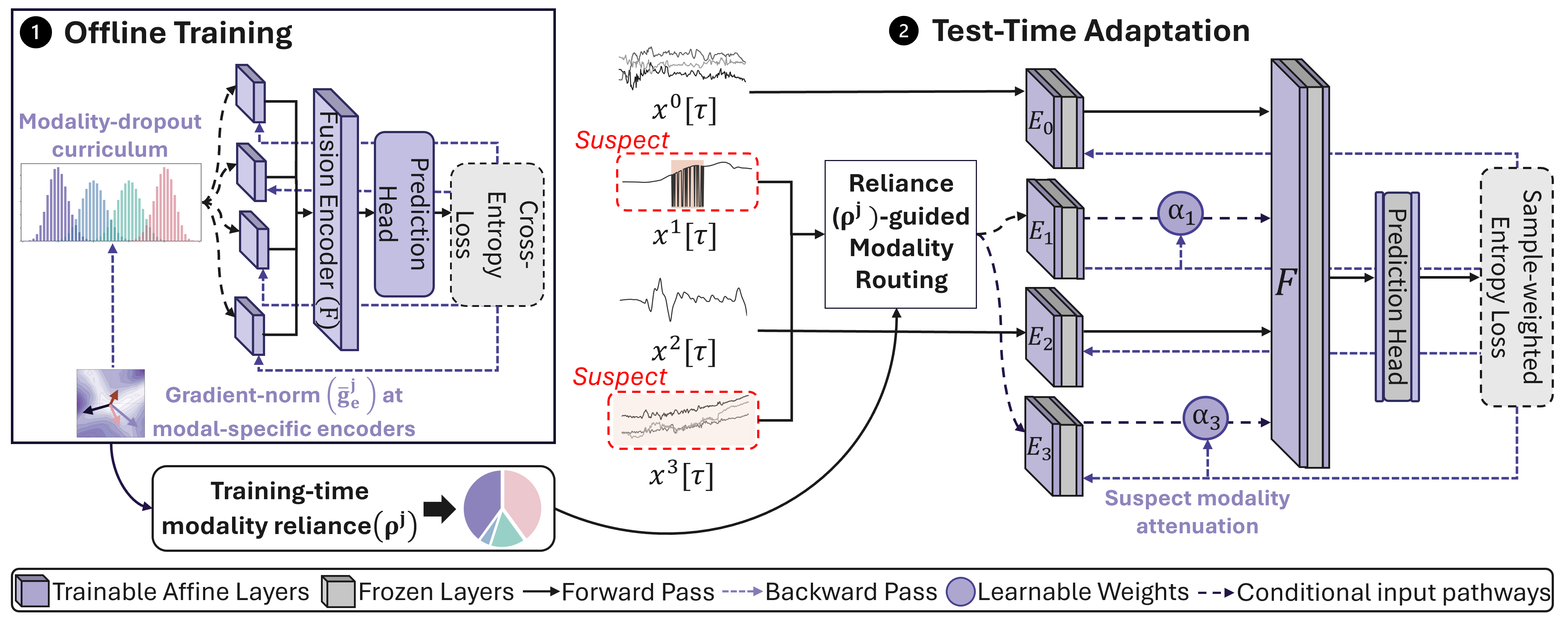}
    \caption{\small \textbf{Overview of \method{}.} During offline training, an asymmetric modality-dropout curriculum adjusts per-modality dropout rates using windowed gradient norms $\bar{g}_e^j$, yielding a training-time reliance profile $\rho^j$ (Section~\ref{sec:reliance}). At test time, a lightweight detector identifies suspect modalities (red; here, $x^1$ and $x^3$), and the frozen reliance profile selects one intervention for the entire suspect set (Section~\ref{sec:adapt}): \emph{modality composition} replaces that set with the missingness symbol $s_0$ when the unflagged modalities retain sufficient reliance mass; otherwise, \emph{latent attenuation} downweights the suspect representations through learnable scalars $a^j$ before fusion. Adaptation updates only normalization and attenuation parameters using a sample-weighted entropy objective.}
    \vspace{-10pt}
    \label{fig:overall_fig}
\end{figure}

\subsection{Problem Formulation}
\label{sec:prob}

We denote each multimodal time-series sample as $\mathcal{M}_i = \left( \{x_i^1[\tau]\}_{\tau=1}^{T^1}, \ldots, \{x_i^M[\tau]\}_{\tau=1}^{T^M} \right)$, where $x_i^j[\tau] \in \mathbb{R}^{D^j}$ is the feature vector at temporal index $\tau$ for modality $j$, and each modality forms a multivariate time series $x_i^j \in \mathbb{R}^{D^j \times T^j}$. Let $f_\theta$ denote a multimodal predictor trained on labeled source data $\mathcal{D}_s=\{(\mathcal{M}_i,y_i)\}_{i=1}^N \sim p_s(\mathcal{M},y)$. At test time, the model receives a sequence of unlabeled batches $\{\mathcal{B}_t\}_{t\ge1}$ in an episodic online adaptation setting. We study corruption-induced input distribution shifts generated by unknown processes $\mathcal{T}_t^j$ acting independently or jointly on sensor observations. These processes introduce additive noise, structured sensor degradation, or missingness, changing the distribution of observed inputs relative to source data while leaving each sample's underlying task label unchanged. Multiple modalities may be corrupted simultaneously; the affected subset and corruption severity are not supplied to the adaptation method. We consider architectures of the form
\[
f_\theta(\mathcal{M}_i)
=
h\!\left(
F(E_1(x_i^1),\ldots,E_M(x_i^M))
\right),
\]
where $E_j$ denotes a modality-specific encoder, $F$ a multimodal fusion module, and $h$ the prediction head. The encoders and fusion module may incorporate transformer layers, gating, routing, or mixture-of-experts mechanisms. Our objective is to control modality participation and update a small subset of model parameters using only unlabeled test batches $\{\mathcal{B}_t\}$, without access to raw source samples during adaptation, to improve robustness under multimodal corruption.

\subsection{Our Approach}
\label{sec:approach}

We propose \method{} (\textbf{C}onditional modality composition \textbf{a}nd \textbf{r}eliance-guided \textbf{at}tenuation for efficient test-time adaptation), a multimodal test-time adaptation framework that combines current input abnormality with source-learned modality reliance. A missingness-resilient backbone supplies a trained operation for omitting inputs; the reliance profile helps decide when to use it. \method{} routes each sample between removal and attenuation without evaluating candidate modality subsets. During training, we replace uniform modality dropout with an asymmetric curriculum that modulates each modality's dropout rate using its windowed input-projection gradient norm, yielding a reliance profile $\{\rho^j\}_{j=1}^{M}$ at convergence (Section~\ref{sec:reliance}). At test time, a lightweight per-modality one-class detector identifies suspect inputs (Section~\ref{sec:adapt}), and the learnt reliance prior determines the intervention. If the unflagged modalities retain sufficient reliance mass, all suspect inputs to $E_j$ are replaced with the missingness symbol; otherwise, the suspect modalities are propagated but attenuated prior to fusion in $F$. Adaptation proceeds by updating only normalization and attenuation parameters using a sample-weighted entropy objective over unlabeled test batches. An overview is shown in Figure~\ref{fig:overall_fig}.

\subsubsection{Modality Reliance Estimation during Training}
\label{sec:reliance}

For the evaluated missingness-resilient backbones~\citep{wang2023multi, NEURIPS2024_b2f2af54, mohapatra2025maestro}, we start from a shared modality-dropout curriculum, exposing the model to easy, complete samples first and progressing to harder, incomplete samples. After a warm-up phase, existing schedules apply a scalar dropout rate $p_{\text{base}}(e)$ uniformly across all modalities at epoch $e$ (full schedule in Appendix~\ref{app:curriculum}). We introduce gradient-guided modulation, drawing on prior observations of imbalanced multimodal learning~\citep{peng2022balanced, wang2020makes}: modalities with larger windowed gradient norms receive higher dropout probabilities, exposing the backbone more often to their absence. As a by-product, the resulting per-modality dropout rate at convergence serves as a proxy for the model's learned reliance on each modality, which we use directly at test time in Section~\ref{sec:adapt}. We follow the symbolic aggregate approximation of~\citet{mohapatra2025maestro} for tokenization, with a reserved symbol $s_0$ indicating missing modalities.

\begin{wrapfigure}{r}{0.33\textwidth}
    \centering
    \captionsetup{font=scriptsize}
    \vspace{-9pt}
    \includegraphics[width=\linewidth]{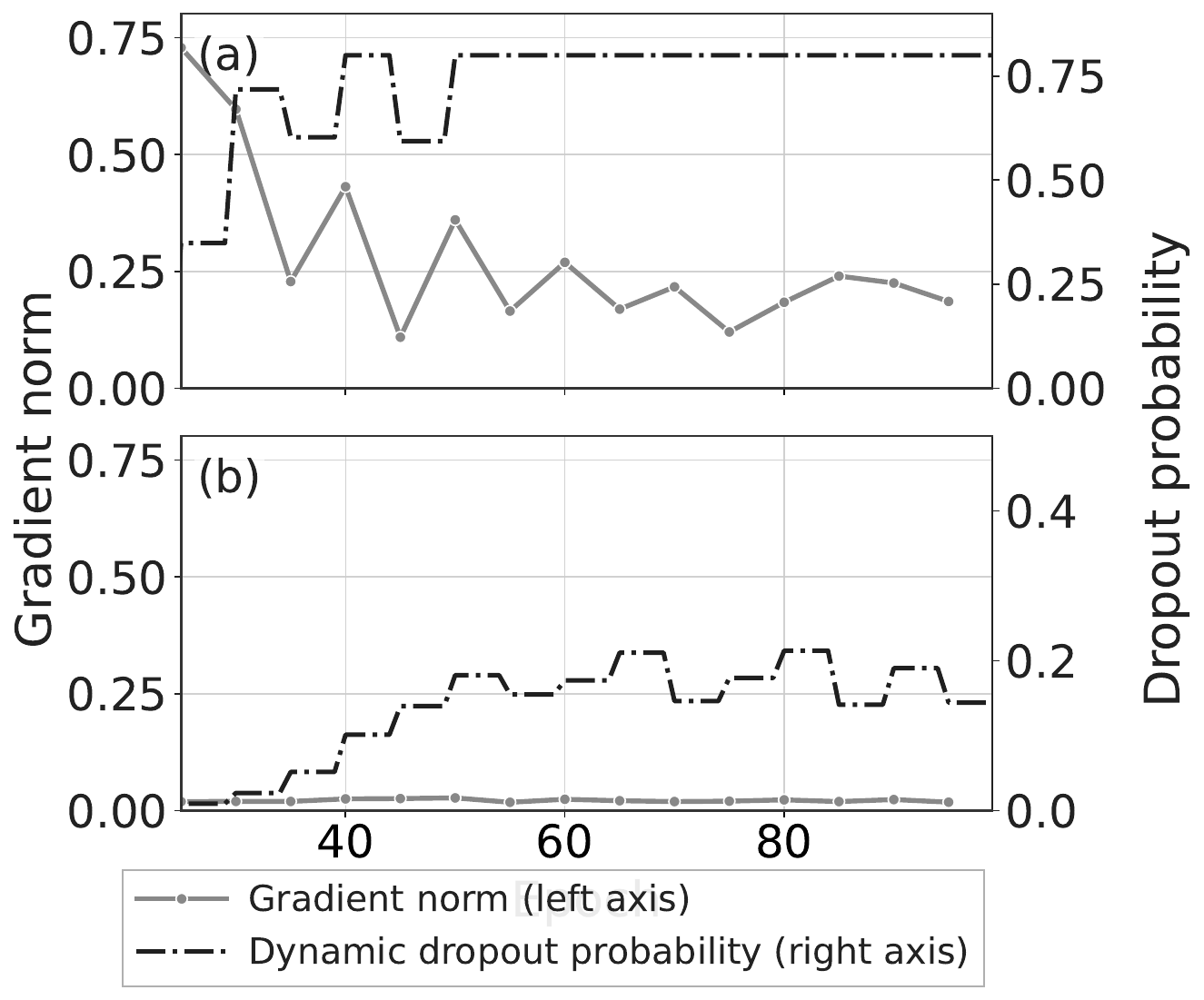}
    \vspace{-20pt}
    \caption{\scriptsize Per-modality gradient norm $\bar{g}_e^j$ and corresponding dropout probability $p_e^j$ on activity recognition (DaliaHAR) dataset for (a) wrist accelerometer and (b) wrist temperature.}
    \vspace{-25pt}
    \label{fig:curr}
\end{wrapfigure} 
Concretely, we use $p_{\text{base}}(e)$ as the target average dropout rate before clipping and let training dynamics redistribute it across modalities. Let $\theta_j \subset \theta$ denote the parameters of modality $j$'s input projection. We measure the gradient norm $g_k^j = \|\nabla_{\theta_j}\mathcal{L}_k(\theta)\|_2$ of the supervised loss $\mathcal{L}_k$ at step $k$ as a local optimization signal for this modality-specific parameter block. At the start of each epoch, we form a per-modality reliance ratio by normalizing windowed gradient norms across modalities.
\begin{equation}
r_e^j \;=\; \frac{\bar{g}_e^j}{\sum_{\ell=1}^{M} \bar{g}_e^\ell},
\qquad
\bar{g}_e^j \;=\; \frac{1}{W}\!\!\sum_{u = k_e - W}^{k_e - 1}\! g_u^j,
\label{eq:reliance_ratio_epoch}
\end{equation}
with $r_e^j \in [0,1]$ and $\sum_j r_e^j = 1$. The Bernoulli dropout probability for modality $j$ during epoch $e$ is then, $p_e^j \;=\; \min\!\big(p_{\text{cap}},\; M \cdot p_{\text{base}}(e) \cdot r_e^j\big),$ held fixed throughout the epoch. The factor $M$ preserves $p_{\text{base}}(e)$ as the mean Bernoulli rate before clipping. We use the normalized gradient magnitudes as an optimization-based proxy for modality reliance. Figure~\ref{fig:curr} illustrates this on DaliaHAR (modalities: accelerometer, temperature, blood volume pulse, electrodermal activity): the gradient norm $\bar{g}_e^j$ is consistently larger for the accelerometer than for temperature, and the schedule translates this gap directly into a higher dropout probability for the accelerometer during training. We log the dropout probability at the trained model $\theta^\star$ as the model's \emph{reliance profile} $\rho^j = p_{e^\star}^j$, which is a clipped, scaled version of the windowed gradient share recorded for that checkpoint. Source preparation therefore supplies both exposure to missing inputs and a frozen, backbone-specific prior for adaptation (Section~\ref{sec:adapt}); their empirical contributions are examined in Section~\ref{sec:ablation} and Appendix~\ref{app:rec_vs_ours_results}.

\noindent\textbf{Optimization-based motivation.} Gradient flow provides a local motivation for this signal~\citep{lubana2020gradient}. For a differentiable loss under $\dot{\theta}=-\nabla_\theta\mathcal{L}$, the contribution of parameter block $\theta_j$ to $\dot{\mathcal{L}}$ is $-\|\nabla_{\theta_j}\mathcal{L}\|_2^2$. This identity motivates measuring gradient magnitude at modality-specific projections. Our windowed, clipped profile uses that signal as a reliance proxy; its connection to predictive dependence is assessed empirically rather than implied by the identity.

\begin{wrapfigure}{r}{0.68\textwidth}
    \centering
    \captionsetup{font=small}
    \vspace{-10pt}

    \begin{minipage}{0.49\linewidth}
        \centering
        \includegraphics[width=\linewidth]{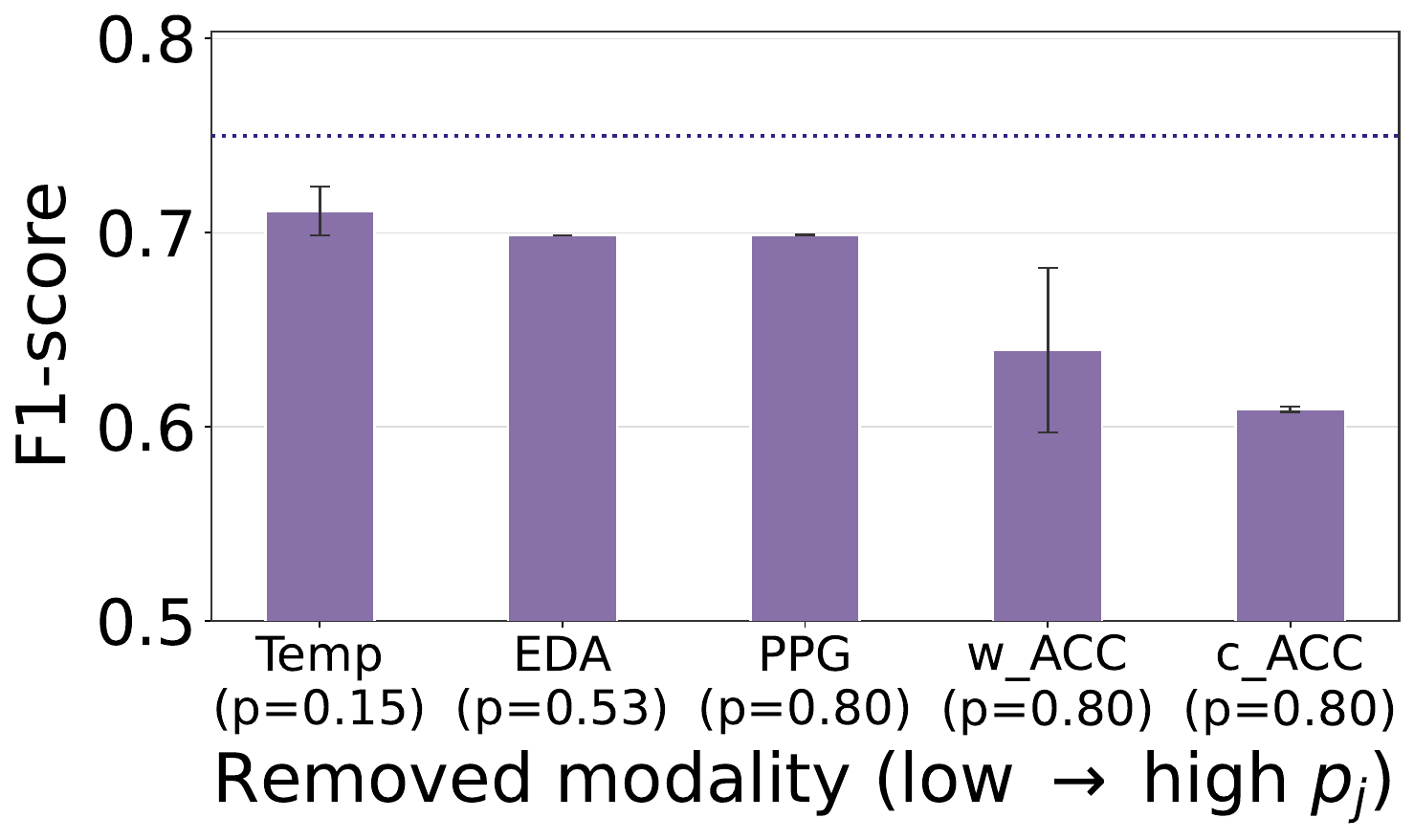}
    \end{minipage}
    \hfill
    \begin{minipage}{0.49\linewidth}
        \centering
        \includegraphics[width=\linewidth]{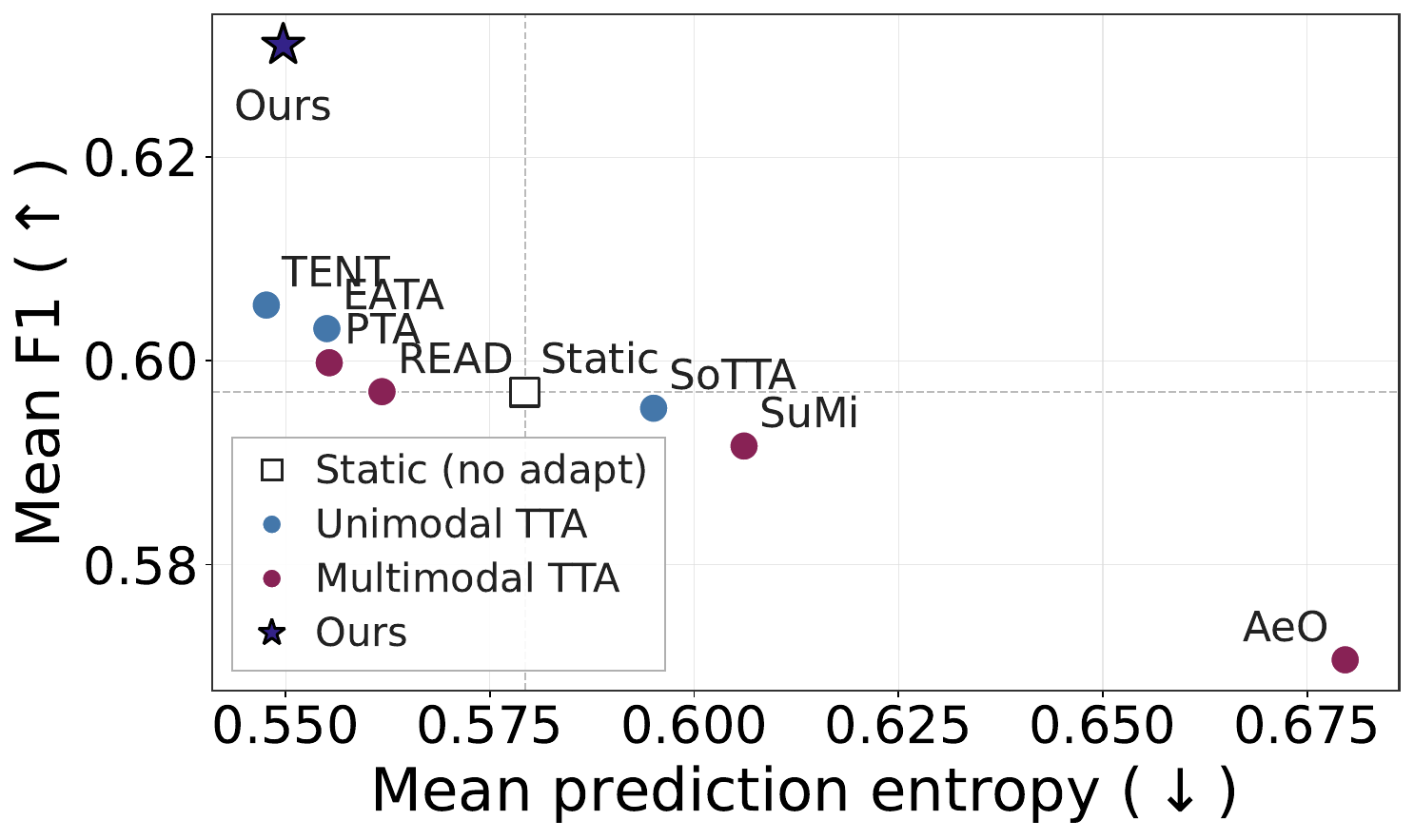}
    \end{minipage}
    \vspace{-8pt}
    \caption{\small \textbf{Left:} Leave-one-modality-out F1 on DaliaHAR--ShaSpec, ordered by terminal dropout probability $\rho^j$. \textbf{Right:} Mean prediction entropy and F1 under Diverse corruption (10 scenarios, 5 corruption types).}
    \vspace{-15pt}
    \label{fig:theory_motivation}
\end{wrapfigure}Figure~\ref{fig:theory_motivation} (left) connects the proxy to the intervention problem: removing the high-reliance accelerometers causes a larger F1 decline than removing low-reliance temperature on DaliaHAR--ShaSpec. This motivates retaining some information from suspect streams when the backbone relies strongly on them. Input removal reuses the missingness symbol seen during training; attenuation instead preserves a downweighted representation when removal may be costly. Figure~\ref{fig:theory_motivation} (right) shows that lower entropy accompanies higher F1 for \method{} in the illustrated evaluation. The prior perturbations in Section~\ref{sec:ablation} directly test the utility of the learned profile, while Appendix~\ref{app:reliance_sketch} details the gradient-flow interpretation and its limits.

\subsubsection{Reliance-Guided Test-Time Adaptation}
\label{sec:adapt}

For each test sample $\mathcal{M}_i$, a per-modality detector $\mathcal{D}^j(\cdot)$ flags suspect modalities. The detector assesses current input abnormality; the source-derived profile guides the intervention. The frozen reliance profile $\{\rho^j\}$ then determines the intervention: if the surviving (unflagged) modalities account for a sufficient share of the reliance budget, the suspect inputs are replaced with the missingness symbol $s_0$ at the input to $E_j$ (dropped); otherwise, they propagate but their pre-fusion contribution is dampened by a learnable attenuator before $F$. 

\noindent \textbf{Suspect modality detection.}
For each sample $\mathcal{M}_i$ and modality $j$, a lightweight one-class detector $\mathcal{D}^j(\cdot)$ computes a corruption score using reconstruction and spectral-distance features relative to compact clean-reference statistics computed during source training~\citep{xu2023fits, gnassounou2024multi, gnassounou2025psdnorm}. No raw source data is accessed during adaptation. Modalities whose score exceeds a fixed threshold are flagged as suspect, producing the routing set $\mathcal{S}_i$. Since \method{} consumes only this binary routing signal, the framework remains agnostic to the specific detector choice. The detectors are trained offline and reused at deployment. Appendix~\ref{app:detector_cost} details their formulation and sensitivity analyses, while Appendix~\ref{app:detector-footprint} reports their training cost, detector-only inference latency, memory footprint, and stored artifacts.

\noindent \textbf{Reliance-guided routing.}
Given $\mathcal{S}_i$, the reliance profile $\{\rho^j\}_{j=1}^{M}$ guides the adaptation action on the suspect modalities. The share of the training-time reliance budget assigned to the unflagged modalities of sample $i$ is
\vspace{-10pt}
\begin{equation}
\beta_i \;=\; \frac{\sum_{j \notin \mathcal{S}_i} \rho^j}{\sum_{\ell=1}^{M} \rho^\ell} \;\in\; [0, 1].
\label{eq:reliability}
\end{equation}
Sample $i$ is routed to modality-composition adaptation when $\beta_i \ge \nu$, and to latent-attenuation adaptation otherwise. This selects one action for the entire suspect set using the relative reliance mass, without evaluating alternative subsets. Both ordering and relative magnitudes can affect $\beta_i$; Table~\ref{tab:ablation-prior} tests the sensitivity to reversing the ordering and to order-preserving perturbations. 

\noindent \textbf{Modality-composition adaptation.}
When $\beta_i \ge \nu$, the unflagged modalities meet the retained-reliance threshold, and we use the backbone's trained missing-input operation. We replace suspect modalities $j \in \mathcal{S}_i$ with the missingness symbol $s_0$ at the encoder input. The substitution is applied to each flagged modality and introduces no additional parameters; it reuses the missingness symbol employed by the modality-dropout curriculum (Section~\ref{sec:reliance}).

\noindent \textbf{Suspect modality attenuation.} When $\beta_i < \nu$, the prior assigns a larger share of reliance to the suspect set, motivating attenuation to retain potentially useful signal while limiting its contribution to fusion. We attenuate each $j \in \mathcal{S}_i$ using a learnable scalar $a^j = \sigma(\alpha^j) \in (0,1)$ applied to its encoder output:
\begin{equation}
E_j^{\,\prime}(\tilde{x}_i^j) =
\begin{cases}
a^j \cdot E_j(\tilde{x}_i^j), & j \in \mathcal{S}_i, \\
E_j(\tilde{x}_i^j), & \text{otherwise}.
\end{cases}
\label{eq:latent_mod}
\end{equation}
The scalar $a^j$ acts as a per-modality gain on the latent stream entering fusion. Since each $\alpha^j$ must adapt within only a few episodic steps, we use a substantially larger learning rate than for LayerNorm affine parameters ($\eta_\alpha \gg \eta_{\text{LN}}$). Appendix Figure~\ref{fig:sensitivity_lr} shows that small learning rates keep the gates near the sigmoid midpoint, while larger values enable meaningful adaptation within the same update budget.

\noindent \textbf{Overall test-time optimization.}
A test batch $\mathcal{B}_t$ may contain both samples with flagged modalities ($\mathcal{S}_i \neq \emptyset$) and samples without flagged modalities ($\mathcal{S}_i = \emptyset$). Following~\citet{wang2020tent}, we minimize a weighted entropy objective:
\begin{equation}
\mathcal{L}_{\text{TTA}}(\theta;\mathcal{B}_t)
=
\frac{\sum_{i=1}^{B} w_i\,\mathcal{H}\!\left(p_\theta(y\mid\mathcal{M}_i)\right)}
{\sum_{i=1}^{B} w_i},
\qquad
w_i=1+\lambda\,\mathbb{1}[\mathcal{S}_i=\emptyset].
\label{eq:tta_objective}
\end{equation}
Here, $\mathcal{H}(\cdot)$ denotes prediction entropy and $\lambda\ge0$ upweights samples without flagged modalities, to emphasize inputs judged less affected by the detector. We update only normalization parameters and attenuation logits $\alpha^j$, keeping the source reliance profile fixed and resetting the adaptive parameters between test batches.

%% file: sections/experiments.tex
\section{Results and Discussion}\label{sec:results}

We evaluate whether source-learned reliance helps a missingness-resilient backbone decide when to remove suspect inputs and when to retain them through attenuation. The experiments test overall robustness and deployment cost (Section~\ref{sec:effectiveness}), then examine the contributions of source preparation, reliance ordering, detection, and online updates (Section~\ref{sec:ablation}).

\noindent\textbf{Benchmark.} We use four wearable sensing datasets: DSADS~\citep{altun2010comparative} (19-class activity recognition from five body-worn IMUs), PAMAP2~\citep{reiss2012pamap2, defrana_physical_activity} (12-class activity recognition from three IMUs and heart rate), DaliaHAR~\citep{reiss2019deep} (7-class activity recognition from wrist motion and physiological signals plus chest acceleration), and WESAD~\citep{schmidt2018introducing} (3-class stress classification from ten chest and wrist streams). The benchmark spans IMU-dominated and heterogeneous sensing configurations. We apply five synthetic sensor corruptions to selected modality subsets: discrete missingness, waveform anomalies, nonlinear degradation, frequency-selective attenuation, and nonstationary noise. The complete grids contain 50, 50, 45, and 95 conditions per backbone on DSADS, DaliaHAR, PAMAP2, and WESAD, respectively. Appendix~\ref{app:dataset} details the sensors and splits; Appendix~\ref{app:noise} specifies corruption models and parameters, with examples in Figure~\ref{fig:noise_illustration}.

% \footnote{A minimal implementation is included in the supplementary materials; we will release the corruption scenarios and baseline implementations.}

\noindent\textbf{Backbones and comparators.} We evaluate MAESTRO~\citep{mohapatra2025maestro}, ShaSpec~\citep{wang2023multi}, and Flex-MoE~\citep{NEURIPS2024_b2f2af54}, which differ in their encoders and fusion mechanisms but support missing inputs and modality-specific attenuation before fusion. Comparators comprise Static (no adaptation) and eight TTA methods: TENT~\citep{wang2020tent}, EATA~\citep{niu2022efficient}, and SoTTA~\citep{gong2023sotta}; multimodal methods READ~\citep{yang2024test}, SuMi~\citep{guo2025smoothing}, AeO~\citep{dong2025towards}, and PTA~\citep{wangpartition}; and time-series-specific PSDNorm~\citep{gnassounou2025psdnorm}. In the main benchmark, all methods share the same source checkpoint trained with our asymmetric curriculum and the same corrupted test batches. Thus, Table~\ref{tab:overall-robustness} compares adaptation on a common source model. Source-training effects are assessed separately in Appendix Table~\ref{tab:rec_ours}, and replacement curricula are examined in Section~\ref{sec:ablation}. Baseline implementations and update schedules are in Appendix~\ref{app:tta}.

\noindent\textbf{Evaluation and reporting.} We use episodic TTA~\citep{wang2020tent}: each method adapts on an unlabeled test batch, predicts that batch, and resets before the next. Table~\ref{tab:overall-robustness} averages macro-F1 within each dataset--backbone setting and weights the 12 settings equally; its descriptive mean rank is computed from the setting means before rounding. In the DaliaHAR ablations, Diverse (Div.) averages ten modality-subset scenarios and five corruption types. One F1 point denotes 0.01 on the 0--1 scale; numerical differences are computed before display rounding when finer source values are available. Appendix~\ref{app:effectiveness_stats} retains the paired Wilcoxon summaries and five corruption-redraw trials from the repeated evaluation. We profile updated parameters, GFLOPs, latency, and CPU energy for batches of 64, including adaptation, inference, and detector overhead (Appendix~\ref{app:efficiency}).

\subsection{Effectiveness under Multimodal Test-Time Corruption}
\label{sec:effectiveness}

\begin{table*}[t]
\centering
\caption{\textbf{Robustness across datasets and backbones.} Mean macro-F1 ($\uparrow$, 0--1) over 50/50/45/95 modality-subset $\times$ corruption conditions per backbone on DSADS/DaliaHAR/PAMAP2/WESAD. Overall F1 equally weights the 12 settings; Rank ($\downarrow$) averages ranks of their means before rounding, with average ranks for exact ties. \textcolor{burgundy}{\textbf{Bold}} / \underline{underline}: best / second-best displayed values; rounded ties are co-marked. MAE: MAESTRO; Sha: ShaSpec; Flex: Flex-MoE.}
\label{tab:overall-robustness}
\vspace{-8pt}
\setlength{\tabcolsep}{2pt}
\renewcommand{\arraystretch}{1.03}
\setlength{\aboverulesep}{1.3pt}
\setlength{\belowrulesep}{1.5pt}
\fontsize{8}{9}\selectfont
\begin{tabular*}{\textwidth}{@{\extracolsep{\fill}}l*{3}{c}|*{3}{c}|*{3}{c}|*{3}{c}|*{2}{>{\columncolor{tolblue!10}}c}@{}}
\toprule
& \multicolumn{3}{c|}{\textbf{DSADS}} & \multicolumn{3}{c|}{\textbf{DaliaHAR}} & \multicolumn{3}{c|}{\textbf{PAMAP2}} & \multicolumn{3}{c|}{\textbf{WESAD}} & \multicolumn{2}{c}{\cellcolor{tolblue!10}\textbf{Overall}} \\
\cmidrule(lr){2-4} \cmidrule(lr){5-7} \cmidrule(lr){8-10} \cmidrule(lr){11-13}
\textbf{Method} & MAE & Sha & Flex & MAE & Sha & Flex & MAE & Sha & Flex & MAE & Sha & Flex & \textbf{F1} $\uparrow$ & \textbf{Rank} $\downarrow$ \\
\midrule
Static & 0.69 & \underline{0.44} & 0.53 & 0.74 & 0.60 & 0.64 & \textcolor{burgundy}{\textbf{0.92}} & 0.70 & 0.75 & \textcolor{burgundy}{\textbf{0.65}} & \underline{0.53} & \underline{0.53} & 0.64 & 5.42 \\
TENT & 0.69 & 0.42 & \underline{0.54} & \textcolor{burgundy}{\textbf{0.76}} & \underline{0.61} & \textcolor{burgundy}{\textbf{0.68}} & \textcolor{burgundy}{\textbf{0.92}} & 0.69 & \textcolor{burgundy}{\textbf{0.77}} & \textcolor{burgundy}{\textbf{0.65}} & \underline{0.53} & \textcolor{burgundy}{\textbf{0.54}} & \underline{0.65} & 3.92 \\
EATA & 0.69 & 0.42 & \underline{0.54} & \textcolor{burgundy}{\textbf{0.76}} & 0.60 & \textcolor{burgundy}{\textbf{0.68}} & \textcolor{burgundy}{\textbf{0.92}} & 0.69 & \textcolor{burgundy}{\textbf{0.77}} & \textcolor{burgundy}{\textbf{0.65}} & \underline{0.53} & \textcolor{burgundy}{\textbf{0.54}} & \underline{0.65} & \underline{3.50} \\
READ & 0.69 & 0.39 & \underline{0.54} & 0.74 & 0.60 & 0.65 & \underline{0.91} & 0.66 & 0.75 & \underline{0.64} & \underline{0.53} & \underline{0.53} & 0.64 & 6.83 \\
SuMi & 0.68 & 0.43 & \underline{0.54} & 0.74 & 0.59 & 0.64 & \underline{0.91} & 0.70 & \underline{0.76} & \underline{0.64} & \underline{0.53} & 0.51 & 0.64 & 6.58 \\
SoTTA & 0.69 & 0.43 & \underline{0.54} & \underline{0.75} & 0.60 & 0.66 & \textcolor{burgundy}{\textbf{0.92}} & 0.69 & \underline{0.76} & \textcolor{burgundy}{\textbf{0.65}} & \underline{0.53} & 0.52 & 0.64 & 4.92 \\
AeO & 0.69 & \underline{0.44} & 0.53 & 0.72 & 0.57 & 0.61 & \underline{0.91} & 0.69 & \underline{0.76} & \underline{0.64} & 0.52 & 0.49 & 0.63 & 7.83 \\
PTA & \underline{0.70} & 0.43 & \underline{0.54} & 0.74 & 0.60 & 0.60 & \textcolor{burgundy}{\textbf{0.92}} & 0.70 & \textcolor{burgundy}{\textbf{0.77}} & 0.63 & 0.52 & 0.50 & 0.64 & 5.42 \\
PSDNorm & 0.60 & \underline{0.44} & 0.46 & 0.68 & 0.57 & 0.61 & 0.85 & \underline{0.71} & 0.72 & \underline{0.64} & 0.48 & 0.51 & 0.61 & 8.17 \\
\midrule
\textbf{\method{} (Ours)} & \textcolor{burgundy}{\textbf{0.72}} & \textcolor{burgundy}{\textbf{0.52}} & \textcolor{burgundy}{\textbf{0.55}} & \textcolor{burgundy}{\textbf{0.76}} & \textcolor{burgundy}{\textbf{0.63}} & \underline{0.67} & \textcolor{burgundy}{\textbf{0.92}} & \textcolor{burgundy}{\textbf{0.72}} & \textcolor{burgundy}{\textbf{0.77}} & 0.63 & \textcolor{burgundy}{\textbf{0.55}} & \textcolor{burgundy}{\textbf{0.54}} & \textcolor{burgundy}{\textbf{0.66}} & \textcolor{burgundy}{\textbf{2.42}} \\
\bottomrule
\end{tabular*}
\vspace{-8pt}
\end{table*}

Table~\ref{tab:overall-robustness} yields three key observations. \textbf{(i) \method{} achieves the highest overall macro-F1 and best mean rank.} It improves on EATA, the strongest baseline overall, by 1.58 F1 points on average, with a median paired improvement of 0.75 points across the 12 settings. Its mean rank is 2.42, compared with 3.50 for EATA. \textbf{(ii) The largest gains occur on ShaSpec across all four datasets.} \method{} leads in every ShaSpec setting, averaging 4.39 points above EATA, compared with 0.15 on MAESTRO and 0.19 on Flex-MoE. \textbf{(iii) \method{} leads or remains close to the best method in 11 of the 12 settings.} It ranks first in seven settings and third in four, with the latter trailing the best method by only 0.18--0.56 F1 points. Appendices~\ref{app:dsads_complete}--\ref{app:wesad_complete} provide every modality-subset and corruption result. Appendix~\ref{app:effectiveness_stats} reports paired effects, intervals, and condition-level ranks; its separate five-redraw DSADS--MAESTRO experiment reports positive paired intervals against all eight included comparators, assessing corruption variability for a fixed checkpoint.

\textbf{The benchmark reveals a structural trend: the relative strength of TTA methods changes with the sensing configuration.} On IMU-dominated DSADS and PAMAP2, multimodal methods such as PTA are competitive; PTA is the strongest baseline on DSADS--MAESTRO and DSADS--Flex-MoE. On heterogeneous DaliaHAR and WESAD, however, TENT and EATA outperform PTA across all six dataset--backbone settings. With Flex-MoE, both obtain approximately 0.68 on DaliaHAR and 0.54 on WESAD, compared with PTA's 0.60 and 0.50. A plausible explanation is that disagreement between sensors measuring different physical processes can reflect complementary information, making it an ambiguous cue for corruption. This observation motivates \method{}'s separation of sensor abnormality from backbone reliance: current detector flags identify suspect inputs, while the source-learned prior guides their omission or attenuation. Appendix Table~\ref{tab:dataset-condition-ranks} provides the corresponding condition-level breakdown. The comparison remains observational, since the datasets also differ in task, sensor count, and corruption difficulty.

% \textbf{The relative performance of TTA methods varies across sensing configurations.} On the IMU-dominated DSADS and PAMAP2 datasets, multimodal methods such as PTA are competitive; PTA is the strongest baseline on DSADS--MAESTRO and DSADS--Flex-MoE before rounding. On heterogeneous DaliaHAR and WESAD, TENT and EATA are at least as competitive: with Flex-MoE, both obtain 0.68 on DaliaHAR and 0.54 on WESAD, versus PTA's 0.60 and 0.50. Condition-level ranks show the same contrast: PTA's mean rank changes from 3.63 on DSADS to 7.97 on WESAD, compared with 3.56 and 4.32 for \method{} (Appendix Table~\ref{tab:dataset-condition-ranks}). This pattern motivates separating current abnormality from source-learned reliance, rather than deriving intervention solely from cross-modal agreement. Dataset differences also include tasks, sensor counts, and corruption difficulty, so the comparison does not isolate heterogeneity as the cause.

\noindent\begin{minipage}{\linewidth}
\setlength{\intextsep}{2pt}
\setlength{\columnsep}{9pt}
\begin{wrapfigure}{r}{0.37\linewidth}
\centering
\includegraphics[width=\linewidth]{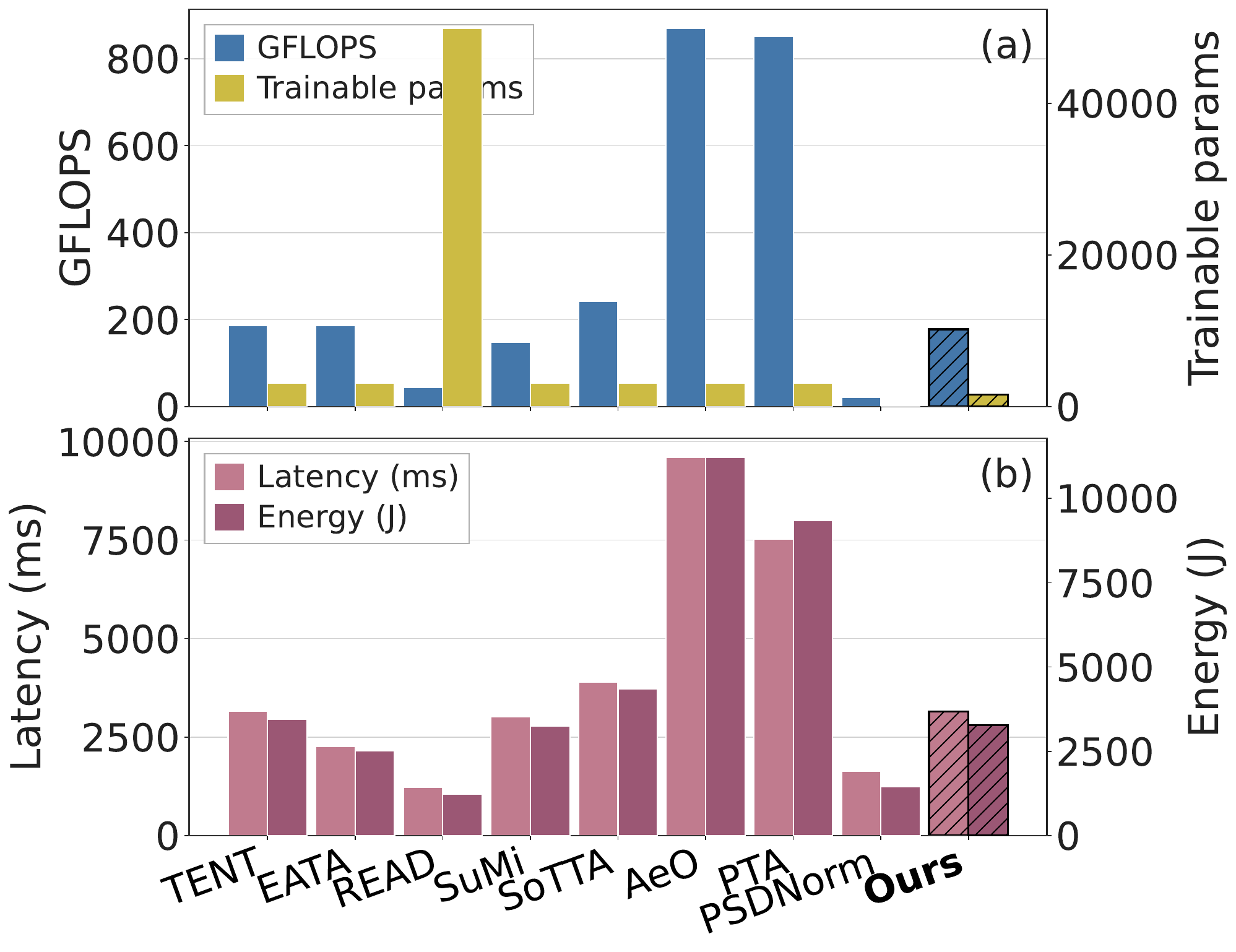}
\captionsetup{font=small,skip=2pt,justification=raggedright,singlelinecheck=false}
\caption{PAMAP2--MAESTRO efficiency for batches of 64, including detector overhead: GFLOPs, updated parameters, latency, and CPU energy. Full measurements: Appendix Table~\ref{tab:efficiency_dalia_pamap2}.}
\label{fig:efficiency}
\end{wrapfigure}
\textbf{\method{} combines modest adaptation compute with substantial runtime savings over multimodal TTA.} Across five profiled configurations, \method{} uses 9.49\% fewer GFLOPs and updates 47.82\% fewer parameters than EATA, measured by median within-configuration ratios. Relative to AeO and PTA, it reduces median latency by 71.03\% and 73.97\%, respectively. These computational savings coexist with implementation overhead: the reference implementation has 44.23\% higher latency and 37.37\% higher CPU energy than EATA. A separate DSADS--MAESTRO study demonstrates that this overhead can be substantially reduced through implementation optimization: batching and reusing detector computations lowers end-to-end latency by 26.3\%, from 1,713 to 1,262 ms, while preserving detector features, suspect flags, and predictions (Appendix Table~\ref{tab:detector-optimization}). Together, these results support \method{} as a practical default when adaptation FLOPs and the number of updated parameters are primary constraints, with further runtime savings demonstrated through detector optimization.
% \textbf{\method{} achieves substantial latency and energy savings over mulitmodal TTA baselines.} Including detector overhead, the PAMAP2--MAESTRO profile in Figure~\ref{fig:efficiency} shows 3.08$\times$ and 2.41$\times$ faster execution, with 71.03\% and 65.21\% less CPU energy, respectively. These savings extend across all five profiled configurations, with median within-configuration latency reductions of 71.03\% and 73.97\% relative to AeO and PTA. Compared with EATA, \method{} updates 47.82\% fewer parameters and uses 9.49\% fewer GFLOPs, with 44.23\% higher latency and 37.37\% higher CPU energy, measured by median within-configuration ratios. A separate DSADS--MAESTRO implementation study demonstrates further savings: batching and reusing detector computations reduces end-to-end latency by 26\% while preserving detector features, suspect flags, and predictions (Appendix Table~\ref{tab:detector-optimization}). Appendix~\ref{app:efficiency} reports the five reference profiles, including Static.
\par\vspace{2pt}\end{minipage}\par

\noindent\begin{minipage}{\linewidth}
\setlength{\intextsep}{2pt}
\setlength{\columnsep}{9pt}
\begin{wrapfigure}{r}{0.35\linewidth}
\centering
\includegraphics[width=0.85\linewidth]{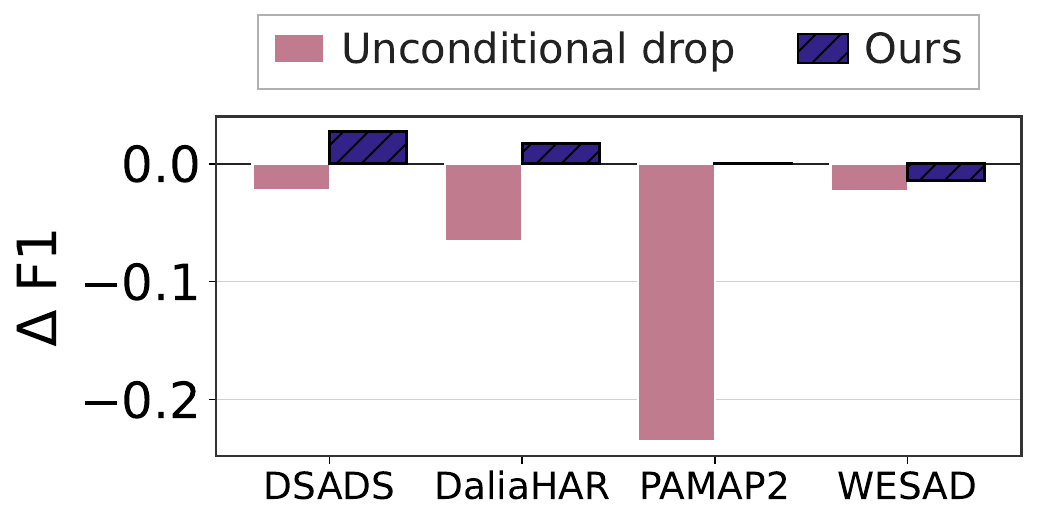}
\captionsetup{font=small,skip=2pt,justification=raggedright,singlelinecheck=false}
\caption{Absolute $\Delta$F1 relative to Static on MAESTRO, averaged across the evaluated modality-subset and corruption conditions. Full results: Appendix Table~\ref{apptab:detect_drop_full}.}
\label{fig:detect-drop}
\end{wrapfigure}
\textbf{Detecting corruption is insufficient to justify removing a modality.} Unconditional dropping replaces every suspect stream with the missingness symbol, bypassing the reliance-guided choice between removal and attenuation. In the MAESTRO comparison (Figure~\ref{fig:detect-drop}), PAMAP2 F1 falls from 0.92 for Static to 0.68 under unconditional dropping, while \method{} retains 0.92. On DaliaHAR, the corresponding scores are 0.74, 0.67, and 0.76 (Appendix Table~\ref{apptab:detect_drop_full}). These losses show that a missingness-resilient backbone alone does not make omission a safe response to every flagged input, supporting intervention that accounts for the information retained by suspect streams. Across Table~\ref{tab:overall-robustness}, \method{} improves on Static in 11 of 12 settings. The complete grids also include all-modality corruption, where unconditional removal would discard every stream (Appendix Tables~\ref{tab:appendix-dsads-maestro-fullgrid}--\ref{tab:appendix-dsads-flexmoe-fullgrid}).
\par\end{minipage}\par

\begingroup
\raggedbottom
\subsection{Ablation and Further Design Analyses}\label{sec:ablation}

We examine DaliaHAR--ShaSpec, whose non-uniform reliance profile distinguishes high-reliance wrist acceleration (\texttt{w\_acc}) from low-reliance temperature (\texttt{w\_temp}); \texttt{all\_wrist} tests joint wrist-sensor corruption, and Div. aggregates 50 conditions. Full reliance profiles are in Appendix~\ref{app:reliance_profiles}. 

\noindent\begin{minipage}{\linewidth}
\setlength{\intextsep}{2pt}
\setlength{\columnsep}{9pt}
\begin{wraptable}{r}{0.45\linewidth}
\centering
\captionsetup{font=small,skip=3pt,justification=raggedright,singlelinecheck=false}
\caption{\textbf{Component ablations.} Macro-F1 ($\uparrow$); \texttt{all\_w}: \texttt{all\_wrist}. R8 disables online updates; R9 changes source training and its prior. \textcolor{burgundy}{\textbf{Bold}}/\underline{underline}: best/second displayed score, including ties.}
\label{tab:ablation-components}
\fontsize{7.5}{8.3}\selectfont
\setlength{\tabcolsep}{2.2pt}
\renewcommand{\arraystretch}{0.98}
\begin{tabular}{@{}clcccc@{}}
\toprule
ID & Variant & \texttt{w\_acc} & \texttt{w\_temp} & \texttt{all\_w} & Div. \\
\midrule
R1 & Always drop & 0.63 & 0.70 & 0.61 & 0.58 \\
R2 & Always attenuate & \textcolor{burgundy}{\textbf{0.70}} & 0.73 & 0.66 & \textcolor{burgundy}{\textbf{0.63}} \\
R3 & Noisy flags & \underline{0.67} & 0.73 & 0.65 & \textcolor{burgundy}{\textbf{0.63}} \\
R4 & No detector & 0.66 & \textcolor{burgundy}{\textbf{0.75}} & 0.60 & 0.61 \\
R5 & Oracle flags & \textcolor{burgundy}{\textbf{0.70}} & \textcolor{burgundy}{\textbf{0.75}} & \textcolor{burgundy}{\textbf{0.68}} & \textcolor{burgundy}{\textbf{0.63}} \\
R6 & No attenuation & \underline{0.67} & \underline{0.74} & 0.66 & \underline{0.62} \\
\midrule
R8 & Frozen $a^j=\rho^j$ & 0.66 & \underline{0.74} & 0.64 & 0.60 \\
R9 & Uniform training & 0.59 & 0.59 & 0.59 & 0.60 \\
\midrule
\rowcolor{tolblue!10}
R7 & \textbf{\method{}} & \textcolor{burgundy}{\textbf{0.70}} & \textcolor{burgundy}{\textbf{0.75}} & \underline{0.67} & \textcolor{burgundy}{\textbf{0.63}} \\
\bottomrule
\end{tabular}
\end{wraptable}
\textbf{Selective intervention and online updates make complementary contributions.} Table~\ref{tab:ablation-components} links the framework's design choices to performance. Always dropping (R1) loses 7 F1 points on \texttt{w\_acc} and 5 on Div.; always attenuating (R2) ties \method{} on these two settings but trails on \texttt{w\_temp} and \texttt{all\_wrist}. Freezing attenuation and disabling the optimizer (R8) costs 4 points on \texttt{w\_acc} and 3 on Div., measuring the joint benefit of normalization and attenuation updates. Removing attenuation alone (R6) lowers all four scores. Detection matters most under joint wrist corruption: removing it (R4) reduces F1 from 0.67 to 0.60, whereas oracle flags (R5) raise it to 0.68. Noisy flags (R3) preserve Div. F1 but lose 3 points on \texttt{w\_acc}. The uniform-training control (R9) reduces Div. F1 to 0.60; it changes both source weights and the prior, so it measures their combined contribution.
\par\end{minipage}\par

\noindent\begin{minipage}{\linewidth}
\setlength{\intextsep}{2pt}
\setlength{\columnsep}{9pt}
\begin{wraptable}{r}{0.3\linewidth}
\centering
\captionsetup{font=scriptsize,skip=3pt,justification=raggedright,singlelinecheck=false}
\caption{\textbf{TTA across source-training regimes.} DaliaHAR--ShaSpec, Div. macro-F1 ($\uparrow$).}
\label{tab:curriculum-replacement}
\fontsize{7.5}{8.3}\selectfont
\setlength{\tabcolsep}{3pt}
\renewcommand{\arraystretch}{0.98}
\begin{tabular}{@{}lccc@{}}
\toprule
TTA & OGM & OPM & MIAM \\
\midrule
Static & 0.57 & 0.59 & 0.58 \\
EATA & 0.62 & 0.63 & 0.61 \\
\rowcolor{tolblue!10}
\method{} & \textcolor{burgundy}{\textbf{0.63}} & \textcolor{burgundy}{\textbf{0.65}} & \textcolor{burgundy}{\textbf{0.63}} \\
\bottomrule
\end{tabular}
\end{wraptable}
\textbf{Asymmetric training improves robustness to missing inputs.} Averaging the reported F1 means across the 12 settings in Appendix Table~\ref{tab:rec_ours}, asymmetric training improves on each backbone's original regime by 9.58 and 14.50 points at 20\% and 40\% missingness. Complete-input effects vary: DaliaHAR--ShaSpec changes from 0.77 to 0.74, while its 40\%-missing F1 rises from 0.39 to 0.61.

\textbf{\method{} remains effective across alternative source-training regimes.} On DaliaHAR--ShaSpec, \method{} exceeds the strongest evaluated TTA baseline by 1, 2, and 2 Div. F1 points with OGM-, OPM-, and MIAM-trained checkpoints, respectively (Table~\ref{tab:curriculum-replacement}); EATA matches the best baseline in each case. OPM-trained \method{} reaches 0.65, compared with 0.63 under the asymmetric curriculum (Table~\ref{tab:ablation-components}). These results support the portability of reliance-guided adaptation across the tested source-training regimes. The asymmetric curriculum's benefits for missing-input robustness are evaluated separately in Appendix Tables~\ref{tab:rec_ours} and~\ref{tab:curriculum-source-full}; Appendix Table~\ref{tab:curriculum-tta-full} gives the full TTA comparison.
\par\end{minipage}\par

\noindent\begin{minipage}{\linewidth}
\setlength{\intextsep}{2pt}
\setlength{\columnsep}{9pt}
\begin{wraptable}{r}{0.3\linewidth}
\centering
\captionsetup{font=scriptsize,skip=3pt,justification=raggedright,singlelinecheck=false}
\caption{\textbf{Prior perturbations.} Rev.: reversed ordering; Rand.: random; Jit.: order-preserving jitter.}
\label{tab:ablation-prior}
\fontsize{7.5}{8.3}\selectfont
\setlength{\tabcolsep}{2pt}
\renewcommand{\arraystretch}{0.98}
\begin{tabular}{@{}lcccc@{}}
\toprule
Prior & \texttt{w\_acc} & \texttt{w\_temp} & \texttt{all\_w} & Div. \\
\midrule
Rev. & 0.64 & 0.71 & 0.62 & \underline{0.61} \\
Rand. & \underline{0.67} & \underline{0.73} & \underline{0.65} & \textcolor{burgundy}{\textbf{0.63}} \\
Jit. & \textcolor{burgundy}{\textbf{0.70}} & \textcolor{burgundy}{\textbf{0.75}} & \textcolor{burgundy}{\textbf{0.67}} & \textcolor{burgundy}{\textbf{0.63}} \\
\midrule
\method{} & \textcolor{burgundy}{\textbf{0.70}} & \textcolor{burgundy}{\textbf{0.75}} & \textcolor{burgundy}{\textbf{0.67}} & \textcolor{burgundy}{\textbf{0.63}} \\
\bottomrule
\end{tabular}
\end{wraptable}
\textbf{Preserving the learned reliance ordering is more effective than reversing it.} With the source checkpoint fixed, reversing the prior lowers \texttt{w\_acc} F1 from 0.70 to 0.64 and Div. F1 from 0.63 to 0.61 (Table~\ref{tab:ablation-prior}). Order-preserving jitter matches \method{} at two-decimal precision in all four settings; random priors reduce the three named-scenario scores but tie on Div. These interventions support the usefulness of the learned profile and tolerance to the tested magnitude perturbations, without implying that routing is invariant to prior values.
\par\end{minipage}\par

\noindent\begin{minipage}{\linewidth}
\setlength{\intextsep}{2pt}
\setlength{\columnsep}{9pt}
\begin{wrapfigure}{r}{0.34\linewidth}
\centering
\includegraphics[width=\linewidth]{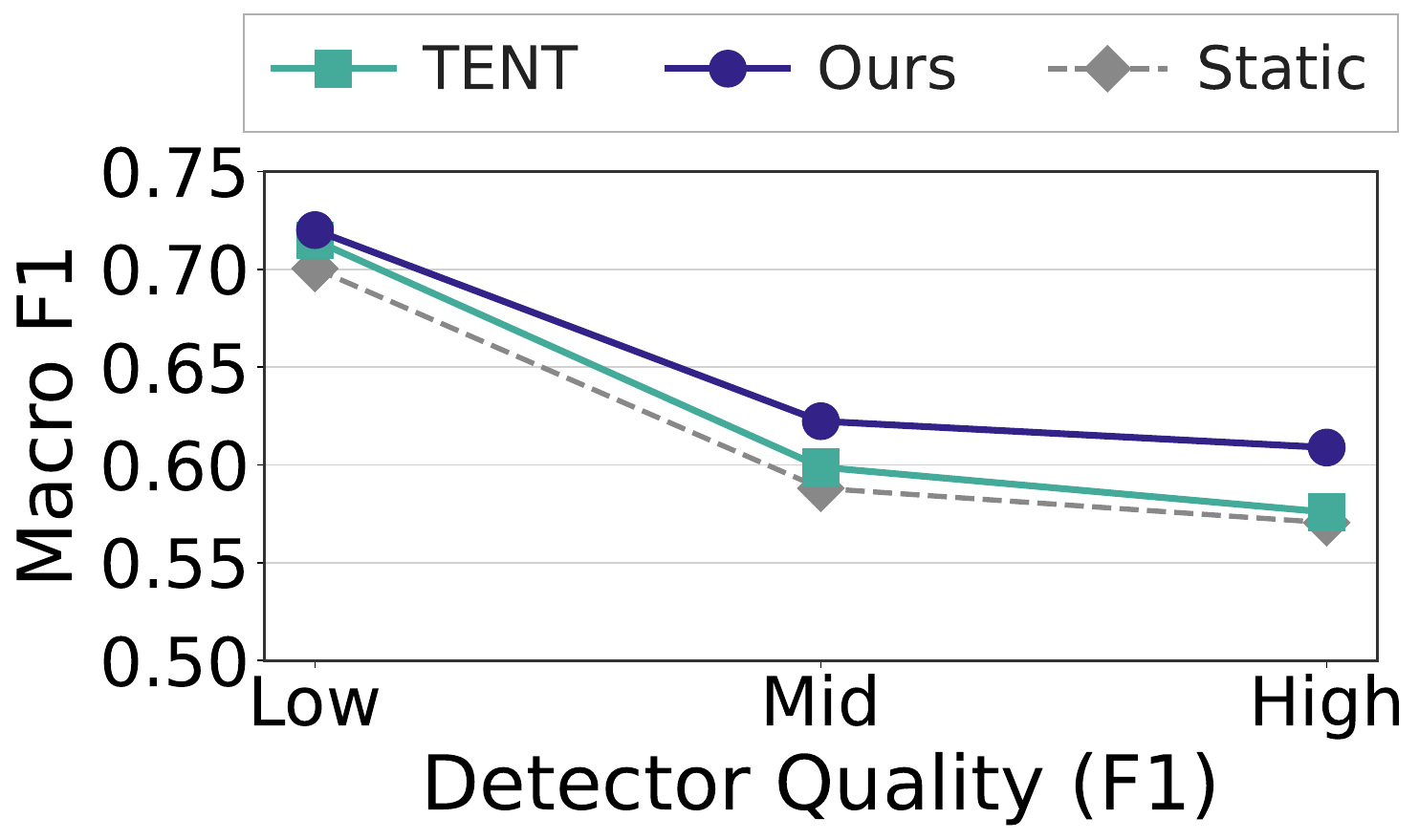}
\captionsetup{font=scriptsize,skip=2pt,justification=raggedright,singlelinecheck=false}
\caption{\textbf{Detector-quality bins on Div.}}
\label{fig:detector-sensitivity}
\end{wrapfigure}
\textbf{\method{} remains ahead of Static and TENT across detector-quality bins.} Its margin grows from low to high detector quality in Figure~\ref{fig:detector-sensitivity}, while absolute F1 falls for all methods. The bins contain different corruption instances, so this trend does not measure the causal effect of improving detector accuracy; the controlled flag ablations in Table~\ref{tab:ablation-components} isolate detector changes. Appendix~\ref{app:sensitivity} reports routing-threshold and batch-size sensitivity (Figures~\ref{fig:nu-sensitivity} and~\ref{fig:batchsize-sensitivity}).
\par\end{minipage}\par

\noindent\begin{minipage}{\linewidth}
\setlength{\intextsep}{2pt}
\setlength{\columnsep}{9pt}
\begin{wraptable}{r}{0.37\linewidth}
\centering
\captionsetup{font=scriptsize,skip=2pt,justification=raggedright,singlelinecheck=false}
\caption{\textbf{DaliaHAR--MAESTRO diagnostics.} T/E/B: wrist temp./EDA/BVP; C: chest acc. $\rho$: terminal dropout; $r$: normalized reliance.}
\label{tab:additional-diagnostics}
\fontsize{7}{7.6}\selectfont
\setlength{\tabcolsep}{1.5pt}
\setlength{\aboverulesep}{0.2ex}
\setlength{\belowrulesep}{0.2ex}
\renewcommand{\arraystretch}{0.95}
\begin{tabular}{@{}lccc@{}}
\toprule
\textbf{A. Noise} & \multicolumn{2}{c}{Max. $\rho$ (noised)} & F1 pts. \\
\midrule
None    & \multicolumn{2}{c}{--}   & 80.20$_{\pm1.40}$ \\
T       & \multicolumn{2}{c}{0.01} & 81.40 \\
T, E    & \multicolumn{2}{c}{0.00} & 82.50 \\
T, E, B & \multicolumn{2}{c}{0.00} & 80.40 \\
\midrule
\textbf{B. Width} & $100r^{\mathrm{C}}$ & $100r^{\mathrm{T}}$ & F1 pts. \\
\midrule
None        & 23.70$_{\pm13.00}$          & 9.80$_{\pm4.10}$           & 80.30$_{\pm0.70}$ \\
C $\times8$ & \textbf{26.60}$_{\pm11.60}$ & 21.30$_{\pm6.20}$          & 80.20$_{\pm0.70}$ \\
T $\times8$ & 23.40$_{\pm3.80}$           & \textbf{14.60}$_{\pm11.50}$ & 79.50$_{\pm2.10}$ \\
\bottomrule
\end{tabular}
\end{wraptable}
\textbf{Additional diagnostics probe uninformative inputs and modality projection width.} In Table~\ref{tab:additional-diagnostics}A, replacing one to three training modalities with variance-matched white noise yields terminal dropout probabilities $\rho^j$ of 0.00--0.01 for those streams and F1 of 80.40--82.50 points. Panel B instead reports normalized gradient-based reliance $r^j$: widening the chest or temperature projection eightfold increases its share by 2.90 or 4.80 percentage points, with F1 changes of $-0.10$ or $-0.80$ points. The profile responds to width, but the widened modality's share does not increase eightfold. These controls characterize the proxy under the tested training changes; they do not establish parameterization invariance. Appendix Tables~\ref{tab:white-noise-control} and~\ref{tab:width-control} give complete results.
\par\vspace{4pt}\end{minipage}

\par
\endgroup

%% file: sections/conclusion.tex
\vspace{-12pt}
\section{Conclusion and Future Work}
\vspace{-10pt}
\method{} amortizes reliance estimation through source training, separating it from runtime corruption detection to guide omission or attenuation without subset search. Across 12 equally weighted dataset--backbone settings, it leads overall macro-F1 by 1.58 points over EATA and achieves the best mean rank (2.42), with the largest gains on ShaSpec. PTA is competitive on IMU-dominated datasets, whereas TENT and EATA are stronger on heterogeneous datasets. These findings support \method{} as a practical option in the evaluated setting. Evaluation covers synthetic wearable-classification corruptions and backbones supporting trained omission and separable pre-fusion representations. The frozen, parameterization-dependent prior may become stale, and attenuation cannot ensure reliability when all streams are uninformative. Future work targets slow-timescale prior refresh, uncertainty-aware adaptation and abstention, finer suspect-set selection, and validation under natural failures across broader tasks and architectures.

%% file: sections/appendix.tex
\onecolumn
\appendix
\section*{\Large \centering Appendix}

% =============== APPENDIX (Section A or wherever) ===============
This Appendix provides additional details supporting the main paper, including corruption-generation procedures and detector construction (Sections~\ref{app:noise} and~\ref{app:detector_cost}), dataset and implementation details for all evaluated backbones and TTA baselines (Sections~\ref{app:dataset},~\ref{app:tta}, and~\ref{app:backbones}), analyses of the asymmetric modality-dropout curriculum and learned reliance profiles (Sections~\ref{app:curriculum},~\ref{app:reliance_sketch}, and~\ref{app:reliance_profiles}), additional sensitivity studies and ablations (Section~\ref{app:sensitivity}), and complete per-condition experimental results beyond the aggregates reported in the main paper. 

All experiments are implemented in PyTorch and conducted on Ubuntu servers equipped with NVIDIA TITAN RTX GPUs. Source-training experiments are repeated across three random seeds \{2711, 2712, 2713\}, and results are reported as mean $\pm$ standard deviation where applicable. 

For test-time adaptation evaluation, each method is evaluated across at least 10 modality-corruption scenarios and 5 corruption types per dataset/backbone combination under the episodic adaptation protocol.

% \section*{Reproducibility Statement}
% We provide a minimal source-code implementation in the Supplementary Materials. All datasets used in this work are publicly available, and detailed preprocessing, training, adaptation, and evaluation settings are documented throughout the following sections.

\section{Test-time Corruption Simulation Details}\label{app:noise}

We describe below each noise model used in this work. Each corruption is applied independently per channel to a univariate signal $x[t] \in \mathbb{R}$ for $t = 1, \ldots, T$, producing a corrupted signal $\tilde{x}[t]$. The set $\mathcal{P}$ is locally redefined per corruption type.

\noindent\textbf{Discrete Missingness}~\citep{little2019statistical}. A subset of $\lceil \rho T \rceil$ indices $\mathcal{P} \subset \{1, \ldots, T\}$ is sampled uniformly at random, and each selected sample is replaced with zero:
\begin{equation}
\tilde{x}[t] = 
\begin{cases}
0, & t \in \mathcal{P}, \\
x[t], & \text{otherwise}.
\end{cases}
\end{equation}
Zero-replacement (rather than masking) is used to remain consistent with downstream backbones that handle missing inputs via zero-imputation.

\noindent\textbf{Waveform Anomaly}~\citep{braei2020anomaly, bashar2020tanogan}. A subset of segments is randomly selected, and within each segment the original signal is replaced by one of two synthetic waveforms --- a square wave or a sinusoid --- chosen with equal probability:
\begin{equation}
\tilde{x}[t] = 
\begin{cases}
w[t], & t \in \mathcal{P}, \\
x[t], & \text{otherwise},
\end{cases}
\qquad
w[t] = 
\begin{cases}
c \sum_{k=0}^{L-1} \frac{1}{2k+1} \sin\!\big(2\pi (2k+1) f\, t\big), & \text{(square wave)} \\[4pt]
a \sin(2\pi f t), & \text{(sinusoid)}
\end{cases}
\end{equation}
where $\mathcal{P}$ is the union of the selected segments, each of radius $r$ centered at a random index. The constant $c$ is a free amplitude parameter (not the Fourier coefficient $4/\pi$ of an idealized square wave); $L$ controls the harmonic order of the truncated expansion.

\noindent\textbf{Nonlinear Sensor Degradation}~\citep{allen2007photoplethysmography}. The signal is amplitude-clipped at the $q$-th percentile of $|x[t]|$ and then compressed via a power law:
\begin{equation}
\tilde{x}[t] = \mathrm{sign}\!\big(\bar{x}[t]\big) \cdot \big|\bar{x}[t]\big|^{\beta}, \qquad \bar{x}[t] = \mathrm{clip}\!\left(x[t],\; -\tau_q,\; +\tau_q\right),
\end{equation}
where $\tau_q$ is the $q$-th percentile of $|x[t]|$ and $\beta \in (0, 1)$ controls compression. This models the saturation-then-compression behavior of degraded transducers.

\noindent\textbf{Frequency-Selective Attenuation}~\citep{oppenheim1999discrete}. A low-pass filter is applied in the frequency domain by attenuating components above a cutoff frequency $f_c$ by $A$\,dB:
\begin{equation}
\tilde{x}[t] = \mathcal{F}^{-1}\!\Big\{ M(f) \cdot \mathcal{F}\{x[t]\} \Big\}, \qquad M(f) = 
\begin{cases}
1, & f \leq f_c, \\
10^{-A/20}, & f > f_c,
\end{cases}
\end{equation}
where $\mathcal{F}$ denotes the discrete Fourier transform and $f_s$ is the modality's sampling rate.

\noindent\textbf{Nonstationary Noise}~\citep{orphanidou2017signal}. Additive Gaussian noise is injected with a time-varying standard deviation, such that the local SNR degrades linearly across the observation window:
\begin{equation}
\tilde{x}[t] = x[t] + \sigma[t]\, \eta[t], \qquad \eta[t] \sim \mathcal{N}(0, 1), \qquad \sigma^2[t] = \frac{\bar{P}_x}{10^{\,\mathrm{SNR}[t]/10}},
\end{equation}
with $\mathrm{SNR}[t] = \mathrm{SNR}_0 + (\mathrm{SNR}_T - \mathrm{SNR}_0)\, \tfrac{t-1}{T-1}$ and $\bar{P}_x = \tfrac{1}{T}\sum_t x[t]^2$.

\paragraph{Default parameters.} Table~\ref{tab:noise-defaults} summarizes the default values used across all experiments unless otherwise stated.

\begin{table}[h]
\centering
\caption{Default parameters for each test-time corruption.}
\label{tab:noise-defaults}
\setlength{\tabcolsep}{6pt}
\renewcommand{\arraystretch}{1.15}
\begin{tabular}{@{}ll@{}}
\toprule
\textbf{Corruption} & \textbf{Default parameters} \\
\midrule
Discrete missingness   & $\rho = 0.05$ \\
Waveform anomaly       & $\rho = 0.05$, $r = 5$, $c = 3.0$, $L = 5$, $f = 0.04$, $a = 2.0$ \\
Nonlinear degradation  & $q = 60$, $\beta = 0.3$ \\
Frequency attenuation  & $f_c = 0.3 \cdot f_s/2$, $A = 20$\,dB \\
Nonstationary noise    & $\mathrm{SNR}_0 = 30$\,dB, $\mathrm{SNR}_T = 5$\,dB \\
\bottomrule
\end{tabular}
\end{table}

\begin{sidewaysfigure}
    \centering
    \includegraphics[width=\textheight]{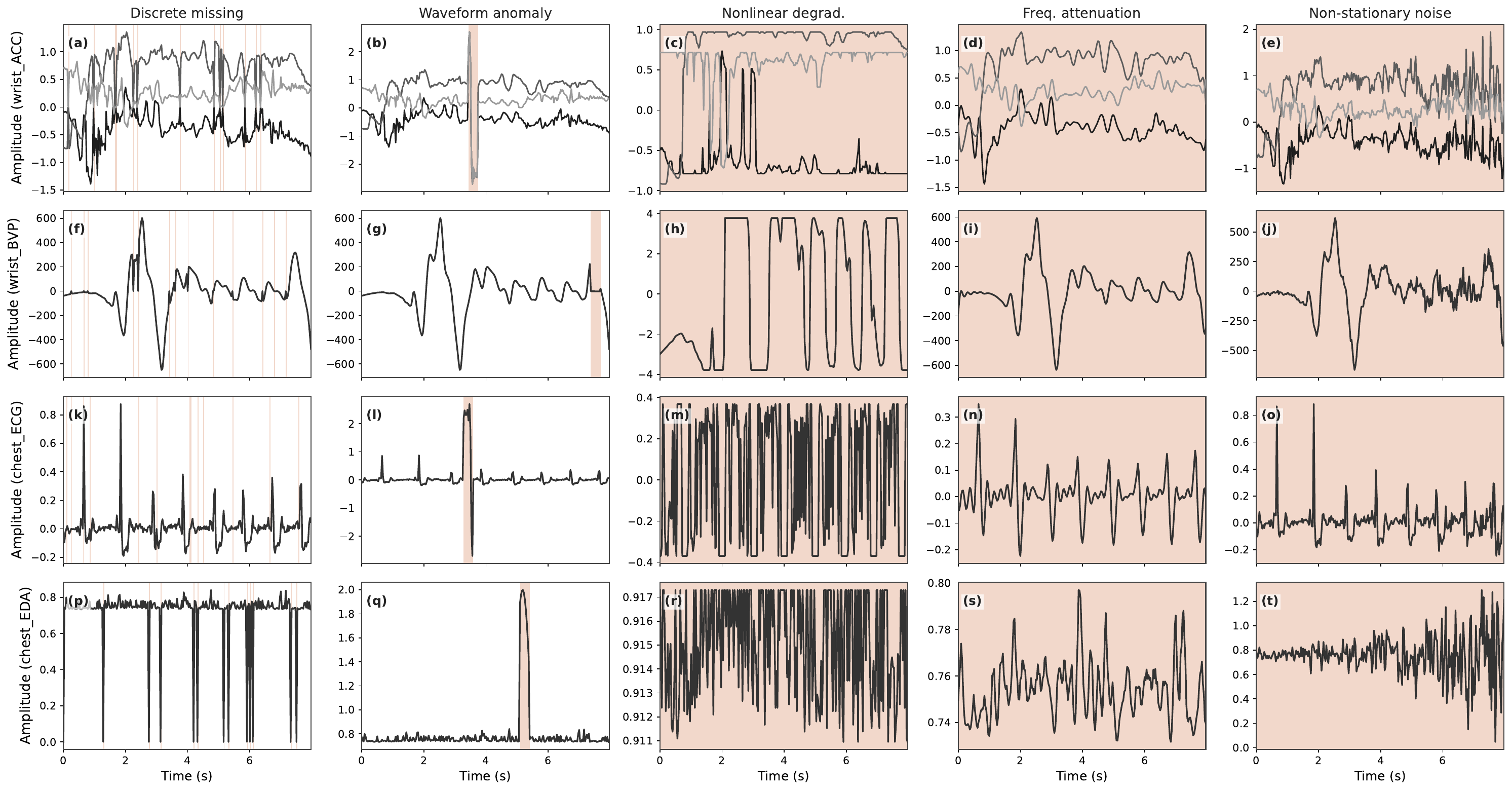}
    \caption{\textbf{Test-time noise injections across heterogeneous wearable sensors.} Rows show four modalities --- wrist\_ACC (motion), wrist\_BVP (photoplethysmography), chest\_ECG (electrocardiogram), and chest\_EDA (electrodermal activity). Columns show the five noise models from Appendix~\ref{app:noise}: \emph{discrete missingness} (sparse zero replacements), \emph{waveform anomaly} (segment substitution by a square wave or sinusoid), \emph{nonlinear sensor degradation} (amplitude clipping with power-law compression), \emph{frequency-selective attenuation} (high-frequency rolloff), and \emph{nonstationary noise} (linearly increasing noise variance). Orange highlights mark localized perturbations; shaded panels indicate window-wide corruption. The same mechanism produces qualitatively different distortions across modalities.}
\label{fig:noise_illustration}
\end{sidewaysfigure}

\subsection{Gradient-Flow Interpretation of the Reliance Profile}
\label{app:reliance_sketch}

We relate the reliance profile $\rho^j$ in Section~\ref{sec:reliance} to the gradient-flow framework of~\citet{lubana2020gradient}. The decomposition motivates the gradient signal used by the curriculum; the connection between the resulting proxy and predictive dependence is evaluated empirically.

\paragraph{Setup.}
Let $f_\theta$ have a disjoint parameter partition $\theta=(\theta_1,\ldots,\theta_M,\theta_{\mathrm{rest}})$, where $\theta_j$ is modality $j$'s input-projection block and $\theta_{\mathrm{rest}}$ collects the remaining encoder, fusion, and prediction parameters. For a fixed differentiable loss $\mathcal{L}(\theta)$, we consider Euclidean gradient flow. The selected blocks identify where gradients are measured; shared downstream parameters can also carry information from multiple modalities.

\paragraph{Gradient-flow decomposition.}
Under gradient flow,
\[
\dot{\theta} = -\nabla_\theta \mathcal{L}(\theta),
\]
the chain rule gives
\[
\dot{\mathcal{L}}(\theta)
=
\langle
\nabla_\theta \mathcal{L},
\dot{\theta}
\rangle
=
-\|\nabla_\theta \mathcal{L}\|_2^2.
\]
Partitioning the parameters into these blocks yields
\[
\dot{\mathcal{L}}(\theta)
=
-\sum_{j=1}^{M}
\|\nabla_{\theta_j}\mathcal{L}\|_2^2
-
\|\nabla_{\theta_{\mathrm{rest}}}\mathcal{L}\|_2^2.
\]
When the selected blocks have nonzero total squared-gradient norm, their normalized contributions to instantaneous loss reduction are
\[
\sigma^j(\theta)
=
\frac{
\|\nabla_{\theta_j}\mathcal{L}\|_2^2
}{
\sum_{\ell=1}^{M}
\|\nabla_{\theta_\ell}\mathcal{L}\|_2^2
},
\qquad
\sum_{j=1}^{M}\sigma^j(\theta)=1.
\]

\paragraph{Connection to the implemented profile.}
The main method first averages gradient norms over the training window, then normalizes them and applies the dropout schedule:
\[
\bar g_e^j=\frac{1}{W}\sum_{u=k_e-W}^{k_e-1}\|\nabla_{\theta_j}\mathcal{L}_u(\theta_u)\|_2,
\qquad
r_e^j=\frac{\bar g_e^j}{\sum_{\ell=1}^{M}\bar g_e^\ell},
\]
\[
\rho^j=p_{e^\star}^j
=\min\!\left(p_{\mathrm{cap}},\,M p_{\mathrm{base}}(e^\star)r_{e^\star}^j\right).
\]
Thus, $r_e^j$ is a normalized gradient share, whereas $\rho^j$ is the terminal dropout probability and need not sum to one. At a fixed parameter state, nonnegative gradient norms and their squares have the same ordering. Window averaging need not preserve the instantaneous ordering or the ordering of averaged squared norms, and clipping can introduce ties. Consequently, the implemented profile is an optimization-based proxy, rather than the exact loss-reduction share $\sigma^j$.

\paragraph{Connection to reliance-guided adaptation.}
The profile connects two stages of \method{}. During training, higher epoch-wise dropout probabilities increase exposure to inputs in which the corresponding modalities are absent. At deployment, the frozen terminal profile supplies the relative reliance mass used by the routing rule. Removal invokes the trained missingness operation, while attenuation retains downweighted suspect representations when the retained-reliance threshold is not met. These interventions act at different locations in the model; the construction does not establish equivalence of their representations or proximity to the source distribution. Their utility is assessed through the omission and prior-perturbation experiments in Section~\ref{sec:results}.

\paragraph{Scope of the interpretation.}
The gradient-flow decomposition holds for the specified parameterization and loss and requires no stationarity assumption. Finite-step optimization, adaptive preconditioning, stochastic batches, and changing dropout masks differ from this idealization. Parameter scaling and projection width can also change the measured norms. We therefore use $\rho^j$ as a backbone-specific routing prior, without interpreting it as a causal attribution of predictive information or a guarantee that a modality can be removed safely. The leave-one-modality-out comparison, prior perturbations, and width diagnostics provide empirical checks of this use.

\section{Details on Datasets}
\label{app:dataset}
% \yue{dataset details for pamap2}

We evaluate on WESAD~\citep{schmidt2018introducing}, DaliaHAR~\citep{reiss2019deep}, DSADS\citep{altun2010comparative}, and PAMAP2~\citep{reiss2012pamap2}. For WESAD, DaliaHAR, and DSADS, we follow the preprocessing protocol of MAESTRO~\citep{mohapatra2025maestro}, including the modality definitions, temporal alignment, resampling, windowing, label filtering, and subject-level split construction. We therefore do not repeat the full dataset specifications for these datasets here. Below, we summarize the PAMAP2 preprocessing used in our experiments, which follows the same window-based setup.

\begin{figure}[t]
\centering
\begin{minipage}[t]{0.47\textwidth}
\vspace{0pt}
\centering
\captionof{table}{PAMAP2 dataset modality details.}
\label{app_tab:pamap2}
\vspace{0.4em}
\small
\begin{tabular}{lcc}
\toprule
\textbf{Modality} & \textbf{Rate (Hz)} & \textbf{Variates} \\
\midrule
IMU\_hand & 100 & 17 \\
IMU\_chest & 100 & 17 \\
IMU\_ankle & 100 & 17 \\
heart\_rate & 9 & 1 \\
\bottomrule
\end{tabular}

\vspace{0.4em}
\footnotesize
Sampling rates denote the original sensor rates before resampling to 64\,Hz.
\end{minipage}
\hfill
\begin{minipage}[t]{0.47\textwidth}
\vspace{0pt}
\centering
\caption{Class distribution of the PAMAP2 activity labels.}
\label{app_fig:pamap2_dist}
\vspace{0.4em}
\includegraphics[width=\linewidth]{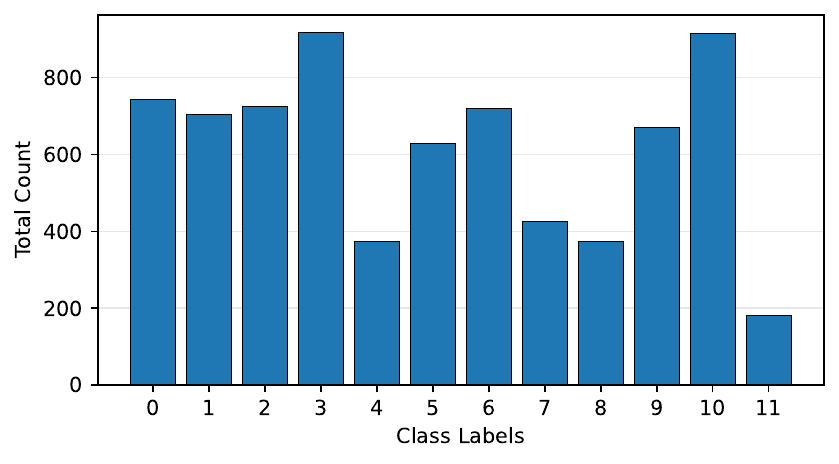}
\end{minipage}
\end{figure}

\noindent\textbf{PAMAP2.}
PAMAP2 is a wearable activity-recognition dataset collected from inertial sensors and a heart-rate monitor. We treat the hand, chest, and ankle IMUs as three separate modalities, and use the heart-rate signal as an additional modality. Each IMU contains 17 channels, including temperature, two accelerometer streams, gyroscope, magnetometer, and orientation signals. We remove transient activities, segment the synchronized recordings into 8-second windows, and resample each window to a common base rate of 64\,Hz, resulting in 512 timesteps per modality. We use 12 activity classes and subject-wise three-fold cross-validation. Table~\ref{app_tab:pamap2} and Figure~\ref{app_fig:pamap2_dist} summarize the processed PAMAP2 setup.

\section{More Details on Curriculum Dropout Schedule}
\label{app:curriculum}

Following~\citep{wang2023multi, NEURIPS2024_b2f2af54, mohapatra2025maestro}, we abstract the 
three-phase modality-dropout curriculum as a single scalar rate:
\[
p_{\text{base}}(e) = \begin{cases} 0 & e < 0.2E \\ p_{\max}\,\frac{e - 0.2E}{0.3E} & 0.2E \le e < 0.5E \\ p_{\max} & e \ge 0.5E \end{cases}
\]
where $p_{\max} \in [0,1]$ is the target missingness rate and $E$ the total number of training 
epochs. Existing methods treat $p_{\text{base}}(e)$ as the per-modality Bernoulli dropout 
probability, applied uniformly across all $M$ modalities.

\section{Detector Analyses}
\label{app:detector_cost}

\paragraph{Overview.}
Our suspect-modality detector is intentionally lightweight and serves
only as a coarse routing signal for downstream adaptation rather than
a standalone corruption classifier. For each sample $\mathcal{M}_i$
and modality $j$, the detector computes a corruption score $s_i^j$
from two complementary feature families: FITS reconstruction-error
statistics and log-spectral-distance (LSD) features relative to clean
source data. Modalities whose score exceeds a clean-reference
threshold are flagged as suspect and routed accordingly during
adaptation.

\paragraph{FITS reconstruction features.}
For each fold, we train a frequency-domain FITS reconstructor
$\mathcal{F}_\theta : \mathbb{R}^{T} \rightarrow \mathbb{R}^{T}$~\citep{xu2023fits}
on clean source data using a standard MSE objective for 30 epochs
($d{=}64$, learning rate $10^{-3}$, batch size $64$). FITS operates
in the DFT domain on the lowest $T/2$ frequency components,
capturing low-rank periodic structure while leaving higher-frequency
corruption in the residual. For each channel $c$, we compute residuals
\[
\epsilon_t = |\hat{x}_{i,t}^c - x_{i,t}^c|,
\]
and standardize them using clean-train statistics
$(\bar{\epsilon}^c, \sigma_\epsilon^c)$.
The resulting FITS feature block is
\[
\phi_{\text{FITS}}^c(x_i)
=
\Big[
\mu(\epsilon),
\max(\epsilon),
\mu(z),
\max(z),
\tfrac{1}{T}\sum_t \mathbb{1}[z_t>2],
\overline{\mathrm{top}_5}(z)
\Big],
\]
where
\[
z_t=(\epsilon_t-\bar{\epsilon}^c)/\sigma_\epsilon^c.
\]
These statistics capture residual magnitude, deviation from the clean
distribution, and sparsity/extremity of anomalies.

\paragraph{Log-spectral-distance features.}
We estimate the per-channel power spectral density
$\mathrm{PSD}(x_i^c)$ using Welch's method
($n_{\text{perseg}}=\min(T,64)$, $50\%$ overlap), and compute a
clean PSD barycenter
$\overline{\mathrm{PSD}}_{\text{src}}^c$
over the clean training set following
\citep{gnassounou2024multi, gnassounou2025psdnorm}.
The log-spectral distance between PSDs $p$ and $\bar p$ is
\[
\mathrm{LSD}(p,\bar p)
=
\Bigg[
\frac{1}{|\mathcal{F}|}
\sum_{f\in\mathcal{F}}
\big(10\log_{10}p(f)-10\log_{10}\bar p(f)\big)^2
\Bigg]^{1/2}.
\]
We additionally separate low/high-frequency bands using
$f_{\text{cut}}{=}4$\,Hz to capture localized spectral shifts.
The resulting LSD feature block is
\[
\phi_{\text{LSD}}^c(x_i)
=
\Big[
\mathrm{LSD}_{\text{tot}},
\mathrm{LSD}_{\text{low}},
\mathrm{LSD}_{\text{high}},
\Delta\arg\max,
\Delta H_{\text{spec}}
\Big],
\]
where $\Delta\arg\max$ measures peak-frequency drift and
$\Delta H_{\text{spec}}$ measures spectral-entropy deviation relative
to the clean PSD barycenter.

\paragraph{Per-modality detector formulation.}
For modality $j$ with channel set $C_j$, we concatenate the FITS and
LSD feature blocks across channels:
\[
\phi_i^j
=
\mathrm{concat}_{c\in C_j}
\big[
\phi_{\text{FITS}}^c(x_i),
\phi_{\text{LSD}}^c(x_i)
\big]
\in \mathbb{R}^{11|C_j|}.
\]
Features are standardized using clean-train statistics, and an
empirical Gaussian reference $(\mu^j,\Sigma^j)$ is fit on the
resulting vectors. The corruption score is computed using the squared
Mahalanobis distance
\[
s_i^j
=
(\phi_i^j-\mu^j)^\top
(\Sigma^j)^{-1}
(\phi_i^j-\mu^j).
\]
A modality is flagged as suspect when
\[
s_i^j > \delta^j,
\]
where $\delta^j$ is set to the 95th percentile of clean-train scores.
The resulting binary routing indicator is
\[
m_i^j = \mathbb{1}[s_i^j > \delta^j],
\]
and the suspect-modality set used by \method{} is
\[
\mathcal{S}_i = \{j : m_i^j = 1\}.
\]

All detector hyperparameters are held fixed across datasets
($z$-threshold $=2$, $f_{\text{cut}}{=}4$\,Hz, percentile $=95$,
FITS $d{=}64$ / 30 epochs).

\paragraph{Why both FITS and LSD?}
The two feature families capture complementary corruption signatures.
FITS residuals are sensitive to transient or localized deviations from
the clean source distribution, but may absorb broad spectral shifts
such as smooth attenuation. In contrast, LSD features capture global
spectral-shape deviations but are less sensitive to localized temporal
anomalies that average out in the PSD estimate. Empirically, combining
both yields more robust suspect-modality detection across diverse
corruption types.

\begin{figure}[!htbp]
    \centering
    \begin{minipage}[t]{0.48\linewidth}
        \centering
        \includegraphics[width=\linewidth]{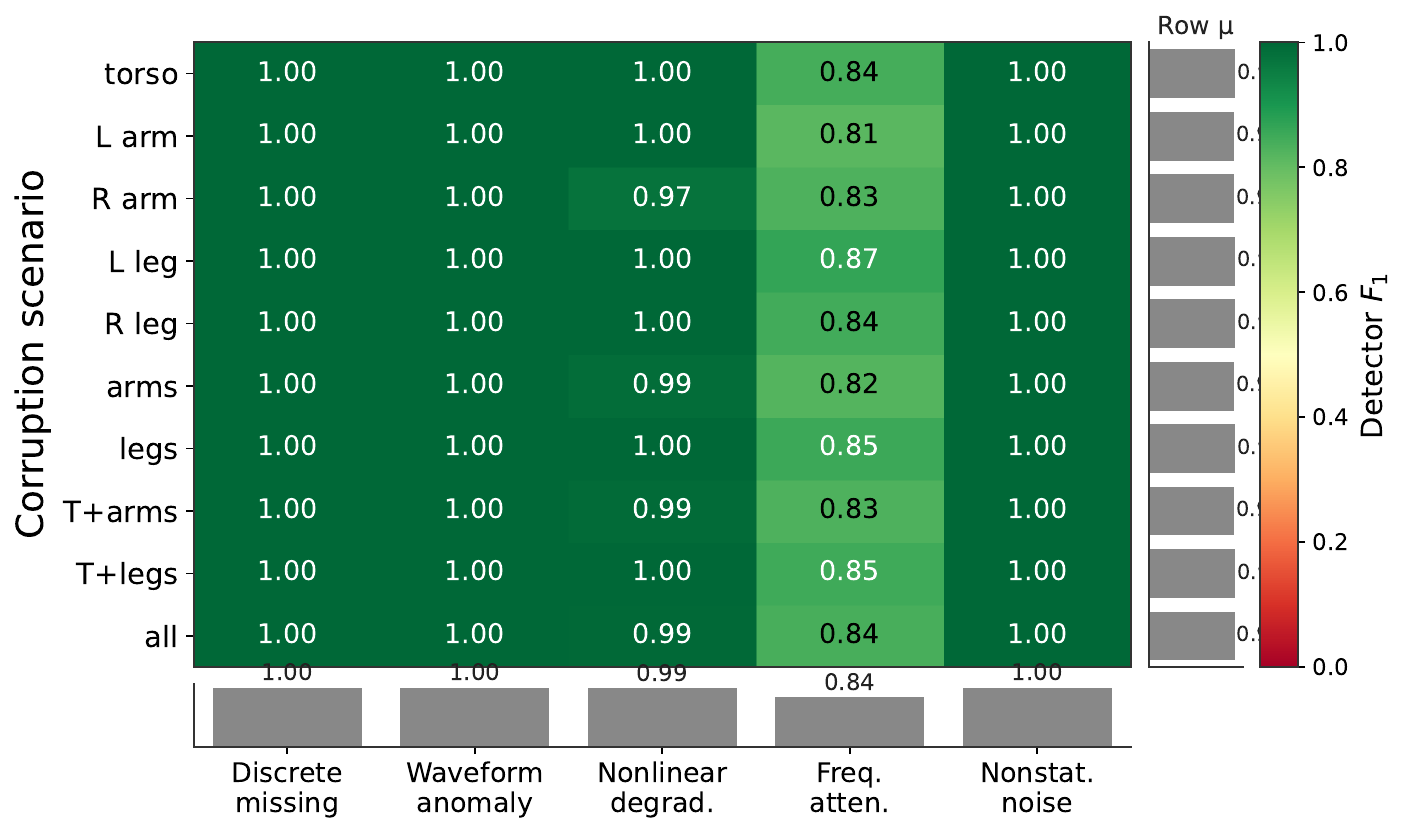}
        \captionof*{figure}{\small (a) DSADS}
    \end{minipage}%
    \hfill
    \begin{minipage}[t]{0.48\linewidth}
        \centering
        \includegraphics[width=\linewidth]{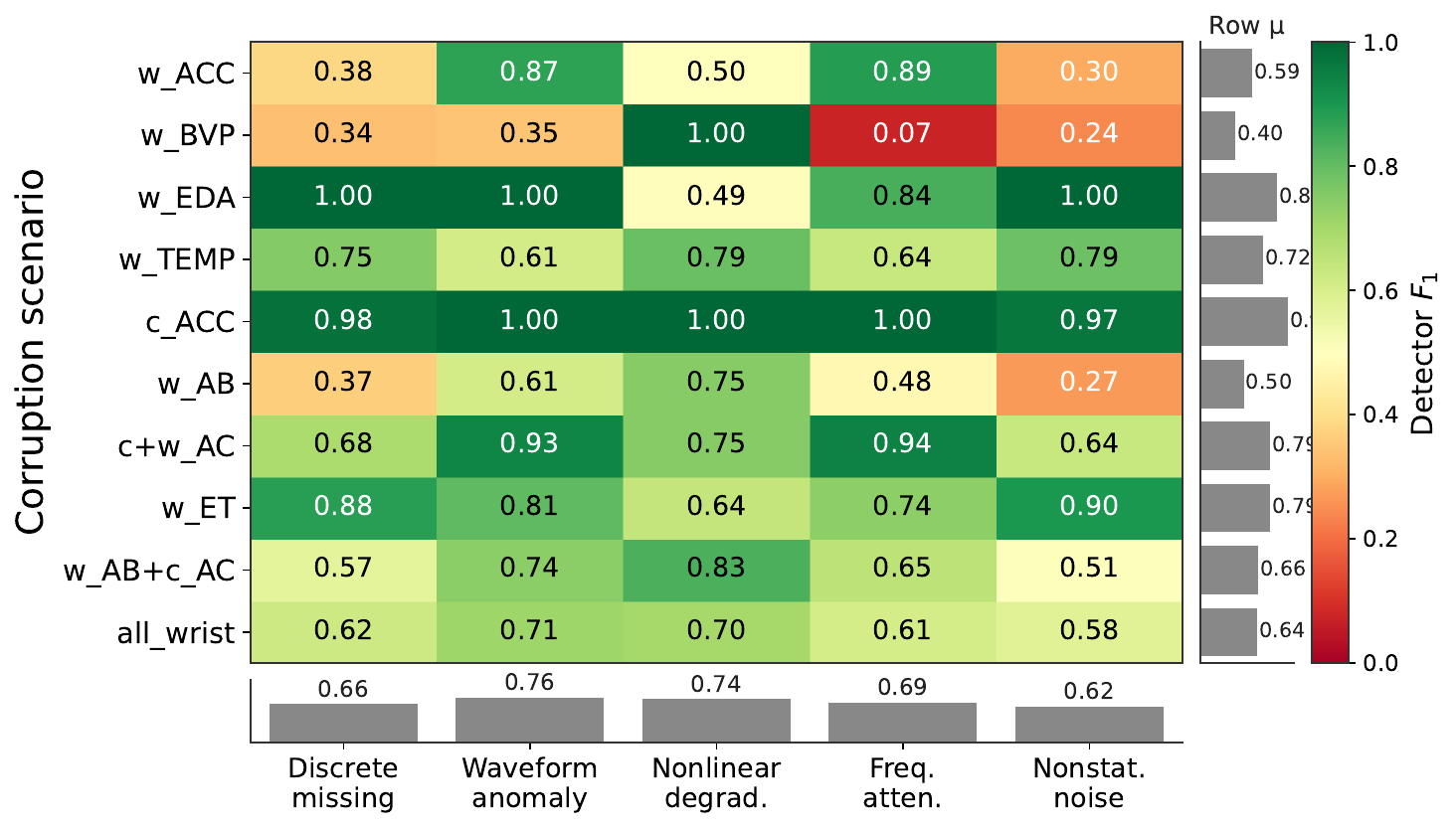}
        \captionof*{figure}{\small (b) DaliaHAR}
    \end{minipage}
    \vspace{6pt}
    \begin{minipage}[t]{0.48\linewidth}
        \centering
        \includegraphics[width=\linewidth]{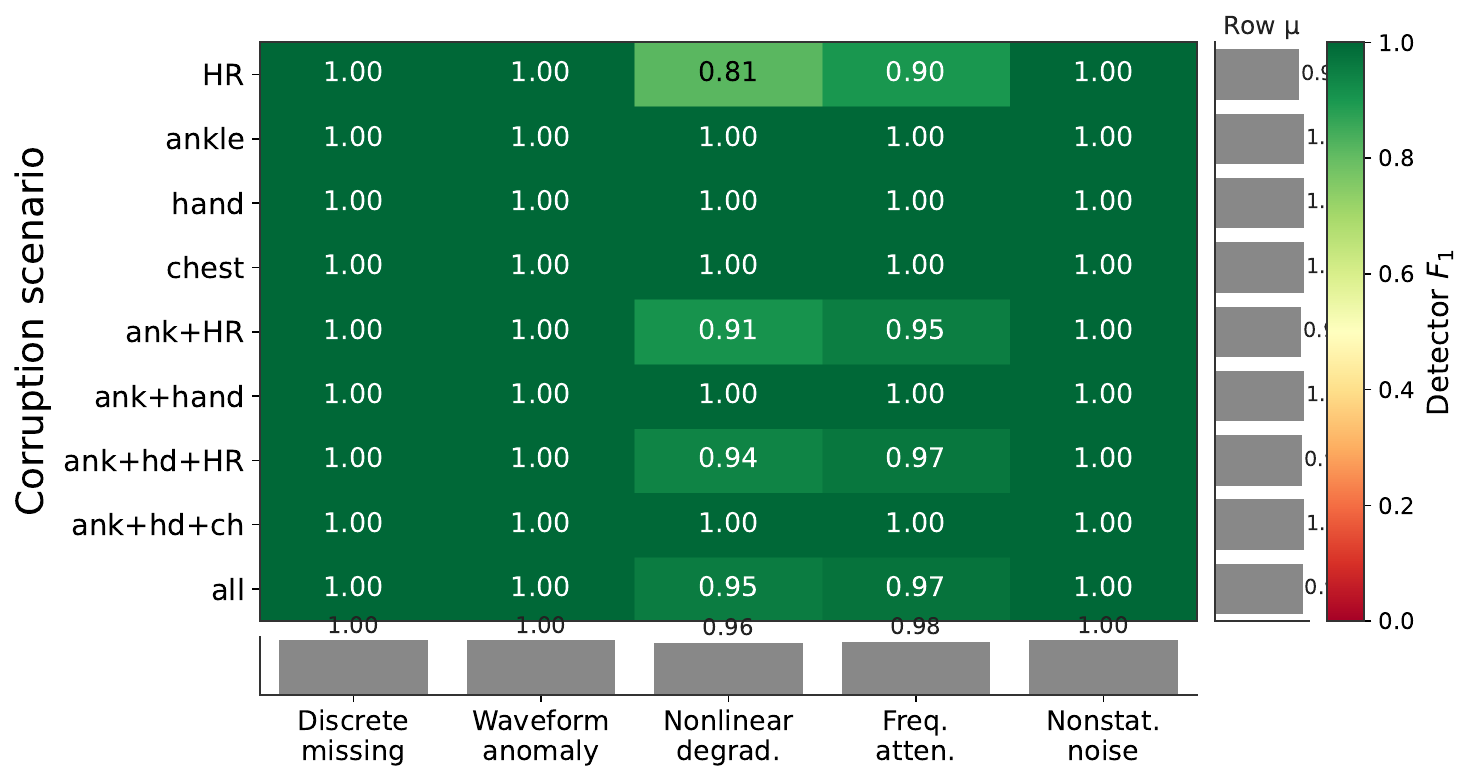}
        \captionof*{figure}{\small (c) PAMAP2}
    \end{minipage}%
    \hfill
    \begin{minipage}[t]{0.48\linewidth}
        \centering
        \includegraphics[width=\linewidth]{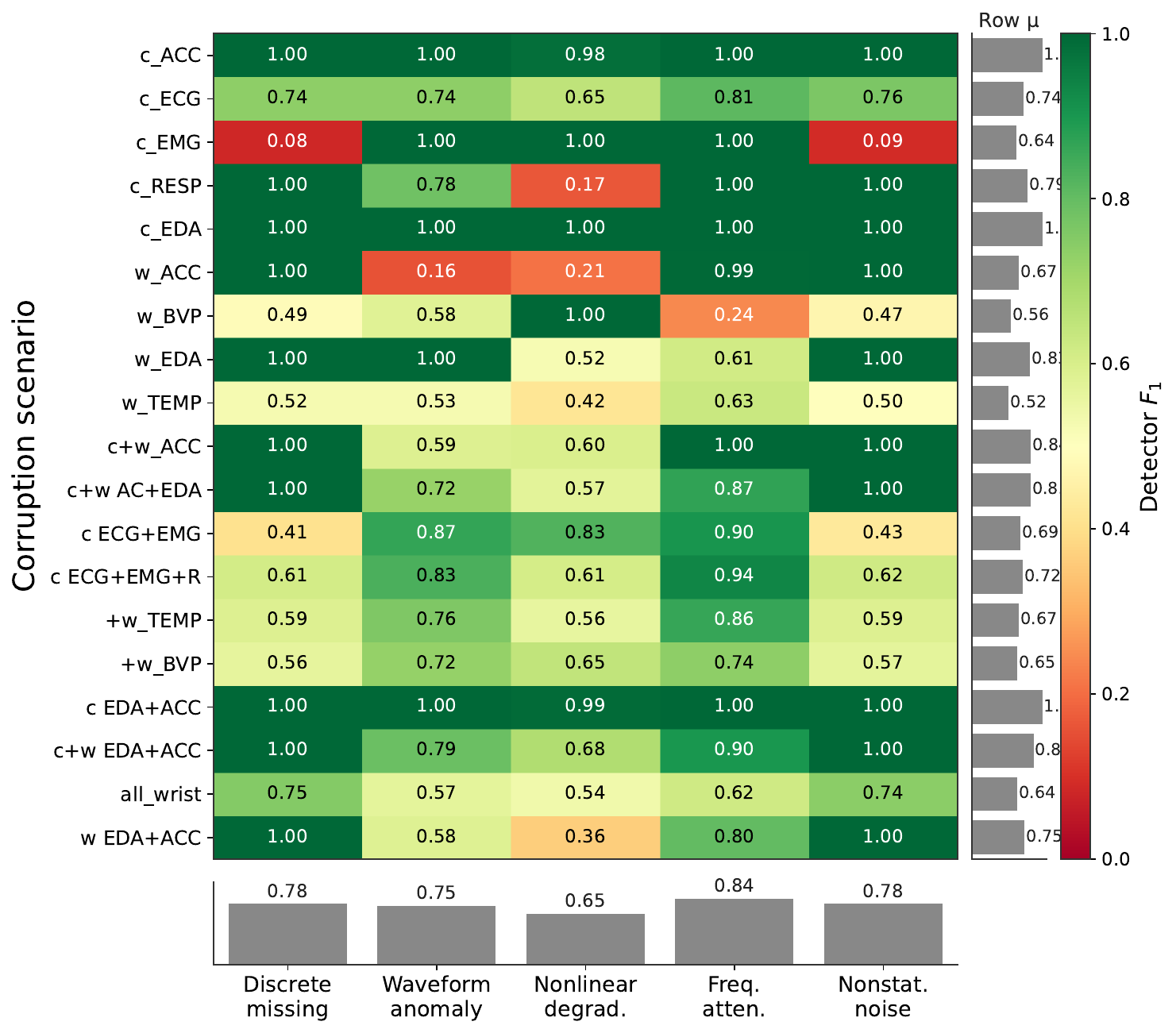}
        \captionof*{figure}{\small (d) WESAD}
    \end{minipage}
    \caption{\small Detector quality (F1 at flagging suspect modalities) across (modality, noise) combinations on each dataset. }
    \label{fig:detector-heatmaps}
\end{figure}

Figure~\ref{fig:detector-heatmaps} reports detector F1 scores across
(modality, corruption) combinations for all datasets. Performance is
generally strongest for structured corruption affecting spectral
content or waveform morphology, while localized corruption and mild
missingness are intentionally harder to separate from clean data under
the coarse one-class setting. Importantly, \method{} remains robust
under moderate detector errors because routing depends only on the
binary suspect indicator rather than calibrated corruption estimates.

\section{Results on Recommended and Asymmetric Curriculum Dropout Schedules}
\label{app:rec_vs_ours_results}

Table~\ref{tab:rec_ours} supports the discussion in Section~\ref{sec:results} by comparing the source checkpoints trained with each backbone's recommended regime and with our asymmetric modality-dropout curriculum under complete and missing-modality inputs with three seeds. We follow the same experimental setup as~\citet{mohapatra2025maestro}.

\begin{table}[t]
\centering
\caption{\small Performance under complete and missing modalities. Accuracy and F1-score ($\uparrow$) on DSADS, DaliaHAR, WESAD, and PAMAP2 across three backbones (MAESTRO, ShaSpec, FlexMoE) under \texttt{Complete}, \texttt{20\% missing}, and \texttt{40\% missing} conditions under their recommended training regime and our proposed asymmetric training regime.}
\label{tab:rec_ours}
\setlength{\tabcolsep}{4pt}
\renewcommand{\arraystretch}{1.15}
\resizebox{\textwidth}{!}{%
\begin{tabular}{@{}llc|ccc|ccc@{}}
\toprule
& & & \multicolumn{3}{c|}{\textbf{Accuracy}} & \multicolumn{3}{c}{\textbf{F1}} \\
\cmidrule(lr){4-6} \cmidrule(lr){7-9}
\textbf{Dataset} & \textbf{Backbone} & \textbf{Variant} 
 & Complete & 20\% miss. & 40\% miss. 
 & Complete & 20\% miss. & 40\% miss. \\
\midrule
% ===== DSADS =====
\multirow{6}{*}{DSADS}
 & \multirow{2}{*}{MAESTRO} & Original 
 & 0.86$_{\pm.00}$ & 0.82$_{\pm.01}$ & 0.78$_{\pm.01}$ 
 & 0.86$_{\pm.00}$ & 0.82$_{\pm.01}$ & 0.77$_{\pm.01}$ \\

 &                          & Ours     
 & 0.87$_{\pm.00}$ & 0.85$_{\pm.01}$ & 0.79$_{\pm.01}$ 
 & 0.86$_{\pm.00}$ & 0.83$_{\pm.01}$ & 0.78$_{\pm.01}$ \\
\cmidrule(l){2-9}
 & \multirow{2}{*}{ShaSpec} & Original & 0.81$_{\pm.01}$ & 0.45$_{\pm.02}$ & 0.22$_{\pm.01}$ & 0.79$_{\pm.00}$ & 0.45$_{\pm.02}$ & 0.20$_{\pm.02}$ \\
 &                          & Ours     & 0.78$_{\pm.00}$ & 0.75$_{\pm.00}$ & 0.64$_{\pm.03}$ & 0.77$_{\pm.00}$ & 0.74$_{\pm.00}$ & 0.64$_{\pm.03}$ \\
\cmidrule(l){2-9}
 & \multirow{2}{*}{FlexMoE} & Original & 0.67$_{\pm.01}$ & 0.35$_{\pm.09}$ & 0.19$_{\pm.06}$ & 0.63$_{\pm.03}$ & 0.34$_{\pm.08}$ & 0.17$_{\pm.07}$ \\
 &                          & Ours     & 0.74$_{\pm.00}$ & 0.73$_{\pm.01}$ & 0.60$_{\pm.04}$ & 0.74$_{\pm.00}$ & 0.73$_{\pm.01}$ & 0.60$_{\pm.05}$ \\
\midrule
% ===== DaliaHAR =====
\multirow{6}{*}{DaliaHAR}
 & \multirow{2}{*}{MAESTRO} & Original & 0.74$_{\pm.00}$ & 0.69$_{\pm.01}$ & 0.60$_{\pm.01}$ & 0.80$_{\pm.00}$ & 0.74$_{\pm.01}$ & 0.63$_{\pm.01}$ \\
 &                          & Ours     & 0.74$_{\pm.00}$ & 0.71$_{\pm.01}$ & 0.66$_{\pm.01}$ & 0.79$_{\pm.00}$ & 0.74$_{\pm.02}$ & 0.67$_{\pm.02}$ \\
\cmidrule(l){2-9}
 & \multirow{2}{*}{ShaSpec} & Original & 0.74$_{\pm.00}$ & 0.55$_{\pm.02}$ & 0.44$_{\pm.02}$ & 0.77$_{\pm.02}$ & 0.57$_{\pm.02}$ & 0.39$_{\pm.01}$ \\
 &                          & Ours     & 0.72$_{\pm.00}$ & 0.66$_{\pm.02}$ & 0.60$_{\pm.02}$ & 0.74$_{\pm.00}$ & 0.68$_{\pm.02}$ & 0.61$_{\pm.02}$ \\
\cmidrule(l){2-9}
 & \multirow{2}{*}{FlexMoE} & Original & 0.70$_{\pm.06}$ & 0.53$_{\pm.04}$ & 0.37$_{\pm.03}$ & 0.70$_{\pm.05}$ & 0.51$_{\pm.06}$ & 0.35$_{\pm.03}$ \\
 &                          & Ours     & 0.73$_{\pm.00}$ & 0.68$_{\pm.00}$ & 0.62$_{\pm.01}$ & 0.77$_{\pm.00}$ & 0.70$_{\pm.00}$ & 0.64$_{\pm.02}$ \\
\midrule
% ===== WESAD =====
\multirow{6}{*}{WESAD}
 & \multirow{2}{*}{MAESTRO} & Original & 0.67$_{\pm.00}$ & 0.64$_{\pm.02}$ & 0.56$_{\pm.05}$ & 0.60$_{\pm.00}$ & 0.57$_{\pm.02}$ & 0.47$_{\pm.05}$ \\
 &                          & Ours     & 0.77$_{\pm.00}$ & 0.74$_{\pm.01}$ & 0.73$_{\pm.01}$ & 0.66$_{\pm.00}$ & 0.64$_{\pm.01}$ & 0.63$_{\pm.01}$ \\
\cmidrule(l){2-9}
 & \multirow{2}{*}{ShaSpec} & Original & 0.65$_{\pm.06}$ & 0.53$_{\pm.05}$ & 0.43$_{\pm.04}$ & 0.54$_{\pm.04}$ & 0.47$_{\pm.04}$ & 0.41$_{\pm.03}$ \\
 &                          & Ours     & 0.74$_{\pm.00}$ & 0.70$_{\pm.01}$ & 0.70$_{\pm.02}$ & 0.56$_{\pm.00}$ & 0.54$_{\pm.00}$ & 0.54$_{\pm.02}$ \\
\cmidrule(l){2-9}
 & \multirow{2}{*}{FlexMoE} & Original & 0.69$_{\pm.00}$ & 0.69$_{\pm.01}$ & 0.65$_{\pm.03}$ & 0.55$_{\pm.00}$ & 0.55$_{\pm.01}$ & 0.50$_{\pm.05}$ \\
 &                          & Ours     & 0.65$_{\pm.00}$ & 0.66$_{\pm.01}$ & 0.62$_{\pm.02}$ & 0.55$_{\pm.00}$ & 0.56$_{\pm.01}$ & 0.51$_{\pm.03}$ \\
\midrule
% ===== PAMAP =====
\multirow{6}{*}{PAMAP2}
 & \multirow{2}{*}{MAESTRO} & Original & 0.95$_{\pm.00}$ & 0.93$_{\pm.00}$ & 0.84$_{\pm.01}$ & 0.95$_{\pm.00}$ & 0.92$_{\pm.00}$ & 0.85$_{\pm.00}$ \\
 &                          & Ours     & 0.95$_{\pm.00}$ & 0.93$_{\pm.00}$ & 0.86$_{\pm.01}$ & 0.95$_{\pm.00}$ & 0.93$_{\pm.00}$ & 0.86$_{\pm.00}$ \\
\cmidrule(l){2-9}
 & \multirow{2}{*}{ShaSpec} & Original & 0.86$_{\pm.00}$ & 0.80$_{\pm.02}$ & 0.71$_{\pm.03}$ & 0.86$_{\pm.00}$ & 0.81$_{\pm.02}$ & 0.73$_{\pm.03}$ \\
 &                          & Ours     & 0.85$_{\pm.00}$ & 0.80$_{\pm.02}$ & 0.72$_{\pm.03}$ & 0.86$_{\pm.00}$ & 0.81$_{\pm.02}$ & 0.73$_{\pm.03}$ \\
\cmidrule(l){2-9}
 & \multirow{2}{*}{FlexMoE} & Original & 0.87$_{\pm.00}$ & 0.83$_{\pm.01}$ & 0.78$_{\pm.03}$ & 0.88$_{\pm.00}$ & 0.84$_{\pm.01}$ & 0.79$_{\pm.03}$ \\
 &                          & Ours     & 0.90$_{\pm.00}$ & 0.84$_{\pm.00}$ & 0.79$_{\pm.05}$ & 0.89$_{\pm.00}$ & 0.84$_{\pm.00}$ & 0.79$_{\pm.05}$ \\
\bottomrule
\end{tabular}%
}
\end{table}

\section{Implementation Details on Baseline TTA Methods}
\label{app:tta}

We evaluate the eight TTA baselines described in Section~\ref{sec:results} under the same episodic protocol. In addition to these baselines, we report Static as the no-adaptation reference. For each test batch, the adaptation model is restored to the source checkpoint, adapted using only the current batch, and then used to predict the same batch. All baselines are evaluated on the same corrupted batches and modality-corruption scenarios.

Unless otherwise specified, we follow the recommended hyperparameter settings of each baseline and use Adam with learning rate $10^{-3}$ for test-time updates. Following the common TTA evaluation setting, all iterative adaptation methods are run for five steps unless the baseline recommends a single-update procedure. In our implementation, TENT~\citep{wang2020tent}, EATA~\citep{niu2022efficient}, SoTTA~\citep{gong2023sotta}, AeO~\citep{dong2025towards}, and PTA~\citep{wangpartition} use five adaptation steps. READ~\citep{yang2024test} uses one update step, and SuMi~\citep{guo2025smoothing} performs one update after its sample-selection stage, following their recommended adaptation procedures. Static and PSDNorm~\citep{gnassounou2025psdnorm} do not update model parameters. Here, $B$ denotes the batch size, $C$ the number of classes, $H(\cdot)$ the prediction entropy, and $p_i^{\max}$ the maximum predicted probability for sample $i$.

\paragraph{Static.}
Static is the frozen source model evaluated directly on the corrupted test batch. It provides the no-adaptation reference.

\paragraph{TENT.}
TENT updates the affine parameters of all LayerNorm modules by minimizing prediction entropy:
\[
\mathcal{L}_{\mathrm{Tent}}
=
-\frac{1}{B}\sum_{i=1}^{B}\sum_c p_{ic}\log p_{ic},
\]
where $p_i=\mathrm{softmax}(z_i)$. All modalities remain active with unit weights throughout adaptation.

\paragraph{EATA.}
EATA follows the same LayerNorm adaptation target as TENT, but applies an entropy-based reliability filter before the update. We use the entropy threshold $0.4\log C$, where $C$ is the number of classes. Adaptation is performed only when the filtered predictions provide a valid entropy-minimization signal; otherwise, the update is skipped.

\paragraph{READ.}
READ updates fusion-stage parameters with a reliability-aware objective. For MAESTRO and Flex-MoE, the trainable parameters are the query and key/value projections in the fusion attention module. For ShaSpec, which does not expose an explicit fusion-attention block, we adapt the Linear and LayerNorm parameters in its composition layer, classifier, and domain classifier. The loss is
\[
\mathcal{L}_{\mathrm{READ}}
=
\mathcal{L}_{\mathrm{ra}}-\mathcal{L}_{\mathrm{bal}},
\]
with
\[
\mathcal{L}_{\mathrm{ra}}
=
\frac{1}{B}\sum_i
p_i^{\max}\left(1-\log p_i^{\max}+\log \gamma\right),
\qquad
\gamma=e^{-1}.
\]
The balance term discourages degenerate batch-level predictions.

\paragraph{SuMi.}
SuMi identifies reliable samples using unimodal and multimodal predictions. We obtain unimodal predictions by activating one modality at a time and masking the remaining modalities. Samples are retained if they pass an interquartile-range feature filter and satisfy
\[
H_{\mathrm{uni}}(x_i)\ge \tau_{\mathrm{uni}},
\qquad
H_{\mathrm{multi}}(x_i)\le \tau_{\mathrm{multi}},
\]
where $\tau_{\mathrm{uni}}=e/10$ and $\tau_{\mathrm{multi}}=\log(C)/2-0.5$. On the selected samples, SuMi performs one LayerNorm update using entropy minimization together with a mutual-information sharing term that aligns unimodal and multimodal predictions via KL divergence.

\paragraph{SoTTA.}
SoTTA uses high-confidence uniform-class sampling followed by sharpness-aware entropy minimization. We use confidence threshold $0.99$ and first retain samples whose maximum softmax probability exceeds this threshold. The retained samples are capped per pseudo-class to form a class-balanced memory for the batch. SoTTA then applies a SAM-style entropy objective with perturbation radius $\rho=0.05$. If the high-confidence memory is empty, adaptation is skipped for that batch.

\paragraph{AeO.}
AeO applies an adaptive entropy objective coupled with cross-modal prediction discrepancy. For each sample, the normalized entropy defines the adaptive weight
\[
w_i=\tanh\left(\beta\left(H(p_i)/\log C-\alpha\right)\right),
\]
with $\alpha=0.5$ and $\beta=5.0$. The final objective combines adaptive entropy, unimodal entropy, pairwise cross-modal prediction discrepancy, and a class-diversity regularizer. We use $\gamma_1=0.5$ and $\gamma_2=0.1$, and update LayerNorm parameters.

\paragraph{PTA.}
PTA partitions the batch by combining confidence with predicted-class frequency. For sample $i$, we compute
\[
r_i = \max_c p_{ic}\left(1-f_{\hat{y}_i}\right),
\]
where $\hat{y}_i=\arg\max_c p_{ic}$ and $f_{\hat{y}_i}$ is the batch frequency of the predicted class. Samples above the 75th percentile of $r_i$ receive weight $2.0$, while samples below the 25th percentile receive weight $0.1$. PTA minimizes the resulting weighted entropy loss and adds a cross-modal alignment term that pulls unimodal predictions toward the detached multimodal prediction.

\paragraph{PSDNorm.}
PSDNorm is a non-parametric frequency-domain correction. We estimate clean source PSD barycenters from the source training set. At test time, when the detector flags a modality as corrupted, each channel in that modality is corrected before SAX tokenization using
\[
G(\omega)
=
\sqrt{
\frac{\mathrm{PSD}_{\mathrm{src}}(\omega)}
{\mathrm{PSD}_{\mathrm{test}}(\omega)+\epsilon}
}.
\]
Unflagged modalities are left unchanged, and no model parameters are updated.

\section{Implementation Details of different backbones}\label{app:backbones}
% \yue{please add details on the training of all threebackbones briefly.}

We evaluate \method{} on three missingness-resilient backbones: MAESTRO~\citep{mohapatra2025maestro}, ShaSpec~\citep{wang2023multi}, and Flex-MoE~\citep{NEURIPS2024_b2f2af54}. To keep the comparison across backbones fair, we preserve the original backbone architectures and supervised training objectives, while applying the asymmetric modality-dropout curriculum introduced in Section~\ref{sec:reliance} during source training. This curriculum changes the modality masking schedule but does not modify the backbone loss or add \method{}-specific modules to the source model. It produces the backbone-specific reliance profile used by \method{} at test time. In our implementation, all source backbones are trained on SAX-tokenized source data. The checkpoint with the best validation accuracy is used as the source checkpoint for \method{} and all baseline TTA methods. 

% All dataset-specific preprocessing, split construction, and optimization settings follow the corresponding training scripts.

\paragraph{MAESTRO.}
MAESTRO follows the original MAESTRO training objective and is optimized with cross-entropy loss over activity labels. During source training, we apply the asymmetric modality-dropout curriculum to estimate the MAESTRO-specific reliance profile. In the curriculum, the base modality-dropout probability is zero during the initial warm-up stage, increases linearly after warm-up, and is then held fixed. The per-modality dropout probabilities are assigned according to running gradient norms of the modality input projections, so modalities with larger estimated reliance are masked more often during training.

\paragraph{Flex-MoE.}
Flex-MoE follows the original Flex-MoE supervised objective. The training loss combines cross-entropy classification loss with the gate load-balancing loss from its fusion module:
\[
\mathcal{L}
=
\mathcal{L}_{\mathrm{ce}}
+
\lambda_{\mathrm{gate}}\mathcal{L}_{\mathrm{gate}},
\qquad
\lambda_{\mathrm{gate}}=10^{-2}.
\]
During source training, we apply the same asymmetric modality-dropout curriculum to estimate the Flex-MoE-specific reliance profile. The curriculum is applied to the modality inputs while keeping the Flex-MoE architecture and fusion objective unchanged.

\paragraph{ShaSpec.}
ShaSpec follows the original ShaSpec shared-specific training objective, which combines task classification, shared-feature alignment, and modality-specific domain classification:
\[
\mathcal{L}
=
\mathcal{L}_{\mathrm{ce}}
+
\alpha \mathcal{L}_{\mathrm{shared}}
+
\beta \mathcal{L}_{\mathrm{specific}},
\qquad
\alpha=0.1,\quad \beta=0.02.
\]
The shared loss aligns shared modality features, and the specific loss trains modality-domain predictions. During source training, we apply the same asymmetric modality-dropout curriculum to estimate the ShaSpec-specific reliance profile, while keeping the ShaSpec architecture and shared-specific objective unchanged.

%===========================================================
\subsection{Modality Reliance Profiles Across Backbones and Datasets}
\label{app:reliance_profiles}
%===========================================================

To better understand how different multimodal backbones distribute optimization importance across sensing streams, Tables~\ref{tab:reliance_daliahar}--\ref{tab:reliance_wesad} report the learned modality reliance for each dataset and backbone combination. Following Section~\ref{sec:reliance}, these weights are derived from the normalized gradient magnitudes accumulated near the deployed epoch and provide a proxy for each modality's contribution to loss reduction under gradient flow. Values are reported as mean $\pm$ standard deviation across the deployed epoch $\pm 1$ snapshot used for reliance estimation.

Several trends emerge. On heterogeneous wearable datasets such as WESAD and DaliaHAR, reliance distributions are highly non-uniform and strongly backbone-dependent, with certain modalities dominating optimization dynamics while others contribute minimally. In contrast, DSADS exhibits substantially more balanced reliance across body locations, reflecting its more homogeneous sensing structure. These observations support two key design choices in \method{}: (i) reliance should be estimated in a model-specific manner rather than assumed uniform across modalities, and (ii) routing decisions during test-time adaptation should preserve partially informative modalities rather than treating all corrupted streams equivalently.

\begin{table}[!htbp]
\centering
\caption{DaliaHAR -- modality reliance scores $\rho_j$.}
\label{tab:reliance_daliahar}
\begin{tabular}{lccc}
\toprule
Modality   & ShaSpec & FlexMoE & MAESTRO \\
\midrule
wrist\_ACC  & $0.80 \pm 0.00$ & $0.70 \pm 0.18$ & $0.65 \pm 0.05$ \\
wrist\_BVP  & $0.47 \pm 0.29$ & $0.19 \pm 0.01$ & $0.45 \pm 0.05$ \\
wrist\_EDA  & $0.71 \pm 0.16$ & $0.22 \pm 0.01$ & $0.13 \pm 0.01$ \\
wrist\_TEMP & $0.17 \pm 0.01$ & $0.22 \pm 0.02$ & $0.17 \pm 0.02$ \\
chest\_ACC  & $0.80 \pm 0.00$ & $0.58 \pm 0.22$ & $0.60 \pm 0.03$ \\
\bottomrule
\end{tabular}
\end{table}

\begin{table}[!htbp]
\centering
\caption{DSADS \textemdash{} modality reliance weights $\rho_j$.}
\label{tab:reliance_dsads}
\begin{tabular}{lccc}
\toprule
Modality  & ShaSpec & FlexMoE & MAESTRO \\
\midrule
torso      & $0.42 \pm 0.18$ & $0.42 \pm 0.03$ & $0.40 \pm 0.05$ \\
right\_arm & $0.37 \pm 0.09$ & $0.40 \pm 0.01$ & $0.39 \pm 0.06$ \\
left\_arm  & $0.45 \pm 0.11$ & $0.42 \pm 0.01$ & $0.41 \pm 0.07$ \\
right\_leg & $0.39 \pm 0.05$ & $0.39 \pm 0.02$ & $0.37 \pm 0.03$ \\
left\_leg  & $0.37 \pm 0.16$ & $0.37 \pm 0.03$ & $0.45 \pm 0.10$ \\
\bottomrule
\end{tabular}
\end{table}

\begin{table}[!htbp]
\centering
\caption{WESAD \textemdash{} modality reliance weights $\rho_j$.}
\label{tab:reliance_wesad}
\begin{tabular}{lccc}
\toprule
Modality   & ShaSpec & FlexMoE & MAESTRO \\
\midrule
chest\_ACC  & $0.80 \pm 0.00$ & $0.70 \pm 0.14$ & $0.78 \pm 0.02$ \\
chest\_ECG  & $0.80 \pm 0.00$ & $0.58 \pm 0.08$ & $0.00 \pm 0.00$ \\
chest\_EMG  & $0.01 \pm 0.00$ & $0.18 \pm 0.00$ & $0.00 \pm 0.00$ \\
chest\_RESP & $0.70 \pm 0.17$ & $0.30 \pm 0.01$ & $0.00 \pm 0.00$ \\
chest\_EDA  & $0.73 \pm 0.12$ & $0.28 \pm 0.02$ & $0.49 \pm 0.05$ \\
chest\_TEMP & $0.65 \pm 0.14$ & $0.23 \pm 0.02$ & $0.02 \pm 0.03$ \\
wrist\_ACC  & $0.80 \pm 0.00$ & $0.73 \pm 0.10$ & $0.75 \pm 0.08$ \\
wrist\_BVP  & $0.56 \pm 0.23$ & $0.33 \pm 0.02$ & $0.76 \pm 0.08$ \\
wrist\_EDA  & $0.72 \pm 0.15$ & $0.33 \pm 0.03$ & $0.06 \pm 0.02$ \\
wrist\_TEMP & $0.55 \pm 0.23$ & $0.27 \pm 0.00$ & $0.01 \pm 0.02$ \\
\bottomrule
\end{tabular}
\end{table}

%===========================================================
\subsection{Sensitivity study}
\label{app:sensitivity}
%===========================================================

\begin{figure}[h]
    \centering
    \captionsetup{font=scriptsize}

    \begin{minipage}[t]{0.48\linewidth}
        \centering
        \includegraphics[width=\linewidth]
        {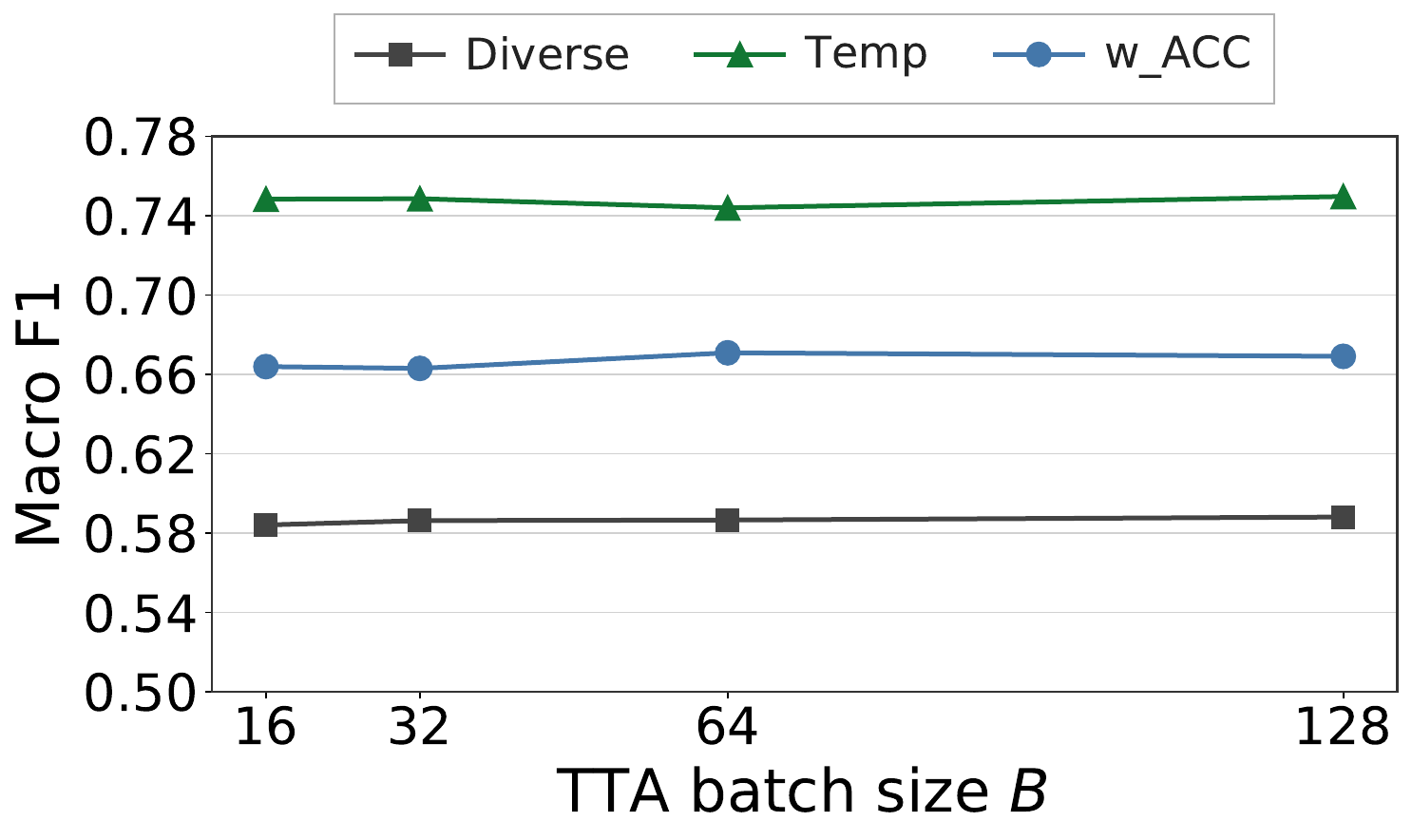}

        \vspace{-6pt}

        \caption{
        \textbf{Batch-size sensitivity.}
        Macro-F1 under varying episodic TTA batch sizes $B$ using the same DaliaHAR + ShaSpec setup as the ablation analyses. Performance remains stable across batch sizes on \texttt{Div.}, \texttt{w\_temp}, and \texttt{w\_acc}.
        }
        \label{fig:batchsize-sensitivity}
    \end{minipage}
    \hfill
    \begin{minipage}[t]{0.48\linewidth}
        \centering
        \includegraphics[width=\linewidth]
        {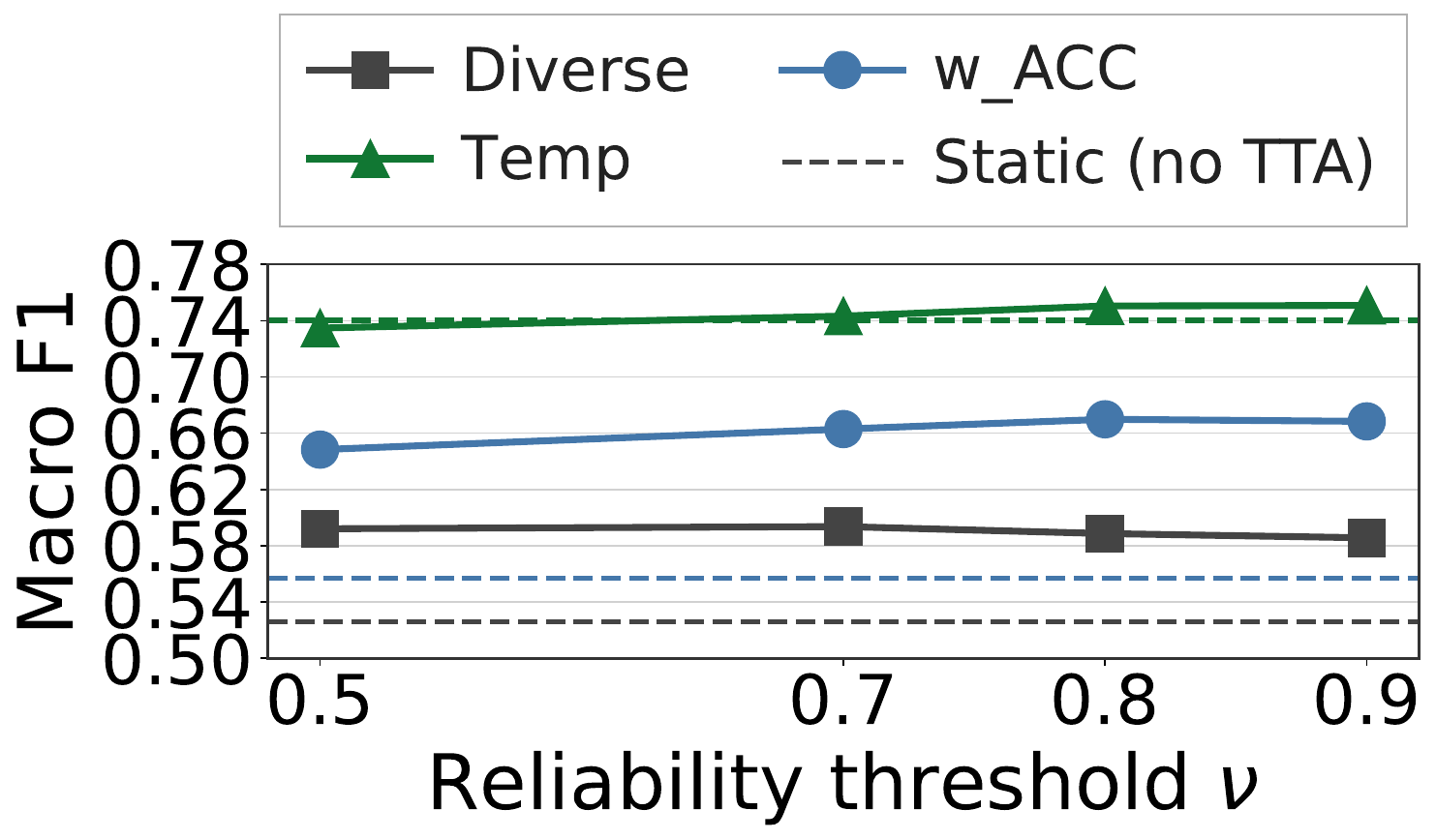}

        \vspace{-6pt}

        \caption{
        \textbf{Routing-threshold sensitivity.}
        Macro-F1 under varying routing thresholds $\nu$ using the same DaliaHAR + ShaSpec setup as the ablation analyses. Performance varies by less than $2\%$ relative macro-F1 across $\nu \in \{0.5,0.7,0.8,0.9\}$.
        }
        \label{fig:nu-sensitivity}
    \end{minipage}

    \vspace{-10pt}
\end{figure}

Figures~\ref{fig:batchsize-sensitivity} and~\ref{fig:nu-sensitivity} evaluate the sensitivity of \method{} to two adaptation hyperparameters using the same DaliaHAR + ShaSpec setup as the main ablations in Section~\ref{sec:ablation}. Across both the pooled diverse setting (\texttt{Div.}) and representative high-/low-reliance modalities (\texttt{w\_acc}, \texttt{w\_temp}), performance remains stable across a broad range of episodic TTA batch sizes and routing thresholds. In particular, varying the routing threshold $\nu$ across $\{0.5,0.7,0.8,0.9\}$ changes macro-F1 by less than $2\%$ relative, suggesting that the routing mechanism depends more on the coarse separation between high- and low-reliance corruption regimes than on precise threshold tuning. We fix $\nu = 0.8$ across all datasets, backbones, and corruption types in the main results. More importantly, $\nu$ exposes a useful system-level knob to the deployer: smaller $\nu$ shifts more suspect modalities into the modality-composition path (drop and replace with $s_0$), which avoids the per-modality attenuator forward pass and reduces test-time compute, at the cost of being more aggressive only toward genuinely low-reliance modalities; larger $\nu$ shifts more suspect modalities into the latent-attenuation path, preserving partially informative streams at modestly higher cost. This makes $\nu$ less of a hyperparameter to tune for accuracy and more of a deployment-time lever to trade efficiency against conservatism, with $\nu = 0.8$ providing a balanced default.

\section{Complete Results on DSADS dataset}\label{app:dsads_complete}

\noindent
For completeness, we provide the full DSADS results behind the aggregate numbers in the main text. 
Tables~\ref{tab:appendix-dsads-maestro-fullgrid}, 
\ref{tab:appendix-dsads-shaspec-fullgrid}, and 
\ref{tab:appendix-dsads-flexmoe-fullgrid} show the per-condition F1 scores for MAESTRO, ShaSpec, and Flex-MoE across corrupted modality subsets and noise types.

\begin{table}[!htbp]
\centering
\caption{Full per-modality-subset, per-noise F1-score ($\uparrow$) on \textbf{DSADS} with the \textbf{MAESTRO} backbone. Each cell is the F1 for one specific (corrupted modality subset, noise type) combination. Noise abbreviations: \emph{miss}=discrete missingness, \emph{wave}=waveform anomaly, \emph{nonlin}=nonlinear degradation, \emph{freq}=frequency attenuation, \emph{nonstat}=nonstationary noise. Clean baseline: F1 = 0.856.}
\label{tab:appendix-dsads-maestro-fullgrid}
\resizebox{1.0\textwidth}{!}{%
\begin{tabular}{@{}lllcccccccccc@{}}
\toprule
\textbf{Card.} & \textbf{Subset} & \textbf{Noise}
& Static & TENT & EATA & READ & SuMi & SoTTA & AeO & PTA & PSDNorm & \cellcolor{gray!10}\textbf{Ours} \\
\midrule

\multirow{5}{*}{1-mod} & \multirow{5}{*}{torso} & miss     & 0.803 & 0.799 & 0.788 & 0.796 & 0.779 & 0.792 & 0.786 & 0.805 & 0.741 & \cellcolor{gray!10}0.794 \\
& & wave     & 0.763 & 0.772 & 0.782 & 0.757 & 0.761 & 0.787 & 0.759 & 0.767 & 0.734 & \cellcolor{gray!10}0.799 \\
& & nonlin   & 0.850 & 0.854 & 0.851 & 0.847 & 0.849 & 0.853 & 0.849 & 0.856 & 0.822 & \cellcolor{gray!10}0.840 \\
& & freq     & 0.850 & 0.854 & 0.862 & 0.862 & 0.843 & 0.865 & 0.855 & 0.859 & 0.819 & \cellcolor{gray!10}0.856 \\
& & nonstat  & 0.809 & 0.783 & 0.790 & 0.793 & 0.783 & 0.787 & 0.783 & 0.789 & 0.745 & \cellcolor{gray!10}0.804 \\
\arrayrulecolor{black!20}\midrule
& \multirow{5}{*}{right\_arm} & miss     & 0.823 & 0.826 & 0.835 & 0.827 & 0.820 & 0.807 & 0.832 & 0.829 & 0.750 & \cellcolor{gray!10}0.824 \\
& & wave     & 0.798 & 0.797 & 0.802 & 0.795 & 0.796 & 0.810 & 0.796 & 0.808 & 0.755 & \cellcolor{gray!10}0.821 \\
& & nonlin   & 0.851 & 0.853 & 0.862 & 0.846 & 0.843 & 0.855 & 0.851 & 0.863 & 0.795 & \cellcolor{gray!10}0.856 \\
& & freq     & 0.842 & 0.852 & 0.858 & 0.843 & 0.848 & 0.853 & 0.850 & 0.851 & 0.800 & \cellcolor{gray!10}0.861 \\
& & nonstat  & 0.836 & 0.828 & 0.835 & 0.834 & 0.823 & 0.825 & 0.834 & 0.840 & 0.754 & \cellcolor{gray!10}0.840 \\
\arrayrulecolor{black!20}\midrule
& \multirow{5}{*}{left\_arm} & miss     & 0.794 & 0.802 & 0.792 & 0.792 & 0.784 & 0.789 & 0.798 & 0.794 & 0.714 & \cellcolor{gray!10}0.793 \\
& & wave     & 0.759 & 0.762 & 0.755 & 0.758 & 0.762 & 0.772 & 0.757 & 0.783 & 0.714 & \cellcolor{gray!10}0.774 \\
& & nonlin   & 0.857 & 0.851 & 0.855 & 0.850 & 0.851 & 0.859 & 0.855 & 0.869 & 0.772 & \cellcolor{gray!10}0.865 \\
& & freq     & 0.830 & 0.844 & 0.833 & 0.834 & 0.821 & 0.843 & 0.837 & 0.848 & 0.777 & \cellcolor{gray!10}0.837 \\
& & nonstat  & 0.803 & 0.798 & 0.802 & 0.795 & 0.802 & 0.793 & 0.784 & 0.810 & 0.718 & \cellcolor{gray!10}0.795 \\
\arrayrulecolor{black!20}\midrule
& \multirow{5}{*}{right\_leg} & miss     & 0.764 & 0.774 & 0.763 & 0.748 & 0.748 & 0.766 & 0.763 & 0.778 & 0.728 & \cellcolor{gray!10}0.838 \\
& & wave     & 0.705 & 0.723 & 0.730 & 0.680 & 0.711 & 0.728 & 0.722 & 0.761 & 0.689 & \cellcolor{gray!10}0.822 \\
& & nonlin   & 0.850 & 0.861 & 0.857 & 0.853 & 0.856 & 0.853 & 0.855 & 0.862 & 0.780 & \cellcolor{gray!10}0.835 \\
& & freq     & 0.829 & 0.828 & 0.830 & 0.825 & 0.842 & 0.838 & 0.853 & 0.834 & 0.787 & \cellcolor{gray!10}0.811 \\
& & nonstat  & 0.817 & 0.811 & 0.822 & 0.803 & 0.801 & 0.816 & 0.817 & 0.814 & 0.740 & \cellcolor{gray!10}0.834 \\
\arrayrulecolor{black!20}\midrule
& \multirow{5}{*}{left\_leg} & miss     & 0.808 & 0.797 & 0.802 & 0.792 & 0.799 & 0.806 & 0.801 & 0.808 & 0.721 & \cellcolor{gray!10}0.786 \\
& & wave     & 0.713 & 0.729 & 0.749 & 0.730 & 0.716 & 0.721 & 0.714 & 0.758 & 0.685 & \cellcolor{gray!10}0.731 \\
& & nonlin   & 0.837 & 0.851 & 0.856 & 0.853 & 0.845 & 0.841 & 0.844 & 0.862 & 0.767 & \cellcolor{gray!10}0.851 \\
& & freq     & 0.838 & 0.848 & 0.835 & 0.836 & 0.841 & 0.850 & 0.847 & 0.847 & 0.785 & \cellcolor{gray!10}0.848 \\
& & nonstat  & 0.812 & 0.826 & 0.809 & 0.801 & 0.807 & 0.814 & 0.798 & 0.822 & 0.744 & \cellcolor{gray!10}0.811 \\
\arrayrulecolor{black}\midrule
\multirow{5}{*}{2-mod} & \multirow{5}{*}{arms} & miss     & 0.720 & 0.727 & 0.722 & 0.700 & 0.699 & 0.714 & 0.704 & 0.713 & 0.610 & \cellcolor{gray!10}0.730 \\
& & wave     & 0.640 & 0.640 & 0.643 & 0.631 & 0.633 & 0.642 & 0.638 & 0.668 & 0.589 & \cellcolor{gray!10}0.670 \\
& & nonlin   & 0.855 & 0.853 & 0.855 & 0.848 & 0.852 & 0.858 & 0.860 & 0.867 & 0.727 & \cellcolor{gray!10}0.850 \\
& & freq     & 0.812 & 0.825 & 0.834 & 0.822 & 0.817 & 0.835 & 0.822 & 0.835 & 0.750 & \cellcolor{gray!10}0.822 \\
& & nonstat  & 0.705 & 0.713 & 0.706 & 0.709 & 0.679 & 0.695 & 0.687 & 0.712 & 0.630 & \cellcolor{gray!10}0.739 \\
\arrayrulecolor{black!20}\midrule
& \multirow{5}{*}{legs} & miss     & 0.602 & 0.592 & 0.603 & 0.605 & 0.603 & 0.616 & 0.618 & 0.635 & 0.412 & \cellcolor{gray!10}0.687 \\
& & wave     & 0.344 & 0.338 & 0.337 & 0.357 & 0.335 & 0.349 & 0.373 & 0.387 & 0.345 & \cellcolor{gray!10}0.527 \\
& & nonlin   & 0.839 & 0.844 & 0.852 & 0.845 & 0.853 & 0.852 & 0.843 & 0.855 & 0.679 & \cellcolor{gray!10}0.839 \\
& & freq     & 0.767 & 0.788 & 0.808 & 0.779 & 0.780 & 0.799 & 0.787 & 0.806 & 0.665 & \cellcolor{gray!10}0.781 \\
& & nonstat  & 0.680 & 0.668 & 0.668 & 0.661 & 0.657 & 0.667 & 0.648 & 0.684 & 0.547 & \cellcolor{gray!10}0.674 \\
\arrayrulecolor{black}\midrule
\multirow{5}{*}{3-mod} & \multirow{5}{*}{torso+arms} & miss     & 0.541 & 0.531 & 0.529 & 0.518 & 0.515 & 0.520 & 0.544 & 0.531 & 0.410 & \cellcolor{gray!10}0.622 \\
& & wave     & 0.377 & 0.378 & 0.373 & 0.359 & 0.379 & 0.384 & 0.386 & 0.422 & 0.331 & \cellcolor{gray!10}0.532 \\
& & nonlin   & 0.841 & 0.856 & 0.856 & 0.831 & 0.845 & 0.843 & 0.851 & 0.849 & 0.704 & \cellcolor{gray!10}0.841 \\
& & freq     & 0.757 & 0.796 & 0.785 & 0.788 & 0.782 & 0.787 & 0.783 & 0.796 & 0.711 & \cellcolor{gray!10}0.774 \\
& & nonstat  & 0.475 & 0.460 & 0.458 & 0.475 & 0.455 & 0.469 & 0.445 & 0.499 & 0.494 & \cellcolor{gray!10}0.561 \\
\arrayrulecolor{black!20}\midrule
& \multirow{5}{*}{torso+legs} & miss     & 0.466 & 0.432 & 0.426 & 0.445 & 0.423 & 0.469 & 0.462 & 0.497 & 0.168 & \cellcolor{gray!10}0.581 \\
& & wave     & 0.112 & 0.089 & 0.092 & 0.095 & 0.089 & 0.115 & 0.112 & 0.121 & 0.113 & \cellcolor{gray!10}0.267 \\
& & nonlin   & 0.834 & 0.849 & 0.840 & 0.837 & 0.819 & 0.831 & 0.831 & 0.844 & 0.550 & \cellcolor{gray!10}0.821 \\
& & freq     & 0.701 & 0.742 & 0.732 & 0.722 & 0.722 & 0.743 & 0.746 & 0.747 & 0.577 & \cellcolor{gray!10}0.731 \\
& & nonstat  & 0.432 & 0.428 & 0.432 & 0.417 & 0.411 & 0.428 & 0.432 & 0.446 & 0.338 & \cellcolor{gray!10}0.524 \\
\arrayrulecolor{black}\midrule
\multirow{5}{*}{5-mod} & \multirow{5}{*}{all} & miss     & 0.170 & 0.162 & 0.152 & 0.174 & 0.157 & 0.164 & 0.169 & 0.172 & 0.026 & \cellcolor{gray!10}0.184 \\
& & wave     & 0.012 & 0.009 & 0.009 & 0.011 & 0.016 & 0.017 & 0.011 & 0.016 & 0.016 & \cellcolor{gray!10}0.018 \\
& & nonlin   & 0.822 & 0.824 & 0.827 & 0.827 & 0.824 & 0.831 & 0.828 & 0.834 & 0.317 & \cellcolor{gray!10}0.812 \\
& & freq     & 0.594 & 0.595 & 0.591 & 0.610 & 0.612 & 0.599 & 0.610 & 0.622 & 0.404 & \cellcolor{gray!10}0.560 \\
& & nonstat  & 0.100 & 0.094 & 0.100 & 0.103 & 0.097 & 0.102 & 0.099 & 0.106 & 0.032 & \cellcolor{gray!10}0.122 \\
\arrayrulecolor{black}\bottomrule

\end{tabular}
}
\end{table}

\begin{table}[!htbp]
\centering
\caption{Full per-modality-subset, per-noise F1-score ($\uparrow$) on \textbf{DSADS} with the \textbf{ShaSpec} backbone. Each cell is the F1 for one specific (corrupted modality subset, noise type) combination. Noise abbreviations: \emph{miss}=discrete missingness, \emph{wave}=waveform anomaly, \emph{nonlin}=nonlinear degradation, \emph{freq}=frequency attenuation, \emph{nonstat}=nonstationary noise. Clean baseline: F1 = 0.765.}
\label{tab:appendix-dsads-shaspec-fullgrid}
\resizebox{1.0\textwidth}{!}{%
\begin{tabular}{@{}lllcccccccccc@{}}
\toprule
\textbf{Card.} & \textbf{Subset} & \textbf{Noise}
& Static & TENT & EATA & READ & SuMi & SoTTA & AeO & PTA & PSDNorm & \cellcolor{gray!10}\textbf{Ours} \\
\midrule

\multirow{5}{*}{1-mod} & \multirow{5}{*}{torso} & miss     & 0.531 & 0.520 & 0.521 & 0.506 & 0.516 & 0.522 & 0.529 & 0.519 & 0.541 & \cellcolor{gray!10}0.708 \\
& & wave     & 0.491 & 0.468 & 0.475 & 0.462 & 0.463 & 0.487 & 0.476 & 0.493 & 0.585 & \cellcolor{gray!10}0.689 \\
& & nonlin   & 0.617 & 0.602 & 0.632 & 0.615 & 0.610 & 0.625 & 0.618 & 0.611 & 0.686 & \cellcolor{gray!10}0.720 \\
& & freq     & 0.748 & 0.726 & 0.742 & 0.709 & 0.740 & 0.735 & 0.715 & 0.751 & 0.706 & \cellcolor{gray!10}0.721 \\
& & nonstat  & 0.719 & 0.676 & 0.699 & 0.658 & 0.693 & 0.688 & 0.670 & 0.696 & 0.653 & \cellcolor{gray!10}0.727 \\
\arrayrulecolor{black!20}\midrule
& \multirow{5}{*}{right\_arm} & miss     & 0.582 & 0.587 & 0.567 & 0.463 & 0.560 & 0.555 & 0.543 & 0.580 & 0.639 & \cellcolor{gray!10}0.680 \\
& & wave     & 0.497 & 0.496 & 0.478 & 0.451 & 0.469 & 0.461 & 0.474 & 0.490 & 0.593 & \cellcolor{gray!10}0.647 \\
& & nonlin   & 0.658 & 0.673 & 0.659 & 0.658 & 0.656 & 0.657 & 0.638 & 0.662 & 0.680 & \cellcolor{gray!10}0.707 \\
& & freq     & 0.775 & 0.763 & 0.754 & 0.694 & 0.764 & 0.763 & 0.753 & 0.773 & 0.723 & \cellcolor{gray!10}0.738 \\
& & nonstat  & 0.705 & 0.700 & 0.693 & 0.662 & 0.700 & 0.703 & 0.689 & 0.707 & 0.667 & \cellcolor{gray!10}0.711 \\
\arrayrulecolor{black!20}\midrule
& \multirow{5}{*}{left\_arm} & miss     & 0.496 & 0.496 & 0.512 & 0.472 & 0.497 & 0.490 & 0.488 & 0.527 & 0.636 & \cellcolor{gray!10}0.667 \\
& & wave     & 0.552 & 0.554 & 0.552 & 0.542 & 0.546 & 0.559 & 0.515 & 0.559 & 0.600 & \cellcolor{gray!10}0.679 \\
& & nonlin   & 0.677 & 0.648 & 0.668 & 0.603 & 0.663 & 0.678 & 0.622 & 0.668 & 0.681 & \cellcolor{gray!10}0.725 \\
& & freq     & 0.764 & 0.754 & 0.755 & 0.686 & 0.755 & 0.728 & 0.751 & 0.752 & 0.723 & \cellcolor{gray!10}0.743 \\
& & nonstat  & 0.727 & 0.722 & 0.701 & 0.647 & 0.703 & 0.705 & 0.698 & 0.723 & 0.680 & \cellcolor{gray!10}0.716 \\
\arrayrulecolor{black!20}\midrule
& \multirow{5}{*}{right\_leg} & miss     & 0.246 & 0.192 & 0.228 & 0.191 & 0.255 & 0.253 & 0.338 & 0.232 & 0.467 & \cellcolor{gray!10}0.553 \\
& & wave     & 0.178 & 0.118 & 0.148 & 0.124 & 0.185 & 0.167 & 0.257 & 0.145 & 0.224 & \cellcolor{gray!10}0.508 \\
& & nonlin   & 0.520 & 0.522 & 0.515 & 0.412 & 0.538 & 0.531 & 0.507 & 0.561 & 0.586 & \cellcolor{gray!10}0.644 \\
& & freq     & 0.758 & 0.736 & 0.736 & 0.671 & 0.725 & 0.720 & 0.727 & 0.739 & 0.638 & \cellcolor{gray!10}0.733 \\
& & nonstat  & 0.528 & 0.559 & 0.550 & 0.513 & 0.527 & 0.564 & 0.500 & 0.579 & 0.524 & \cellcolor{gray!10}0.628 \\
\arrayrulecolor{black!20}\midrule
& \multirow{5}{*}{left\_leg} & miss     & 0.229 & 0.106 & 0.112 & 0.127 & 0.224 & 0.191 & 0.391 & 0.210 & 0.541 & \cellcolor{gray!10}0.585 \\
& & wave     & 0.154 & 0.122 & 0.137 & 0.144 & 0.171 & 0.149 & 0.210 & 0.138 & 0.228 & \cellcolor{gray!10}0.479 \\
& & nonlin   & 0.549 & 0.526 & 0.541 & 0.512 & 0.546 & 0.552 & 0.548 & 0.569 & 0.618 & \cellcolor{gray!10}0.650 \\
& & freq     & 0.761 & 0.747 & 0.740 & 0.694 & 0.744 & 0.718 & 0.743 & 0.748 & 0.671 & \cellcolor{gray!10}0.734 \\
& & nonstat  & 0.593 & 0.608 & 0.592 & 0.573 & 0.571 & 0.608 & 0.525 & 0.625 & 0.544 & \cellcolor{gray!10}0.666 \\
\arrayrulecolor{black}\midrule
\multirow{5}{*}{2-mod} & \multirow{5}{*}{arms} & miss     & 0.259 & 0.240 & 0.240 & 0.242 & 0.254 & 0.270 & 0.277 & 0.248 & 0.450 & \cellcolor{gray!10}0.546 \\
& & wave     & 0.299 & 0.275 & 0.299 & 0.294 & 0.275 & 0.310 & 0.306 & 0.276 & 0.416 & \cellcolor{gray!10}0.554 \\
& & nonlin   & 0.569 & 0.533 & 0.545 & 0.502 & 0.537 & 0.550 & 0.553 & 0.573 & 0.591 & \cellcolor{gray!10}0.676 \\
& & freq     & 0.760 & 0.758 & 0.738 & 0.676 & 0.743 & 0.755 & 0.727 & 0.747 & 0.718 & \cellcolor{gray!10}0.724 \\
& & nonstat  & 0.652 & 0.650 & 0.654 & 0.591 & 0.629 & 0.625 & 0.612 & 0.656 & 0.590 & \cellcolor{gray!10}0.675 \\
\arrayrulecolor{black!20}\midrule
& \multirow{5}{*}{legs} & miss     & 0.038 & 0.017 & 0.012 & 0.011 & 0.039 & 0.027 & 0.060 & 0.025 & 0.116 & \cellcolor{gray!10}0.067 \\
& & wave     & 0.080 & 0.056 & 0.055 & 0.046 & 0.072 & 0.064 & 0.096 & 0.061 & 0.066 & \cellcolor{gray!10}0.123 \\
& & nonlin   & 0.243 & 0.173 & 0.176 & 0.173 & 0.235 & 0.241 & 0.319 & 0.248 & 0.246 & \cellcolor{gray!10}0.360 \\
& & freq     & 0.685 & 0.668 & 0.660 & 0.653 & 0.679 & 0.674 & 0.665 & 0.679 & 0.429 & \cellcolor{gray!10}0.689 \\
& & nonstat  & 0.109 & 0.042 & 0.054 & 0.034 & 0.083 & 0.101 & 0.176 & 0.090 & 0.126 & \cellcolor{gray!10}0.195 \\
\arrayrulecolor{black}\midrule
\multirow{5}{*}{3-mod} & \multirow{5}{*}{torso+arms} & miss     & 0.212 & 0.207 & 0.212 & 0.109 & 0.212 & 0.208 & 0.218 & 0.223 & 0.343 & \cellcolor{gray!10}0.436 \\
& & wave     & 0.168 & 0.157 & 0.168 & 0.138 & 0.177 & 0.173 & 0.189 & 0.157 & 0.277 & \cellcolor{gray!10}0.469 \\
& & nonlin   & 0.485 & 0.424 & 0.414 & 0.361 & 0.478 & 0.455 & 0.478 & 0.470 & 0.490 & \cellcolor{gray!10}0.609 \\
& & freq     & 0.728 & 0.723 & 0.727 & 0.658 & 0.696 & 0.699 & 0.706 & 0.722 & 0.668 & \cellcolor{gray!10}0.697 \\
& & nonstat  & 0.525 & 0.499 & 0.511 & 0.448 & 0.508 & 0.493 & 0.466 & 0.533 & 0.475 & \cellcolor{gray!10}0.610 \\
\arrayrulecolor{black!20}\midrule
& \multirow{5}{*}{torso+legs} & miss     & 0.028 & 0.024 & 0.028 & 0.042 & 0.032 & 0.033 & 0.046 & 0.027 & 0.062 & \cellcolor{gray!10}0.041 \\
& & wave     & 0.044 & 0.042 & 0.044 & 0.050 & 0.043 & 0.042 & 0.052 & 0.043 & 0.034 & \cellcolor{gray!10}0.068 \\
& & nonlin   & 0.263 & 0.215 & 0.214 & 0.204 & 0.267 & 0.250 & 0.303 & 0.254 & 0.174 & \cellcolor{gray!10}0.278 \\
& & freq     & 0.684 & 0.681 & 0.675 & 0.624 & 0.670 & 0.660 & 0.656 & 0.675 & 0.366 & \cellcolor{gray!10}0.670 \\
& & nonstat  & 0.076 & 0.052 & 0.058 & 0.048 & 0.079 & 0.088 & 0.086 & 0.057 & 0.071 & \cellcolor{gray!10}0.112 \\
\arrayrulecolor{black}\midrule
\multirow{5}{*}{5-mod} & \multirow{5}{*}{all} & miss     & 0.018 & 0.016 & 0.018 & 0.020 & 0.022 & 0.018 & 0.025 & 0.017 & 0.012 & \cellcolor{gray!10}0.017 \\
& & wave     & 0.024 & 0.024 & 0.024 & 0.022 & 0.031 & 0.030 & 0.028 & 0.021 & 0.016 & \cellcolor{gray!10}0.024 \\
& & nonlin   & 0.153 & 0.083 & 0.123 & 0.092 & 0.155 & 0.148 & 0.215 & 0.096 & 0.097 & \cellcolor{gray!10}0.140 \\
& & freq     & 0.653 & 0.645 & 0.632 & 0.589 & 0.644 & 0.645 & 0.620 & 0.635 & 0.247 & \cellcolor{gray!10}0.631 \\
& & nonstat  & 0.021 & 0.014 & 0.020 & 0.019 & 0.032 & 0.027 & 0.034 & 0.015 & 0.030 & \cellcolor{gray!10}0.022 \\
\arrayrulecolor{black}\bottomrule

\end{tabular}
}
\end{table}

\begin{table}[!htbp]
\centering
\caption{Full per-modality-subset, per-noise F1-score ($\uparrow$) on \textbf{DSADS} with the \textbf{Flex-MoE} backbone. Each cell is the F1 for one specific (corrupted modality subset, noise type) combination. Noise abbreviations: \emph{miss}=discrete missingness, \emph{wave}=waveform anomaly, \emph{nonlin}=nonlinear degradation, \emph{freq}=frequency attenuation, \emph{nonstat}=nonstationary noise. Clean baseline: F1 = 0.740.}
\label{tab:appendix-dsads-flexmoe-fullgrid}
\resizebox{1.0\textwidth}{!}{%
\begin{tabular}{@{}lllcccccccccc@{}}
\toprule
\textbf{Card.} & \textbf{Subset} & \textbf{Noise}
& Static & TENT & EATA & READ & SuMi & SoTTA & AeO & PTA & PSDNorm & \cellcolor{gray!10}\textbf{Ours} \\
\midrule

\multirow{5}{*}{1-mod} & \multirow{5}{*}{torso} & miss     & 0.680 & 0.691 & 0.695 & 0.671 & 0.680 & 0.701 & 0.679 & 0.682 & 0.644 & \cellcolor{gray!10}0.719 \\
& & wave     & 0.670 & 0.697 & 0.706 & 0.696 & 0.700 & 0.704 & 0.698 & 0.710 & 0.613 & \cellcolor{gray!10}0.719 \\
& & nonlin   & 0.693 & 0.693 & 0.682 & 0.679 & 0.683 & 0.698 & 0.684 & 0.687 & 0.637 & \cellcolor{gray!10}0.709 \\
& & freq     & 0.703 & 0.733 & 0.733 & 0.737 & 0.725 & 0.738 & 0.715 & 0.739 & 0.660 & \cellcolor{gray!10}0.718 \\
& & nonstat  & 0.670 & 0.713 & 0.705 & 0.690 & 0.685 & 0.709 & 0.694 & 0.703 & 0.626 & \cellcolor{gray!10}0.685 \\
\arrayrulecolor{black!20}\midrule
& \multirow{5}{*}{right\_arm} & miss     & 0.683 & 0.697 & 0.686 & 0.696 & 0.688 & 0.694 & 0.675 & 0.676 & 0.646 & \cellcolor{gray!10}0.699 \\
& & wave     & 0.655 & 0.689 & 0.687 & 0.685 & 0.673 & 0.692 & 0.674 & 0.680 & 0.610 & \cellcolor{gray!10}0.698 \\
& & nonlin   & 0.687 & 0.716 & 0.704 & 0.711 & 0.715 & 0.698 & 0.703 & 0.716 & 0.659 & \cellcolor{gray!10}0.692 \\
& & freq     & 0.733 & 0.747 & 0.767 & 0.765 & 0.762 & 0.748 & 0.733 & 0.752 & 0.681 & \cellcolor{gray!10}0.753 \\
& & nonstat  & 0.683 & 0.707 & 0.700 & 0.698 & 0.717 & 0.693 & 0.684 & 0.704 & 0.643 & \cellcolor{gray!10}0.694 \\
\arrayrulecolor{black!20}\midrule
& \multirow{5}{*}{left\_arm} & miss     & 0.685 & 0.667 & 0.675 & 0.698 & 0.669 & 0.674 & 0.665 & 0.690 & 0.619 & \cellcolor{gray!10}0.706 \\
& & wave     & 0.703 & 0.718 & 0.718 & 0.712 & 0.706 & 0.699 & 0.705 & 0.715 & 0.647 & \cellcolor{gray!10}0.712 \\
& & nonlin   & 0.700 & 0.707 & 0.700 & 0.706 & 0.712 & 0.702 & 0.699 & 0.702 & 0.640 & \cellcolor{gray!10}0.742 \\
& & freq     & 0.746 & 0.759 & 0.760 & 0.737 & 0.752 & 0.738 & 0.754 & 0.763 & 0.685 & \cellcolor{gray!10}0.746 \\
& & nonstat  & 0.720 & 0.730 & 0.727 & 0.704 & 0.713 & 0.714 & 0.721 & 0.724 & 0.651 & \cellcolor{gray!10}0.712 \\
\arrayrulecolor{black!20}\midrule
& \multirow{5}{*}{right\_leg} & miss     & 0.627 & 0.668 & 0.659 & 0.663 & 0.637 & 0.663 & 0.633 & 0.665 & 0.534 & \cellcolor{gray!10}0.705 \\
& & wave     & 0.640 & 0.645 & 0.624 & 0.655 & 0.652 & 0.659 & 0.642 & 0.658 & 0.535 & \cellcolor{gray!10}0.704 \\
& & nonlin   & 0.736 & 0.771 & 0.763 & 0.744 & 0.732 & 0.744 & 0.734 & 0.754 & 0.597 & \cellcolor{gray!10}0.725 \\
& & freq     & 0.671 & 0.681 & 0.697 & 0.715 & 0.693 & 0.682 & 0.688 & 0.705 & 0.567 & \cellcolor{gray!10}0.672 \\
& & nonstat  & 0.532 & 0.575 & 0.569 & 0.646 & 0.559 & 0.555 & 0.563 & 0.604 & 0.469 & \cellcolor{gray!10}0.663 \\
\arrayrulecolor{black!20}\midrule
& \multirow{5}{*}{left\_leg} & miss     & 0.660 & 0.667 & 0.668 & 0.660 & 0.662 & 0.666 & 0.626 & 0.677 & 0.579 & \cellcolor{gray!10}0.678 \\
& & wave     & 0.631 & 0.615 & 0.614 & 0.617 & 0.613 & 0.614 & 0.608 & 0.624 & 0.559 & \cellcolor{gray!10}0.674 \\
& & nonlin   & 0.698 & 0.680 & 0.679 & 0.661 & 0.682 & 0.679 & 0.658 & 0.696 & 0.610 & \cellcolor{gray!10}0.712 \\
& & freq     & 0.696 & 0.721 & 0.709 & 0.731 & 0.720 & 0.714 & 0.708 & 0.713 & 0.608 & \cellcolor{gray!10}0.708 \\
& & nonstat  & 0.571 & 0.604 & 0.610 & 0.614 & 0.588 & 0.608 & 0.551 & 0.618 & 0.556 & \cellcolor{gray!10}0.675 \\
\arrayrulecolor{black}\midrule
\multirow{5}{*}{2-mod} & \multirow{5}{*}{arms} & miss     & 0.566 & 0.546 & 0.563 & 0.574 & 0.557 & 0.579 & 0.540 & 0.566 & 0.533 & \cellcolor{gray!10}0.558 \\
& & wave     & 0.555 & 0.586 & 0.582 & 0.583 & 0.568 & 0.569 & 0.557 & 0.588 & 0.524 & \cellcolor{gray!10}0.578 \\
& & nonlin   & 0.627 & 0.626 & 0.644 & 0.634 & 0.638 & 0.635 & 0.641 & 0.632 & 0.581 & \cellcolor{gray!10}0.626 \\
& & freq     & 0.732 & 0.757 & 0.752 & 0.746 & 0.750 & 0.745 & 0.763 & 0.761 & 0.669 & \cellcolor{gray!10}0.744 \\
& & nonstat  & 0.634 & 0.658 & 0.660 & 0.637 & 0.647 & 0.640 & 0.646 & 0.650 & 0.608 & \cellcolor{gray!10}0.662 \\
\arrayrulecolor{black!20}\midrule
& \multirow{5}{*}{legs} & miss     & 0.317 & 0.313 & 0.302 & 0.355 & 0.337 & 0.351 & 0.340 & 0.341 & 0.236 & \cellcolor{gray!10}0.336 \\
& & wave     & 0.281 & 0.267 & 0.279 & 0.271 & 0.285 & 0.260 & 0.283 & 0.278 & 0.222 & \cellcolor{gray!10}0.271 \\
& & nonlin   & 0.556 & 0.519 & 0.529 & 0.517 & 0.528 & 0.514 & 0.518 & 0.540 & 0.302 & \cellcolor{gray!10}0.519 \\
& & freq     & 0.556 & 0.584 & 0.566 & 0.639 & 0.605 & 0.578 & 0.589 & 0.586 & 0.325 & \cellcolor{gray!10}0.594 \\
& & nonstat  & 0.179 & 0.184 & 0.184 & 0.174 & 0.189 & 0.204 & 0.219 & 0.196 & 0.182 & \cellcolor{gray!10}0.194 \\
\arrayrulecolor{black}\midrule
\multirow{5}{*}{3-mod} & \multirow{5}{*}{torso+arms} & miss     & 0.293 & 0.361 & 0.340 & 0.352 & 0.347 & 0.353 & 0.363 & 0.370 & 0.396 & \cellcolor{gray!10}0.370 \\
& & wave     & 0.354 & 0.387 & 0.394 & 0.405 & 0.392 & 0.403 & 0.394 & 0.404 & 0.354 & \cellcolor{gray!10}0.402 \\
& & nonlin   & 0.552 & 0.539 & 0.534 & 0.552 & 0.533 & 0.545 & 0.536 & 0.541 & 0.477 & \cellcolor{gray!10}0.528 \\
& & freq     & 0.709 & 0.728 & 0.735 & 0.724 & 0.729 & 0.756 & 0.721 & 0.732 & 0.640 & \cellcolor{gray!10}0.733 \\
& & nonstat  & 0.533 & 0.560 & 0.550 & 0.538 & 0.517 & 0.546 & 0.489 & 0.570 & 0.474 & \cellcolor{gray!10}0.567 \\
\arrayrulecolor{black!20}\midrule
& \multirow{5}{*}{torso+legs} & miss     & 0.174 & 0.166 & 0.163 & 0.196 & 0.185 & 0.177 & 0.200 & 0.194 & 0.121 & \cellcolor{gray!10}0.190 \\
& & wave     & 0.144 & 0.142 & 0.136 & 0.132 & 0.141 & 0.147 & 0.152 & 0.133 & 0.110 & \cellcolor{gray!10}0.140 \\
& & nonlin   & 0.372 & 0.354 & 0.352 & 0.381 & 0.354 & 0.331 & 0.345 & 0.360 & 0.205 & \cellcolor{gray!10}0.359 \\
& & freq     & 0.553 & 0.562 & 0.567 & 0.613 & 0.581 & 0.551 & 0.558 & 0.554 & 0.309 & \cellcolor{gray!10}0.558 \\
& & nonstat  & 0.095 & 0.090 & 0.084 & 0.091 & 0.092 & 0.096 & 0.096 & 0.092 & 0.088 & \cellcolor{gray!10}0.100 \\
\arrayrulecolor{black}\midrule
\multirow{5}{*}{5-mod} & \multirow{5}{*}{all} & miss     & 0.009 & 0.012 & 0.012 & 0.009 & 0.011 & 0.011 & 0.013 & 0.012 & 0.024 & \cellcolor{gray!10}0.011 \\
& & wave     & 0.028 & 0.017 & 0.024 & 0.026 & 0.024 & 0.024 & 0.027 & 0.024 & 0.017 & \cellcolor{gray!10}0.025 \\
& & nonlin   & 0.115 & 0.085 & 0.087 & 0.123 & 0.101 & 0.094 & 0.141 & 0.115 & 0.074 & \cellcolor{gray!10}0.106 \\
& & freq     & 0.555 & 0.560 & 0.584 & 0.606 & 0.571 & 0.573 & 0.582 & 0.574 & 0.222 & \cellcolor{gray!10}0.568 \\
& & nonstat  & 0.033 & 0.039 & 0.036 & 0.033 & 0.035 & 0.034 & 0.028 & 0.040 & 0.024 & \cellcolor{gray!10}0.035 \\
\arrayrulecolor{black}\bottomrule

\end{tabular}
}
\end{table}

\section{Complete Results on DaliaHAR dataset}\label{app:dalia_complete}

\noindent
For completeness, we provide the full DaliaHAR results behind the aggregate numbers in the main text.
Tables~\ref{tab:appendix-daliahar-maestro-fullgrid},
\ref{tab:appendix-daliahar-shaspec-fullgrid}, and
\ref{tab:appendix-daliahar-flexmoe-fullgrid} show the per-condition F1 scores for MAESTRO, ShaSpec, and Flex-MoE
across corrupted modality subsets and noise types.

\begin{table}[!htbp]
\centering
\caption{Full per-modality-subset, per-noise F1-score ($\uparrow$) on \textbf{DaliaHAR} with the \textbf{MAESTRO} backbone. Each cell is the F1 for one specific (corrupted modality subset, noise type) combination. Noise abbreviations: \emph{miss}=discrete missingness, \emph{wave}=waveform anomaly, \emph{nonlin}=nonlinear degradation, \emph{freq}=frequency attenuation, \emph{nonstat}=nonstationary noise. Clean baseline: F1 = 0.788.}
\label{tab:appendix-daliahar-maestro-fullgrid}
\resizebox{1.0\textwidth}{!}{%
\begin{tabular}{@{}lllcccccccccc@{}}
\toprule
\textbf{Card.} & \textbf{Subset} & \textbf{Noise}
& Static & TENT & EATA & READ & SuMi & SoTTA & AeO & PTA & PSDNorm & \cellcolor{gray!10}\textbf{Ours} \\
\midrule

\multirow{5}{*}{1-mod} & \multirow{5}{*}{wrist\_ACC} & miss     & 0.751 & 0.765 & 0.767 & 0.748 & 0.750 & 0.771 & 0.733 & 0.757 & 0.699 & \cellcolor{gray!10}0.768 \\
& & wave     & 0.756 & 0.785 & 0.781 & 0.760 & 0.756 & 0.781 & 0.735 & 0.755 & 0.714 & \cellcolor{gray!10}0.779 \\
& & nonlin   & 0.756 & 0.775 & 0.777 & 0.762 & 0.757 & 0.774 & 0.736 & 0.765 & 0.722 & \cellcolor{gray!10}0.776 \\
& & freq     & 0.749 & 0.773 & 0.773 & 0.754 & 0.748 & 0.760 & 0.738 & 0.739 & 0.715 & \cellcolor{gray!10}0.774 \\
& & nonstat  & 0.750 & 0.771 & 0.772 & 0.749 & 0.748 & 0.766 & 0.727 & 0.751 & 0.704 & \cellcolor{gray!10}0.771 \\
\arrayrulecolor{black!20}\midrule
& \multirow{5}{*}{wrist\_BVP} & miss     & 0.787 & 0.803 & 0.800 & 0.787 & 0.785 & 0.796 & 0.768 & 0.788 & 0.757 & \cellcolor{gray!10}0.797 \\
& & wave     & 0.785 & 0.802 & 0.802 & 0.783 & 0.784 & 0.796 & 0.768 & 0.787 & 0.758 & \cellcolor{gray!10}0.800 \\
& & nonlin   & 0.752 & 0.761 & 0.759 & 0.758 & 0.748 & 0.764 & 0.735 & 0.767 & 0.722 & \cellcolor{gray!10}0.757 \\
& & freq     & 0.787 & 0.801 & 0.800 & 0.782 & 0.785 & 0.794 & 0.775 & 0.785 & 0.752 & \cellcolor{gray!10}0.801 \\
& & nonstat  & 0.781 & 0.802 & 0.794 & 0.784 & 0.780 & 0.799 & 0.762 & 0.788 & 0.750 & \cellcolor{gray!10}0.794 \\
\arrayrulecolor{black!20}\midrule
& \multirow{5}{*}{chest\_ACC} & miss     & 0.728 & 0.753 & 0.758 & 0.729 & 0.728 & 0.753 & 0.709 & 0.734 & 0.564 & \cellcolor{gray!10}0.750 \\
& & wave     & 0.606 & 0.600 & 0.604 & 0.590 & 0.590 & 0.607 & 0.591 & 0.608 & 0.619 & \cellcolor{gray!10}0.599 \\
& & nonlin   & 0.742 & 0.770 & 0.771 & 0.745 & 0.742 & 0.765 & 0.719 & 0.739 & 0.616 & \cellcolor{gray!10}0.758 \\
& & freq     & 0.735 & 0.754 & 0.752 & 0.729 & 0.733 & 0.740 & 0.717 & 0.732 & 0.651 & \cellcolor{gray!10}0.745 \\
& & nonstat  & 0.717 & 0.733 & 0.742 & 0.714 & 0.711 & 0.737 & 0.693 & 0.722 & 0.617 & \cellcolor{gray!10}0.734 \\
\arrayrulecolor{black!20}\midrule
& \multirow{5}{*}{wrist\_EDA} & miss     & 0.774 & 0.788 & 0.799 & 0.774 & 0.770 & 0.787 & 0.751 & 0.773 & 0.762 & \cellcolor{gray!10}0.794 \\
& & wave     & 0.761 & 0.772 & 0.773 & 0.759 & 0.755 & 0.773 & 0.742 & 0.762 & 0.732 & \cellcolor{gray!10}0.776 \\
& & nonlin   & 0.791 & 0.811 & 0.813 & 0.795 & 0.790 & 0.809 & 0.773 & 0.789 & 0.758 & \cellcolor{gray!10}0.805 \\
& & freq     & 0.789 & 0.805 & 0.807 & 0.791 & 0.785 & 0.802 & 0.768 & 0.793 & 0.758 & \cellcolor{gray!10}0.805 \\
& & nonstat  & 0.784 & 0.805 & 0.804 & 0.781 & 0.780 & 0.803 & 0.766 & 0.778 & 0.758 & \cellcolor{gray!10}0.803 \\
\arrayrulecolor{black!20}\midrule
& \multirow{5}{*}{wrist\_TEMP} & miss     & 0.815 & 0.840 & 0.839 & 0.816 & 0.817 & 0.824 & 0.800 & 0.809 & 0.786 & \cellcolor{gray!10}0.837 \\
& & wave     & 0.821 & 0.847 & 0.844 & 0.820 & 0.821 & 0.834 & 0.799 & 0.818 & 0.793 & \cellcolor{gray!10}0.841 \\
& & nonlin   & 0.798 & 0.815 & 0.816 & 0.795 & 0.794 & 0.813 & 0.772 & 0.798 & 0.769 & \cellcolor{gray!10}0.816 \\
& & freq     & 0.790 & 0.806 & 0.806 & 0.787 & 0.790 & 0.799 & 0.766 & 0.794 & 0.756 & \cellcolor{gray!10}0.804 \\
& & nonstat  & 0.819 & 0.838 & 0.835 & 0.818 & 0.810 & 0.834 & 0.798 & 0.817 & 0.786 & \cellcolor{gray!10}0.836 \\
\arrayrulecolor{black}\midrule
\multirow{5}{*}{2-mod} & \multirow{5}{*}{W\_ACC+\,W\_BVP} & miss     & 0.749 & 0.765 & 0.767 & 0.744 & 0.752 & 0.767 & 0.729 & 0.755 & 0.692 & \cellcolor{gray!10}0.762 \\
& & wave     & 0.761 & 0.784 & 0.788 & 0.757 & 0.759 & 0.776 & 0.735 & 0.751 & 0.713 & \cellcolor{gray!10}0.787 \\
& & nonlin   & 0.715 & 0.735 & 0.724 & 0.724 & 0.725 & 0.731 & 0.697 & 0.730 & 0.689 & \cellcolor{gray!10}0.728 \\
& & freq     & 0.744 & 0.767 & 0.767 & 0.747 & 0.744 & 0.761 & 0.732 & 0.740 & 0.707 & \cellcolor{gray!10}0.769 \\
& & nonstat  & 0.745 & 0.752 & 0.764 & 0.747 & 0.744 & 0.757 & 0.720 & 0.752 & 0.697 & \cellcolor{gray!10}0.761 \\
\arrayrulecolor{black!20}\midrule
& \multirow{5}{*}{C\_ACC+\,W\_ACC} & miss     & 0.660 & 0.681 & 0.686 & 0.655 & 0.654 & 0.684 & 0.632 & 0.656 & 0.442 & \cellcolor{gray!10}0.679 \\
& & wave     & 0.522 & 0.540 & 0.532 & 0.524 & 0.518 & 0.535 & 0.510 & 0.517 & 0.501 & \cellcolor{gray!10}0.530 \\
& & nonlin   & 0.685 & 0.715 & 0.709 & 0.676 & 0.677 & 0.694 & 0.647 & 0.664 & 0.533 & \cellcolor{gray!10}0.706 \\
& & freq     & 0.699 & 0.730 & 0.725 & 0.695 & 0.697 & 0.704 & 0.687 & 0.681 & 0.588 & \cellcolor{gray!10}0.719 \\
& & nonstat  & 0.652 & 0.676 & 0.683 & 0.656 & 0.648 & 0.682 & 0.627 & 0.662 & 0.492 & \cellcolor{gray!10}0.666 \\
\arrayrulecolor{black!20}\midrule
& \multirow{5}{*}{W\_EDA+\,W\_TEMP} & miss     & 0.802 & 0.827 & 0.831 & 0.797 & 0.804 & 0.819 & 0.783 & 0.801 & 0.790 & \cellcolor{gray!10}0.822 \\
& & wave     & 0.793 & 0.809 & 0.807 & 0.798 & 0.795 & 0.804 & 0.770 & 0.806 & 0.766 & \cellcolor{gray!10}0.813 \\
& & nonlin   & 0.800 & 0.819 & 0.817 & 0.800 & 0.799 & 0.812 & 0.782 & 0.795 & 0.769 & \cellcolor{gray!10}0.819 \\
& & freq     & 0.788 & 0.810 & 0.807 & 0.789 & 0.787 & 0.803 & 0.766 & 0.791 & 0.760 & \cellcolor{gray!10}0.802 \\
& & nonstat  & 0.808 & 0.829 & 0.832 & 0.803 & 0.803 & 0.821 & 0.782 & 0.795 & 0.786 & \cellcolor{gray!10}0.832 \\
\arrayrulecolor{black}\midrule
\multirow{5}{*}{3-mod} & \multirow{5}{*}{W\_ACC+\,W\_BVP+\,C\_ACC} & miss     & 0.649 & 0.672 & 0.670 & 0.648 & 0.646 & 0.675 & 0.630 & 0.657 & 0.439 & \cellcolor{gray!10}0.664 \\
& & wave     & 0.517 & 0.525 & 0.537 & 0.518 & 0.521 & 0.526 & 0.507 & 0.524 & 0.491 & \cellcolor{gray!10}0.530 \\
& & nonlin   & 0.636 & 0.652 & 0.653 & 0.628 & 0.631 & 0.634 & 0.619 & 0.624 & 0.474 & \cellcolor{gray!10}0.663 \\
& & freq     & 0.698 & 0.726 & 0.718 & 0.693 & 0.691 & 0.705 & 0.686 & 0.680 & 0.589 & \cellcolor{gray!10}0.719 \\
& & nonstat  & 0.636 & 0.655 & 0.646 & 0.627 & 0.633 & 0.650 & 0.608 & 0.647 & 0.476 & \cellcolor{gray!10}0.644 \\
\arrayrulecolor{black}\midrule
\multirow{5}{*}{4-mod} & \multirow{5}{*}{all\_wrist} & miss     & 0.763 & 0.780 & 0.785 & 0.769 & 0.768 & 0.780 & 0.743 & 0.756 & 0.725 & \cellcolor{gray!10}0.781 \\
& & wave     & 0.758 & 0.783 & 0.781 & 0.768 & 0.763 & 0.784 & 0.738 & 0.764 & 0.727 & \cellcolor{gray!10}0.786 \\
& & nonlin   & 0.732 & 0.750 & 0.742 & 0.732 & 0.732 & 0.749 & 0.710 & 0.741 & 0.702 & \cellcolor{gray!10}0.759 \\
& & freq     & 0.749 & 0.769 & 0.773 & 0.745 & 0.748 & 0.765 & 0.729 & 0.741 & 0.707 & \cellcolor{gray!10}0.771 \\
& & nonstat  & 0.762 & 0.780 & 0.780 & 0.760 & 0.755 & 0.774 & 0.728 & 0.742 & 0.724 & \cellcolor{gray!10}0.775 \\
\arrayrulecolor{black}\bottomrule

\end{tabular}
}
\end{table}

\begin{table}[!htbp]
\centering
\caption{Full per-modality-subset, per-noise F1-score ($\uparrow$) on \textbf{DaliaHAR} with the \textbf{ShaSpec} backbone. Each cell is the F1 for one specific (corrupted modality subset, noise type) combination. Noise abbreviations: \emph{miss}=discrete missingness, \emph{wave}=waveform anomaly, \emph{nonlin}=nonlinear degradation, \emph{freq}=frequency attenuation, \emph{nonstat}=nonstationary noise. Clean baseline: F1 = 0.741.}
\label{tab:appendix-daliahar-shaspec-fullgrid}
\resizebox{1.0\textwidth}{!}{%
\begin{tabular}{@{}lllcccccccccc@{}}
\toprule
\textbf{Card.} & \textbf{Subset} & \textbf{Noise}
& Static & TENT & EATA & READ & SuMi & SoTTA & AeO & PTA & PSDNorm & \cellcolor{gray!10}\textbf{Ours} \\
\midrule

\multirow{5}{*}{1-mod} & \multirow{5}{*}{wrist\_ACC} & miss     & 0.720 & 0.738 & 0.733 & 0.724 & 0.719 & 0.726 & 0.690 & 0.726 & 0.637 & \cellcolor{gray!10}0.731 \\
& & wave     & 0.557 & 0.575 & 0.573 & 0.562 & 0.548 & 0.561 & 0.520 & 0.547 & 0.505 & \cellcolor{gray!10}0.662 \\
& & nonlin   & 0.580 & 0.597 & 0.588 & 0.576 & 0.571 & 0.573 & 0.528 & 0.595 & 0.531 & \cellcolor{gray!10}0.599 \\
& & freq     & 0.586 & 0.597 & 0.593 & 0.585 & 0.588 & 0.586 & 0.566 & 0.593 & 0.566 & \cellcolor{gray!10}0.592 \\
& & nonstat  & 0.603 & 0.627 & 0.623 & 0.624 & 0.611 & 0.618 & 0.587 & 0.617 & 0.585 & \cellcolor{gray!10}0.652 \\
\arrayrulecolor{black!20}\midrule
& \multirow{5}{*}{wrist\_BVP} & miss     & 0.740 & 0.749 & 0.747 & 0.740 & 0.736 & 0.740 & 0.714 & 0.746 & 0.711 & \cellcolor{gray!10}0.749 \\
& & wave     & 0.739 & 0.745 & 0.745 & 0.738 & 0.739 & 0.735 & 0.717 & 0.747 & 0.707 & \cellcolor{gray!10}0.748 \\
& & nonlin   & 0.657 & 0.669 & 0.668 & 0.650 & 0.643 & 0.645 & 0.595 & 0.663 & 0.645 & \cellcolor{gray!10}0.712 \\
& & freq     & 0.742 & 0.750 & 0.750 & 0.737 & 0.737 & 0.742 & 0.716 & 0.744 & 0.711 & \cellcolor{gray!10}0.750 \\
& & nonstat  & 0.741 & 0.753 & 0.748 & 0.737 & 0.735 & 0.741 & 0.716 & 0.744 & 0.710 & \cellcolor{gray!10}0.750 \\
\arrayrulecolor{black!20}\midrule
& \multirow{5}{*}{chest\_ACC} & miss     & 0.501 & 0.500 & 0.494 & 0.491 & 0.493 & 0.501 & 0.488 & 0.499 & 0.578 & \cellcolor{gray!10}0.594 \\
& & wave     & 0.327 & 0.332 & 0.323 & 0.315 & 0.327 & 0.325 & 0.332 & 0.323 & 0.362 & \cellcolor{gray!10}0.465 \\
& & nonlin   & 0.524 & 0.529 & 0.529 & 0.512 & 0.512 & 0.513 & 0.481 & 0.518 & 0.469 & \cellcolor{gray!10}0.589 \\
& & freq     & 0.700 & 0.706 & 0.705 & 0.706 & 0.694 & 0.700 & 0.669 & 0.699 & 0.643 & \cellcolor{gray!10}0.729 \\
& & nonstat  & 0.569 & 0.557 & 0.556 & 0.557 & 0.556 & 0.565 & 0.543 & 0.561 & 0.565 & \cellcolor{gray!10}0.646 \\
\arrayrulecolor{black!20}\midrule
& \multirow{5}{*}{wrist\_EDA} & miss     & 0.712 & 0.716 & 0.714 & 0.713 & 0.698 & 0.702 & 0.672 & 0.716 & 0.703 & \cellcolor{gray!10}0.734 \\
& & wave     & 0.708 & 0.713 & 0.708 & 0.715 & 0.695 & 0.694 & 0.669 & 0.715 & 0.686 & \cellcolor{gray!10}0.732 \\
& & nonlin   & 0.740 & 0.751 & 0.745 & 0.741 & 0.734 & 0.739 & 0.710 & 0.747 & 0.710 & \cellcolor{gray!10}0.748 \\
& & freq     & 0.741 & 0.752 & 0.750 & 0.736 & 0.738 & 0.741 & 0.717 & 0.747 & 0.713 & \cellcolor{gray!10}0.747 \\
& & nonstat  & 0.715 & 0.719 & 0.720 & 0.717 & 0.706 & 0.708 & 0.689 & 0.721 & 0.702 & \cellcolor{gray!10}0.740 \\
\arrayrulecolor{black!20}\midrule
& \multirow{5}{*}{wrist\_TEMP} & miss     & 0.740 & 0.753 & 0.750 & 0.740 & 0.737 & 0.743 & 0.713 & 0.746 & 0.711 & \cellcolor{gray!10}0.748 \\
& & wave     & 0.740 & 0.750 & 0.752 & 0.736 & 0.734 & 0.743 & 0.716 & 0.747 & 0.711 & \cellcolor{gray!10}0.751 \\
& & nonlin   & 0.741 & 0.751 & 0.747 & 0.736 & 0.736 & 0.739 & 0.715 & 0.744 & 0.711 & \cellcolor{gray!10}0.749 \\
& & freq     & 0.741 & 0.753 & 0.748 & 0.735 & 0.735 & 0.742 & 0.710 & 0.746 & 0.711 & \cellcolor{gray!10}0.753 \\
& & nonstat  & 0.740 & 0.754 & 0.749 & 0.737 & 0.737 & 0.745 & 0.713 & 0.746 & 0.711 & \cellcolor{gray!10}0.751 \\
\arrayrulecolor{black}\midrule
\multirow{5}{*}{2-mod} & \multirow{5}{*}{W\_ACC+\,W\_BVP} & miss     & 0.719 & 0.735 & 0.730 & 0.729 & 0.717 & 0.720 & 0.686 & 0.724 & 0.634 & \cellcolor{gray!10}0.730 \\
& & wave     & 0.558 & 0.577 & 0.577 & 0.562 & 0.551 & 0.566 & 0.520 & 0.547 & 0.499 & \cellcolor{gray!10}0.666 \\
& & nonlin   & 0.436 & 0.437 & 0.447 & 0.429 & 0.420 & 0.427 & 0.414 & 0.427 & 0.449 & \cellcolor{gray!10}0.518 \\
& & freq     & 0.584 & 0.599 & 0.590 & 0.592 & 0.591 & 0.586 & 0.568 & 0.595 & 0.563 & \cellcolor{gray!10}0.596 \\
& & nonstat  & 0.600 & 0.620 & 0.616 & 0.615 & 0.606 & 0.614 & 0.584 & 0.610 & 0.583 & \cellcolor{gray!10}0.650 \\
\arrayrulecolor{black!20}\midrule
& \multirow{5}{*}{C\_ACC+\,W\_ACC} & miss     & 0.481 & 0.481 & 0.478 & 0.473 & 0.474 & 0.479 & 0.462 & 0.479 & 0.514 & \cellcolor{gray!10}0.561 \\
& & wave     & 0.209 & 0.208 & 0.209 & 0.209 & 0.209 & 0.207 & 0.211 & 0.210 & 0.173 & \cellcolor{gray!10}0.229 \\
& & nonlin   & 0.298 & 0.298 & 0.302 & 0.302 & 0.299 & 0.300 & 0.286 & 0.304 & 0.247 & \cellcolor{gray!10}0.306 \\
& & freq     & 0.522 & 0.525 & 0.523 & 0.521 & 0.519 & 0.519 & 0.503 & 0.514 & 0.442 & \cellcolor{gray!10}0.503 \\
& & nonstat  & 0.497 & 0.514 & 0.508 & 0.499 & 0.494 & 0.496 & 0.456 & 0.503 & 0.438 & \cellcolor{gray!10}0.527 \\
\arrayrulecolor{black!20}\midrule
& \multirow{5}{*}{W\_EDA+\,W\_TEMP} & miss     & 0.712 & 0.715 & 0.714 & 0.715 & 0.699 & 0.697 & 0.665 & 0.716 & 0.701 & \cellcolor{gray!10}0.736 \\
& & wave     & 0.707 & 0.714 & 0.708 & 0.712 & 0.691 & 0.696 & 0.666 & 0.709 & 0.685 & \cellcolor{gray!10}0.732 \\
& & nonlin   & 0.740 & 0.747 & 0.745 & 0.741 & 0.734 & 0.741 & 0.713 & 0.742 & 0.709 & \cellcolor{gray!10}0.749 \\
& & freq     & 0.741 & 0.753 & 0.752 & 0.739 & 0.735 & 0.739 & 0.716 & 0.748 & 0.712 & \cellcolor{gray!10}0.752 \\
& & nonstat  & 0.715 & 0.721 & 0.721 & 0.716 & 0.708 & 0.708 & 0.688 & 0.720 & 0.700 & \cellcolor{gray!10}0.739 \\
\arrayrulecolor{black}\midrule
\multirow{5}{*}{3-mod} & \multirow{5}{*}{W\_ACC+\,W\_BVP+\,C\_ACC} & miss     & 0.478 & 0.485 & 0.478 & 0.474 & 0.477 & 0.481 & 0.463 & 0.476 & 0.510 & \cellcolor{gray!10}0.548 \\
& & wave     & 0.211 & 0.211 & 0.210 & 0.216 & 0.207 & 0.212 & 0.210 & 0.209 & 0.160 & \cellcolor{gray!10}0.226 \\
& & nonlin   & 0.261 & 0.259 & 0.267 & 0.259 & 0.256 & 0.256 & 0.248 & 0.264 & 0.238 & \cellcolor{gray!10}0.270 \\
& & freq     & 0.520 & 0.525 & 0.519 & 0.519 & 0.517 & 0.518 & 0.502 & 0.520 & 0.442 & \cellcolor{gray!10}0.498 \\
& & nonstat  & 0.491 & 0.504 & 0.504 & 0.493 & 0.486 & 0.487 & 0.450 & 0.498 & 0.433 & \cellcolor{gray!10}0.523 \\
\arrayrulecolor{black}\midrule
\multirow{5}{*}{4-mod} & \multirow{5}{*}{all\_wrist} & miss     & 0.683 & 0.692 & 0.690 & 0.685 & 0.673 & 0.675 & 0.647 & 0.688 & 0.620 & \cellcolor{gray!10}0.709 \\
& & wave     & 0.502 & 0.508 & 0.510 & 0.497 & 0.487 & 0.493 & 0.467 & 0.497 & 0.460 & \cellcolor{gray!10}0.632 \\
& & nonlin   & 0.428 & 0.420 & 0.435 & 0.417 & 0.416 & 0.423 & 0.411 & 0.426 & 0.446 & \cellcolor{gray!10}0.510 \\
& & freq     & 0.585 & 0.602 & 0.589 & 0.595 & 0.586 & 0.588 & 0.566 & 0.591 & 0.564 & \cellcolor{gray!10}0.595 \\
& & nonstat  & 0.563 & 0.584 & 0.577 & 0.581 & 0.570 & 0.571 & 0.544 & 0.573 & 0.566 & \cellcolor{gray!10}0.627 \\
\arrayrulecolor{black}\bottomrule

\end{tabular}
}
\end{table}

\begin{table}[!htbp]
\centering
\caption{Full per-modality-subset, per-noise F1-score ($\uparrow$) on \textbf{DaliaHAR} with the \textbf{Flex-MoE} backbone. Each cell is the F1 for one specific (corrupted modality subset, noise type) combination. Noise abbreviations: \emph{miss}=discrete missingness, \emph{wave}=waveform anomaly, \emph{nonlin}=nonlinear degradation, \emph{freq}=frequency attenuation, \emph{nonstat}=nonstationary noise. Clean baseline: F1 = 0.768.}
\label{tab:appendix-daliahar-flexmoe-fullgrid}
\resizebox{1.0\textwidth}{!}{%
\begin{tabular}{@{}lllcccccccccc@{}}
\toprule
\textbf{Card.} & \textbf{Subset} & \textbf{Noise}
& Static & TENT & EATA & READ & SuMi & SoTTA & AeO & PTA & PSDNorm & \cellcolor{gray!10}\textbf{Ours} \\
\midrule

\multirow{5}{*}{1-mod} & \multirow{5}{*}{wrist\_ACC} & miss     & 0.743 & 0.787 & 0.784 & 0.748 & 0.758 & 0.780 & 0.726 & 0.711 & 0.693 & \cellcolor{gray!10}0.788 \\
& & wave     & 0.622 & 0.653 & 0.656 & 0.608 & 0.637 & 0.680 & 0.590 & 0.552 & 0.561 & \cellcolor{gray!10}0.654 \\
& & nonlin   & 0.624 & 0.644 & 0.650 & 0.608 & 0.659 & 0.698 & 0.627 & 0.595 & 0.608 & \cellcolor{gray!10}0.650 \\
& & freq     & 0.675 & 0.736 & 0.738 & 0.692 & 0.676 & 0.710 & 0.631 & 0.613 & 0.612 & \cellcolor{gray!10}0.729 \\
& & nonstat  & 0.586 & 0.631 & 0.628 & 0.604 & 0.611 & 0.639 & 0.573 & 0.554 & 0.571 & \cellcolor{gray!10}0.625 \\
\arrayrulecolor{black!20}\midrule
& \multirow{5}{*}{wrist\_BVP} & miss     & 0.767 & 0.801 & 0.809 & 0.786 & 0.787 & 0.801 & 0.752 & 0.738 & 0.745 & \cellcolor{gray!10}0.808 \\
& & wave     & 0.767 & 0.801 & 0.806 & 0.782 & 0.780 & 0.795 & 0.759 & 0.747 & 0.743 & \cellcolor{gray!10}0.801 \\
& & nonlin   & 0.699 & 0.750 & 0.753 & 0.726 & 0.678 & 0.697 & 0.651 & 0.634 & 0.687 & \cellcolor{gray!10}0.757 \\
& & freq     & 0.770 & 0.807 & 0.811 & 0.788 & 0.774 & 0.788 & 0.755 & 0.744 & 0.744 & \cellcolor{gray!10}0.800 \\
& & nonstat  & 0.770 & 0.807 & 0.807 & 0.781 & 0.782 & 0.799 & 0.760 & 0.742 & 0.744 & \cellcolor{gray!10}0.798 \\
\arrayrulecolor{black!20}\midrule
& \multirow{5}{*}{chest\_ACC} & miss     & 0.644 & 0.698 & 0.703 & 0.670 & 0.633 & 0.638 & 0.599 & 0.625 & 0.638 & \cellcolor{gray!10}0.701 \\
& & wave     & 0.510 & 0.501 & 0.508 & 0.498 & 0.473 & 0.468 & 0.461 & 0.462 & 0.495 & \cellcolor{gray!10}0.508 \\
& & nonlin   & 0.549 & 0.571 & 0.572 & 0.537 & 0.560 & 0.592 & 0.519 & 0.510 & 0.482 & \cellcolor{gray!10}0.571 \\
& & freq     & 0.731 & 0.769 & 0.774 & 0.755 & 0.733 & 0.755 & 0.706 & 0.701 & 0.687 & \cellcolor{gray!10}0.769 \\
& & nonstat  & 0.665 & 0.708 & 0.718 & 0.686 & 0.636 & 0.638 & 0.607 & 0.633 & 0.636 & \cellcolor{gray!10}0.705 \\
\arrayrulecolor{black!20}\midrule
& \multirow{5}{*}{wrist\_EDA} & miss     & 0.707 & 0.760 & 0.760 & 0.746 & 0.713 & 0.736 & 0.696 & 0.702 & 0.742 & \cellcolor{gray!10}0.753 \\
& & wave     & 0.716 & 0.772 & 0.769 & 0.754 & 0.735 & 0.763 & 0.715 & 0.693 & 0.717 & \cellcolor{gray!10}0.764 \\
& & nonlin   & 0.759 & 0.804 & 0.797 & 0.775 & 0.778 & 0.803 & 0.744 & 0.730 & 0.735 & \cellcolor{gray!10}0.804 \\
& & freq     & 0.768 & 0.812 & 0.807 & 0.793 & 0.787 & 0.812 & 0.756 & 0.738 & 0.743 & \cellcolor{gray!10}0.803 \\
& & nonstat  & 0.731 & 0.788 & 0.784 & 0.748 & 0.762 & 0.779 & 0.725 & 0.705 & 0.734 & \cellcolor{gray!10}0.772 \\
\arrayrulecolor{black!20}\midrule
& \multirow{5}{*}{wrist\_TEMP} & miss     & 0.766 & 0.792 & 0.794 & 0.789 & 0.774 & 0.793 & 0.734 & 0.720 & 0.737 & \cellcolor{gray!10}0.792 \\
& & wave     & 0.763 & 0.790 & 0.795 & 0.778 & 0.783 & 0.807 & 0.758 & 0.737 & 0.749 & \cellcolor{gray!10}0.798 \\
& & nonlin   & 0.766 & 0.810 & 0.801 & 0.775 & 0.783 & 0.802 & 0.762 & 0.732 & 0.742 & \cellcolor{gray!10}0.808 \\
& & freq     & 0.767 & 0.805 & 0.805 & 0.793 & 0.783 & 0.802 & 0.756 & 0.737 & 0.744 & \cellcolor{gray!10}0.803 \\
& & nonstat  & 0.753 & 0.792 & 0.799 & 0.797 & 0.768 & 0.784 & 0.720 & 0.713 & 0.733 & \cellcolor{gray!10}0.798 \\
\arrayrulecolor{black}\midrule

\multirow{5}{*}{2-mod} & \multirow{5}{*}{W\_ACC+\,W\_BVP} & miss     & 0.743 & 0.783 & 0.788 & 0.766 & 0.753 & 0.776 & 0.722 & 0.709 & 0.692 & \cellcolor{gray!10}0.784 \\
& & wave     & 0.619 & 0.646 & 0.649 & 0.614 & 0.610 & 0.639 & 0.574 & 0.545 & 0.560 & \cellcolor{gray!10}0.653 \\
& & nonlin   & 0.496 & 0.518 & 0.520 & 0.529 & 0.422 & 0.431 & 0.428 & 0.394 & 0.504 & \cellcolor{gray!10}0.519 \\
& & freq     & 0.673 & 0.730 & 0.735 & 0.695 & 0.676 & 0.705 & 0.640 & 0.621 & 0.609 & \cellcolor{gray!10}0.732 \\
& & nonstat  & 0.579 & 0.626 & 0.619 & 0.599 & 0.605 & 0.617 & 0.568 & 0.568 & 0.565 & \cellcolor{gray!10}0.611 \\
\arrayrulecolor{black!20}\midrule
& \multirow{5}{*}{C\_ACC+\,W\_ACC} & miss     & 0.599 & 0.644 & 0.656 & 0.635 & 0.575 & 0.574 & 0.553 & 0.560 & 0.554 & \cellcolor{gray!10}0.651 \\
& & wave     & 0.300 & 0.296 & 0.293 & 0.269 & 0.279 & 0.284 & 0.289 & 0.265 & 0.268 & \cellcolor{gray!10}0.298 \\
& & nonlin   & 0.277 & 0.270 & 0.272 & 0.261 & 0.255 & 0.267 & 0.260 & 0.238 & 0.280 & \cellcolor{gray!10}0.273 \\
& & freq     & 0.633 & 0.699 & 0.695 & 0.663 & 0.632 & 0.655 & 0.583 & 0.585 & 0.499 & \cellcolor{gray!10}0.699 \\
& & nonstat  & 0.468 & 0.515 & 0.518 & 0.471 & 0.443 & 0.450 & 0.413 & 0.390 & 0.421 & \cellcolor{gray!10}0.518 \\
\arrayrulecolor{black!20}\midrule
& \multirow{5}{*}{W\_EDA+\,W\_TEMP} & miss     & 0.688 & 0.736 & 0.743 & 0.707 & 0.712 & 0.730 & 0.693 & 0.688 & 0.735 & \cellcolor{gray!10}0.741 \\
& & wave     & 0.695 & 0.754 & 0.747 & 0.732 & 0.701 & 0.732 & 0.681 & 0.678 & 0.710 & \cellcolor{gray!10}0.751 \\
& & nonlin   & 0.757 & 0.793 & 0.798 & 0.757 & 0.771 & 0.788 & 0.743 & 0.724 & 0.728 & \cellcolor{gray!10}0.797 \\
& & freq     & 0.768 & 0.801 & 0.803 & 0.779 & 0.783 & 0.806 & 0.757 & 0.746 & 0.744 & \cellcolor{gray!10}0.799 \\
& & nonstat  & 0.710 & 0.774 & 0.775 & 0.745 & 0.716 & 0.749 & 0.687 & 0.656 & 0.726 & \cellcolor{gray!10}0.762 \\
\arrayrulecolor{black}\midrule

\multirow{5}{*}{3-mod} & \multirow{5}{*}{W\_ACC+\,W\_BVP+\,C\_ACC} & miss     & 0.592 & 0.644 & 0.638 & 0.622 & 0.589 & 0.604 & 0.560 & 0.562 & 0.547 & \cellcolor{gray!10}0.642 \\
& & wave     & 0.305 & 0.298 & 0.296 & 0.274 & 0.261 & 0.251 & 0.264 & 0.250 & 0.261 & \cellcolor{gray!10}0.295 \\
& & nonlin   & 0.224 & 0.204 & 0.203 & 0.206 & 0.185 & 0.183 & 0.200 & 0.199 & 0.250 & \cellcolor{gray!10}0.214 \\
& & freq     & 0.632 & 0.699 & 0.698 & 0.646 & 0.635 & 0.666 & 0.582 & 0.587 & 0.498 & \cellcolor{gray!10}0.692 \\
& & nonstat  & 0.464 & 0.515 & 0.511 & 0.468 & 0.453 & 0.457 & 0.418 & 0.414 & 0.417 & \cellcolor{gray!10}0.514 \\
\arrayrulecolor{black}\midrule

\multirow{5}{*}{4-mod} & \multirow{5}{*}{all\_wrist} & miss     & 0.686 & 0.749 & 0.748 & 0.730 & 0.739 & 0.756 & 0.696 & 0.691 & 0.676 & \cellcolor{gray!10}0.747 \\
& & wave     & 0.612 & 0.644 & 0.636 & 0.609 & 0.576 & 0.597 & 0.549 & 0.533 & 0.577 & \cellcolor{gray!10}0.650 \\
& & nonlin   & 0.463 & 0.486 & 0.481 & 0.493 & 0.407 & 0.414 & 0.414 & 0.385 & 0.457 & \cellcolor{gray!10}0.483 \\
& & freq     & 0.671 & 0.743 & 0.747 & 0.689 & 0.688 & 0.713 & 0.642 & 0.628 & 0.608 & \cellcolor{gray!10}0.731 \\
& & nonstat  & 0.545 & 0.600 & 0.596 & 0.572 & 0.519 & 0.554 & 0.476 & 0.479 & 0.540 & \cellcolor{gray!10}0.589 \\
\arrayrulecolor{black}\bottomrule

\end{tabular}
}
\end{table}

\section{Complete Results on PAMAP2 dataset}\label{app:pamap_complete}

\noindent
For completeness, we provide the full PAMAP2 results behind the aggregate numbers in the main text.
Tables~\ref{tab:appendix-pamap2-maestro-fullgrid},
\ref{tab:appendix-pamap2-shaspec-fullgrid}, and
\ref{tab:appendix-pamap2-flexmoe-fullgrid} show the per-condition F1 scores for MAESTRO, ShaSpec, and Flex-MoE
across corrupted modality subsets and noise types.

\begin{table}[!htbp]
\centering
\caption{Full per-modality-subset, per-noise F1-score ($\uparrow$) on \textbf{PAMAP2} with the \textbf{MAESTRO} backbone. Each cell is the F1 for one specific (corrupted modality subset, noise type) combination. Noise abbreviations: \emph{miss}=discrete missingness, \emph{wave}=waveform anomaly, \emph{nonlin}=nonlinear degradation, \emph{freq}=frequency attenuation, \emph{nonstat}=nonstationary noise. Clean baseline: F1 = 0.951.}
\label{tab:appendix-pamap2-maestro-fullgrid}
\resizebox{1.0\textwidth}{!}{%
\begin{tabular}{@{}lllcccccccccc@{}}
\toprule
\textbf{Card.} & \textbf{Subset} & \textbf{Noise}
& Static & TENT & EATA & READ & SuMi & SoTTA & AeO & PTA & PSDNorm & \cellcolor{gray!10}\textbf{Ours} \\
\midrule

\multirow{5}{*}{1-mod} & \multirow{5}{*}{IMU\_ankle} & miss     & 0.940 & 0.938 & 0.946 & 0.943 & 0.935 & 0.940 & 0.937 & 0.941 & 0.901 & \cellcolor{gray!10}0.935 \\
& & wave     & 0.926 & 0.931 & 0.932 & 0.926 & 0.927 & 0.929 & 0.924 & 0.931 & 0.897 & \cellcolor{gray!10}0.925 \\
& & nonlin   & 0.940 & 0.933 & 0.938 & 0.932 & 0.941 & 0.935 & 0.938 & 0.943 & 0.903 & \cellcolor{gray!10}0.939 \\
& & freq     & 0.941 & 0.942 & 0.942 & 0.936 & 0.938 & 0.938 & 0.939 & 0.939 & 0.918 & \cellcolor{gray!10}0.943 \\
& & nonstat  & 0.943 & 0.938 & 0.943 & 0.936 & 0.932 & 0.937 & 0.937 & 0.942 & 0.892 & \cellcolor{gray!10}0.932 \\
\arrayrulecolor{black!20}\midrule
& \multirow{5}{*}{IMU\_hand} & miss     & 0.944 & 0.941 & 0.945 & 0.933 & 0.937 & 0.944 & 0.939 & 0.945 & 0.910 & \cellcolor{gray!10}0.946 \\
& & wave     & 0.945 & 0.948 & 0.946 & 0.939 & 0.939 & 0.943 & 0.944 & 0.943 & 0.926 & \cellcolor{gray!10}0.948 \\
& & nonlin   & 0.949 & 0.943 & 0.948 & 0.943 & 0.949 & 0.947 & 0.949 & 0.947 & 0.914 & \cellcolor{gray!10}0.947 \\
& & freq     & 0.937 & 0.936 & 0.939 & 0.941 & 0.943 & 0.941 & 0.944 & 0.934 & 0.929 & \cellcolor{gray!10}0.937 \\
& & nonstat  & 0.943 & 0.937 & 0.944 & 0.937 & 0.933 & 0.941 & 0.937 & 0.938 & 0.923 & \cellcolor{gray!10}0.935 \\
\arrayrulecolor{black!20}\midrule
& \multirow{5}{*}{IMU\_chest} & miss     & 0.943 & 0.940 & 0.943 & 0.933 & 0.941 & 0.940 & 0.938 & 0.942 & 0.906 & \cellcolor{gray!10}0.941 \\
& & wave     & 0.915 & 0.902 & 0.907 & 0.906 & 0.907 & 0.901 & 0.906 & 0.909 & 0.895 & \cellcolor{gray!10}0.921 \\
& & nonlin   & 0.946 & 0.947 & 0.941 & 0.939 & 0.945 & 0.942 & 0.945 & 0.936 & 0.927 & \cellcolor{gray!10}0.945 \\
& & freq     & 0.944 & 0.944 & 0.944 & 0.945 & 0.945 & 0.943 & 0.944 & 0.951 & 0.935 & \cellcolor{gray!10}0.943 \\
& & nonstat  & 0.935 & 0.936 & 0.934 & 0.927 & 0.926 & 0.935 & 0.929 & 0.931 & 0.913 & \cellcolor{gray!10}0.926 \\
\arrayrulecolor{black!20}\midrule
& \multirow{5}{*}{heart\_rate} & miss     & 0.950 & 0.948 & 0.950 & 0.947 & 0.949 & 0.953 & 0.947 & 0.947 & 0.947 & \cellcolor{gray!10}0.945 \\
& & wave     & 0.952 & 0.948 & 0.950 & 0.948 & 0.948 & 0.949 & 0.946 & 0.950 & 0.947 & \cellcolor{gray!10}0.943 \\
& & nonlin   & 0.951 & 0.951 & 0.951 & 0.948 & 0.948 & 0.947 & 0.948 & 0.947 & 0.947 & \cellcolor{gray!10}0.948 \\
& & freq     & 0.953 & 0.946 & 0.949 & 0.949 & 0.946 & 0.946 & 0.949 & 0.950 & 0.946 & \cellcolor{gray!10}0.949 \\
& & nonstat  & 0.950 & 0.947 & 0.949 & 0.944 & 0.946 & 0.946 & 0.948 & 0.945 & 0.944 & \cellcolor{gray!10}0.949 \\
\arrayrulecolor{black}\midrule
\multirow{5}{*}{2-mod} & \multirow{5}{*}{ankle+\,hand} & miss     & 0.933 & 0.933 & 0.932 & 0.920 & 0.927 & 0.926 & 0.923 & 0.932 & 0.795 & \cellcolor{gray!10}0.930 \\
& & wave     & 0.916 & 0.915 & 0.920 & 0.908 & 0.912 & 0.917 & 0.907 & 0.916 & 0.827 & \cellcolor{gray!10}0.927 \\
& & nonlin   & 0.928 & 0.921 & 0.926 & 0.925 & 0.924 & 0.926 & 0.921 & 0.932 & 0.805 & \cellcolor{gray!10}0.927 \\
& & freq     & 0.889 & 0.911 & 0.928 & 0.902 & 0.905 & 0.905 & 0.891 & 0.918 & 0.860 & \cellcolor{gray!10}0.923 \\
& & nonstat  & 0.922 & 0.915 & 0.922 & 0.911 & 0.903 & 0.920 & 0.911 & 0.921 & 0.807 & \cellcolor{gray!10}0.923 \\
\arrayrulecolor{black!20}\midrule
& \multirow{5}{*}{ankle+\,HR} & miss     & 0.941 & 0.944 & 0.944 & 0.935 & 0.939 & 0.934 & 0.940 & 0.940 & 0.901 & \cellcolor{gray!10}0.936 \\
& & wave     & 0.931 & 0.933 & 0.935 & 0.934 & 0.930 & 0.931 & 0.926 & 0.931 & 0.901 & \cellcolor{gray!10}0.931 \\
& & nonlin   & 0.943 & 0.943 & 0.943 & 0.932 & 0.938 & 0.943 & 0.938 & 0.938 & 0.898 & \cellcolor{gray!10}0.939 \\
& & freq     & 0.937 & 0.945 & 0.944 & 0.940 & 0.936 & 0.941 & 0.938 & 0.942 & 0.918 & \cellcolor{gray!10}0.941 \\
& & nonstat  & 0.938 & 0.940 & 0.940 & 0.934 & 0.938 & 0.936 & 0.936 & 0.939 & 0.901 & \cellcolor{gray!10}0.936 \\
\arrayrulecolor{black}\midrule
\multirow{5}{*}{3-mod} & \multirow{5}{*}{ankle+\,hand+\,chest} & miss     & 0.896 & 0.889 & 0.893 & 0.883 & 0.884 & 0.889 & 0.881 & 0.895 & 0.698 & \cellcolor{gray!10}0.880 \\
& & wave     & 0.803 & 0.805 & 0.801 & 0.800 & 0.791 & 0.797 & 0.774 & 0.807 & 0.690 & \cellcolor{gray!10}0.791 \\
& & nonlin   & 0.901 & 0.889 & 0.904 & 0.885 & 0.885 & 0.896 & 0.887 & 0.899 & 0.671 & \cellcolor{gray!10}0.894 \\
& & freq     & 0.832 & 0.855 & 0.858 & 0.838 & 0.835 & 0.857 & 0.829 & 0.850 & 0.793 & \cellcolor{gray!10}0.855 \\
& & nonstat  & 0.848 & 0.834 & 0.839 & 0.848 & 0.837 & 0.847 & 0.842 & 0.850 & 0.695 & \cellcolor{gray!10}0.841 \\
\arrayrulecolor{black!20}\midrule
& \multirow{5}{*}{ankle+\,hand+\,HR} & miss     & 0.931 & 0.927 & 0.931 & 0.919 & 0.921 & 0.932 & 0.924 & 0.931 & 0.791 & \cellcolor{gray!10}0.936 \\
& & wave     & 0.915 & 0.918 & 0.924 & 0.911 & 0.918 & 0.917 & 0.914 & 0.923 & 0.834 & \cellcolor{gray!10}0.919 \\
& & nonlin   & 0.933 & 0.925 & 0.924 & 0.922 & 0.916 & 0.923 & 0.918 & 0.929 & 0.799 & \cellcolor{gray!10}0.932 \\
& & freq     & 0.892 & 0.921 & 0.911 & 0.903 & 0.901 & 0.906 & 0.901 & 0.914 & 0.857 & \cellcolor{gray!10}0.925 \\
& & nonstat  & 0.925 & 0.924 & 0.922 & 0.914 & 0.917 & 0.919 & 0.909 & 0.925 & 0.813 & \cellcolor{gray!10}0.920 \\
\arrayrulecolor{black}\midrule
\multirow{5}{*}{4-mod} & \multirow{5}{*}{all} & miss     & 0.885 & 0.884 & 0.883 & 0.883 & 0.878 & 0.888 & 0.882 & 0.882 & 0.696 & \cellcolor{gray!10}0.890 \\
& & wave     & 0.798 & 0.803 & 0.803 & 0.795 & 0.806 & 0.797 & 0.776 & 0.803 & 0.692 & \cellcolor{gray!10}0.788 \\
& & nonlin   & 0.892 & 0.887 & 0.895 & 0.880 & 0.887 & 0.889 & 0.886 & 0.906 & 0.675 & \cellcolor{gray!10}0.890 \\
& & freq     & 0.830 & 0.858 & 0.852 & 0.845 & 0.833 & 0.853 & 0.829 & 0.864 & 0.788 & \cellcolor{gray!10}0.847 \\
& & nonstat  & 0.845 & 0.837 & 0.833 & 0.838 & 0.839 & 0.835 & 0.843 & 0.848 & 0.704 & \cellcolor{gray!10}0.840 \\
\arrayrulecolor{black}\bottomrule

\end{tabular}
}
\end{table}

\begin{table}[!htbp]
\centering
\caption{Full per-modality-subset, per-noise F1-score ($\uparrow$) on \textbf{PAMAP2} with the \textbf{ShaSpec} backbone. Each cell is the F1 for one specific (corrupted modality subset, noise type) combination. Noise abbreviations: \emph{miss}=discrete missingness, \emph{wave}=waveform anomaly, \emph{nonlin}=nonlinear degradation, \emph{freq}=frequency attenuation, \emph{nonstat}=nonstationary noise. Clean baseline: F1 = 0.862.}
\label{tab:appendix-pamap2-shaspec-fullgrid}
\resizebox{1.0\textwidth}{!}{%
\begin{tabular}{@{}lllcccccccccc@{}}
\toprule
\textbf{Card.} & \textbf{Subset} & \textbf{Noise}
& Static & TENT & EATA & READ & SuMi & SoTTA & AeO & PTA & PSDNorm & \cellcolor{gray!10}\textbf{Ours} \\
\midrule

\multirow{5}{*}{1-mod} & \multirow{5}{*}{IMU\_ankle} & miss     & 0.819 & 0.812 & 0.808 & 0.763 & 0.802 & 0.811 & 0.803 & 0.803 & 0.804 & \cellcolor{gray!10}0.824 \\
& & wave     & 0.779 & 0.764 & 0.764 & 0.725 & 0.770 & 0.759 & 0.757 & 0.767 & 0.784 & \cellcolor{gray!10}0.798 \\
& & nonlin   & 0.270 & 0.223 & 0.239 & 0.214 & 0.267 & 0.242 & 0.289 & 0.265 & 0.292 & \cellcolor{gray!10}0.511 \\
& & freq     & 0.860 & 0.850 & 0.845 & 0.807 & 0.851 & 0.843 & 0.833 & 0.843 & 0.849 & \cellcolor{gray!10}0.847 \\
& & nonstat  & 0.850 & 0.834 & 0.838 & 0.794 & 0.839 & 0.841 & 0.828 & 0.843 & 0.833 & \cellcolor{gray!10}0.822 \\
\arrayrulecolor{black!20}\midrule
& \multirow{5}{*}{IMU\_hand} & miss     & 0.839 & 0.829 & 0.823 & 0.816 & 0.833 & 0.826 & 0.825 & 0.830 & 0.837 & \cellcolor{gray!10}0.845 \\
& & wave     & 0.813 & 0.798 & 0.802 & 0.781 & 0.806 & 0.791 & 0.794 & 0.801 & 0.800 & \cellcolor{gray!10}0.827 \\
& & nonlin   & 0.724 & 0.712 & 0.705 & 0.691 & 0.709 & 0.714 & 0.699 & 0.725 & 0.772 & \cellcolor{gray!10}0.798 \\
& & freq     & 0.860 & 0.851 & 0.843 & 0.816 & 0.857 & 0.847 & 0.843 & 0.847 & 0.851 & \cellcolor{gray!10}0.845 \\
& & nonstat  & 0.857 & 0.843 & 0.848 & 0.817 & 0.846 & 0.842 & 0.837 & 0.836 & 0.841 & \cellcolor{gray!10}0.848 \\
\arrayrulecolor{black!20}\midrule
& \multirow{5}{*}{IMU\_chest} & miss     & 0.784 & 0.788 & 0.786 & 0.759 & 0.781 & 0.780 & 0.766 & 0.790 & 0.798 & \cellcolor{gray!10}0.794 \\
& & wave     & 0.700 & 0.693 & 0.688 & 0.701 & 0.695 & 0.705 & 0.691 & 0.701 & 0.690 & \cellcolor{gray!10}0.746 \\
& & nonlin   & 0.432 & 0.414 & 0.429 & 0.434 & 0.429 & 0.443 & 0.436 & 0.437 & 0.484 & \cellcolor{gray!10}0.568 \\
& & freq     & 0.860 & 0.854 & 0.849 & 0.820 & 0.852 & 0.849 & 0.847 & 0.847 & 0.851 & \cellcolor{gray!10}0.829 \\
& & nonstat  & 0.800 & 0.796 & 0.798 & 0.763 & 0.792 & 0.796 & 0.788 & 0.802 & 0.828 & \cellcolor{gray!10}0.783 \\
\arrayrulecolor{black!20}\midrule
& \multirow{5}{*}{heart\_rate} & miss     & 0.857 & 0.845 & 0.847 & 0.819 & 0.849 & 0.850 & 0.850 & 0.856 & 0.850 & \cellcolor{gray!10}0.852 \\
& & wave     & 0.861 & 0.849 & 0.849 & 0.817 & 0.849 & 0.839 & 0.848 & 0.849 & 0.852 & \cellcolor{gray!10}0.851 \\
& & nonlin   & 0.863 & 0.851 & 0.850 & 0.820 & 0.853 & 0.851 & 0.852 & 0.848 & 0.856 & \cellcolor{gray!10}0.859 \\
& & freq     & 0.864 & 0.853 & 0.846 & 0.826 & 0.854 & 0.847 & 0.843 & 0.856 & 0.859 & \cellcolor{gray!10}0.857 \\
& & nonstat  & 0.856 & 0.857 & 0.849 & 0.814 & 0.850 & 0.846 & 0.845 & 0.854 & 0.853 & \cellcolor{gray!10}0.853 \\
\arrayrulecolor{black}\midrule
\multirow{5}{*}{2-mod} & \multirow{5}{*}{ankle+\,hand} & miss     & 0.810 & 0.807 & 0.816 & 0.743 & 0.803 & 0.800 & 0.791 & 0.816 & 0.771 & \cellcolor{gray!10}0.799 \\
& & wave     & 0.743 & 0.726 & 0.708 & 0.673 & 0.724 & 0.717 & 0.717 & 0.740 & 0.725 & \cellcolor{gray!10}0.748 \\
& & nonlin   & 0.186 & 0.159 & 0.145 & 0.138 & 0.174 & 0.164 & 0.215 & 0.163 & 0.224 & \cellcolor{gray!10}0.321 \\
& & freq     & 0.852 & 0.845 & 0.841 & 0.813 & 0.840 & 0.839 & 0.844 & 0.835 & 0.834 & \cellcolor{gray!10}0.839 \\
& & nonstat  & 0.846 & 0.826 & 0.834 & 0.772 & 0.836 & 0.832 & 0.826 & 0.841 & 0.821 & \cellcolor{gray!10}0.819 \\
\arrayrulecolor{black!20}\midrule
& \multirow{5}{*}{ankle+\,HR} & miss     & 0.807 & 0.800 & 0.810 & 0.749 & 0.788 & 0.783 & 0.782 & 0.807 & 0.803 & \cellcolor{gray!10}0.822 \\
& & wave     & 0.781 & 0.772 & 0.774 & 0.713 & 0.771 & 0.776 & 0.752 & 0.770 & 0.783 & \cellcolor{gray!10}0.801 \\
& & nonlin   & 0.264 & 0.220 & 0.230 & 0.209 & 0.259 & 0.236 & 0.274 & 0.282 & 0.294 & \cellcolor{gray!10}0.517 \\
& & freq     & 0.859 & 0.849 & 0.846 & 0.807 & 0.850 & 0.846 & 0.838 & 0.844 & 0.852 & \cellcolor{gray!10}0.849 \\
& & nonstat  & 0.846 & 0.839 & 0.841 & 0.790 & 0.849 & 0.846 & 0.831 & 0.833 & 0.842 & \cellcolor{gray!10}0.825 \\
\arrayrulecolor{black}\midrule
\multirow{5}{*}{3-mod} & \multirow{5}{*}{ankle+\,hand+\,chest} & miss     & 0.686 & 0.678 & 0.671 & 0.634 & 0.662 & 0.681 & 0.674 & 0.678 & 0.706 & \cellcolor{gray!10}0.663 \\
& & wave     & 0.565 & 0.554 & 0.553 & 0.546 & 0.559 & 0.569 & 0.555 & 0.567 & 0.577 & \cellcolor{gray!10}0.540 \\
& & nonlin   & 0.060 & 0.056 & 0.055 & 0.051 & 0.064 & 0.064 & 0.069 & 0.062 & 0.121 & \cellcolor{gray!10}0.063 \\
& & freq     & 0.838 & 0.829 & 0.827 & 0.791 & 0.839 & 0.832 & 0.838 & 0.838 & 0.819 & \cellcolor{gray!10}0.834 \\
& & nonstat  & 0.741 & 0.735 & 0.739 & 0.683 & 0.716 & 0.758 & 0.715 & 0.751 & 0.770 & \cellcolor{gray!10}0.724 \\
\arrayrulecolor{black!20}\midrule
& \multirow{5}{*}{ankle+\,hand+\,HR} & miss     & 0.804 & 0.798 & 0.789 & 0.728 & 0.788 & 0.790 & 0.766 & 0.807 & 0.768 & \cellcolor{gray!10}0.809 \\
& & wave     & 0.710 & 0.700 & 0.689 & 0.662 & 0.698 & 0.701 & 0.693 & 0.704 & 0.721 & \cellcolor{gray!10}0.746 \\
& & nonlin   & 0.183 & 0.145 & 0.147 & 0.132 & 0.178 & 0.160 & 0.208 & 0.162 & 0.223 & \cellcolor{gray!10}0.308 \\
& & freq     & 0.853 & 0.844 & 0.847 & 0.808 & 0.842 & 0.846 & 0.838 & 0.839 & 0.836 & \cellcolor{gray!10}0.847 \\
& & nonstat  & 0.848 & 0.839 & 0.837 & 0.773 & 0.832 & 0.825 & 0.824 & 0.840 & 0.830 & \cellcolor{gray!10}0.815 \\
\arrayrulecolor{black}\midrule
\multirow{5}{*}{4-mod} & \multirow{5}{*}{all} & miss     & 0.682 & 0.667 & 0.683 & 0.627 & 0.667 & 0.666 & 0.652 & 0.682 & 0.690 & \cellcolor{gray!10}0.669 \\
& & wave     & 0.539 & 0.530 & 0.523 & 0.510 & 0.532 & 0.529 & 0.531 & 0.532 & 0.563 & \cellcolor{gray!10}0.529 \\
& & nonlin   & 0.059 & 0.053 & 0.065 & 0.048 & 0.063 & 0.062 & 0.070 & 0.060 & 0.119 & \cellcolor{gray!10}0.056 \\
& & freq     & 0.838 & 0.834 & 0.835 & 0.801 & 0.834 & 0.834 & 0.844 & 0.839 & 0.821 & \cellcolor{gray!10}0.846 \\
& & nonstat  & 0.744 & 0.733 & 0.734 & 0.700 & 0.727 & 0.752 & 0.730 & 0.756 & 0.775 & \cellcolor{gray!10}0.728 \\
\arrayrulecolor{black}\bottomrule

\end{tabular}
}
\end{table}

\begin{table}[!htbp]
\centering
\caption{Full per-modality-subset, per-noise F1-score ($\uparrow$) on \textbf{PAMAP2} with the \textbf{Flex-MoE} backbone. Each cell is the F1 for one specific (corrupted modality subset, noise type) combination. Noise abbreviations: \emph{miss}=discrete missingness, \emph{wave}=waveform anomaly, \emph{nonlin}=nonlinear degradation, \emph{freq}=frequency attenuation, \emph{nonstat}=nonstationary noise. Clean baseline: F1 = 0.893.}
\label{tab:appendix-pamap2-flexmoe-fullgrid}
\resizebox{1.0\textwidth}{!}{%
\begin{tabular}{@{}lllcccccccccc@{}}
\toprule
\textbf{Card.} & \textbf{Subset} & \textbf{Noise}
& Static & TENT & EATA & READ & SuMi & SoTTA & AeO & PTA & PSDNorm & \cellcolor{gray!10}\textbf{Ours} \\
\midrule

\multirow{5}{*}{1-mod} & \multirow{5}{*}{IMU\_ankle} & miss     & 0.855 & 0.865 & 0.862 & 0.834 & 0.861 & 0.858 & 0.863 & 0.864 & 0.824 & \cellcolor{gray!10}0.862 \\
& & wave     & 0.844 & 0.864 & 0.863 & 0.827 & 0.862 & 0.856 & 0.866 & 0.859 & 0.793 & \cellcolor{gray!10}0.867 \\
& & nonlin   & 0.709 & 0.685 & 0.689 & 0.684 & 0.693 & 0.680 & 0.678 & 0.683 & 0.608 & \cellcolor{gray!10}0.685 \\
& & freq     & 0.862 & 0.874 & 0.880 & 0.849 & 0.866 & 0.882 & 0.870 & 0.885 & 0.819 & \cellcolor{gray!10}0.877 \\
& & nonstat  & 0.861 & 0.870 & 0.871 & 0.830 & 0.874 & 0.873 & 0.870 & 0.872 & 0.811 & \cellcolor{gray!10}0.875 \\
\arrayrulecolor{black!20}\midrule
& \multirow{5}{*}{IMU\_hand} & miss     & 0.883 & 0.887 & 0.890 & 0.864 & 0.888 & 0.885 & 0.888 & 0.891 & 0.870 & \cellcolor{gray!10}0.892 \\
& & wave     & 0.868 & 0.872 & 0.880 & 0.860 & 0.882 & 0.875 & 0.884 & 0.877 & 0.869 & \cellcolor{gray!10}0.884 \\
& & nonlin   & 0.819 & 0.854 & 0.841 & 0.813 & 0.843 & 0.838 & 0.849 & 0.857 & 0.817 & \cellcolor{gray!10}0.852 \\
& & freq     & 0.883 & 0.896 & 0.895 & 0.865 & 0.892 & 0.893 & 0.891 & 0.901 & 0.876 & \cellcolor{gray!10}0.895 \\
& & nonstat  & 0.866 & 0.868 & 0.873 & 0.849 & 0.868 & 0.876 & 0.873 & 0.868 & 0.872 & \cellcolor{gray!10}0.876 \\
\arrayrulecolor{black!20}\midrule
& \multirow{5}{*}{IMU\_chest} & miss     & 0.805 & 0.829 & 0.818 & 0.798 & 0.817 & 0.815 & 0.818 & 0.814 & 0.851 & \cellcolor{gray!10}0.818 \\
& & wave     & 0.775 & 0.776 & 0.772 & 0.751 & 0.769 & 0.772 & 0.764 & 0.770 & 0.749 & \cellcolor{gray!10}0.769 \\
& & nonlin   & 0.754 & 0.763 & 0.754 & 0.734 & 0.761 & 0.761 & 0.757 & 0.766 & 0.772 & \cellcolor{gray!10}0.763 \\
& & freq     & 0.878 & 0.895 & 0.892 & 0.870 & 0.887 & 0.891 & 0.889 & 0.886 & 0.861 & \cellcolor{gray!10}0.886 \\
& & nonstat  & 0.809 & 0.811 & 0.812 & 0.809 & 0.805 & 0.793 & 0.812 & 0.815 & 0.825 & \cellcolor{gray!10}0.802 \\
\arrayrulecolor{black!20}\midrule
& \multirow{5}{*}{heart\_rate} & miss     & 0.886 & 0.895 & 0.893 & 0.873 & 0.891 & 0.897 & 0.888 & 0.895 & 0.855 & \cellcolor{gray!10}0.895 \\
& & wave     & 0.885 & 0.894 & 0.900 & 0.869 & 0.891 & 0.892 & 0.890 & 0.901 & 0.836 & \cellcolor{gray!10}0.896 \\
& & nonlin   & 0.892 & 0.899 & 0.896 & 0.868 & 0.898 & 0.898 & 0.890 & 0.895 & 0.890 & \cellcolor{gray!10}0.903 \\
& & freq     & 0.891 & 0.903 & 0.896 & 0.872 & 0.900 & 0.894 & 0.897 & 0.892 & 0.883 & \cellcolor{gray!10}0.901 \\
& & nonstat  & 0.829 & 0.858 & 0.850 & 0.826 & 0.855 & 0.868 & 0.850 & 0.860 & 0.837 & \cellcolor{gray!10}0.891 \\
\arrayrulecolor{black}\midrule
\multirow{5}{*}{2-mod} & \multirow{5}{*}{ankle+\,hand} & miss     & 0.830 & 0.843 & 0.844 & 0.824 & 0.835 & 0.836 & 0.831 & 0.830 & 0.788 & \cellcolor{gray!10}0.849 \\
& & wave     & 0.806 & 0.819 & 0.826 & 0.797 & 0.816 & 0.825 & 0.821 & 0.815 & 0.752 & \cellcolor{gray!10}0.825 \\
& & nonlin   & 0.494 & 0.462 & 0.450 & 0.468 & 0.458 & 0.453 & 0.459 & 0.459 & 0.423 & \cellcolor{gray!10}0.463 \\
& & freq     & 0.857 & 0.868 & 0.868 & 0.847 & 0.864 & 0.864 & 0.867 & 0.870 & 0.780 & \cellcolor{gray!10}0.872 \\
& & nonstat  & 0.829 & 0.840 & 0.843 & 0.814 & 0.835 & 0.840 & 0.832 & 0.845 & 0.765 & \cellcolor{gray!10}0.840 \\
\arrayrulecolor{black!20}\midrule
& \multirow{5}{*}{ankle+\,HR} & miss     & 0.840 & 0.858 & 0.854 & 0.840 & 0.851 & 0.857 & 0.856 & 0.857 & 0.760 & \cellcolor{gray!10}0.866 \\
& & wave     & 0.838 & 0.850 & 0.848 & 0.831 & 0.838 & 0.846 & 0.850 & 0.847 & 0.713 & \cellcolor{gray!10}0.856 \\
& & nonlin   & 0.710 & 0.686 & 0.683 & 0.677 & 0.689 & 0.673 & 0.680 & 0.693 & 0.617 & \cellcolor{gray!10}0.699 \\
& & freq     & 0.860 & 0.875 & 0.876 & 0.861 & 0.878 & 0.874 & 0.877 & 0.875 & 0.813 & \cellcolor{gray!10}0.870 \\
& & nonstat  & 0.761 & 0.807 & 0.817 & 0.798 & 0.808 & 0.811 & 0.810 & 0.802 & 0.720 & \cellcolor{gray!10}0.849 \\
\arrayrulecolor{black}\midrule
\multirow{5}{*}{3-mod} & \multirow{5}{*}{ankle+\,hand+\,chest} & miss     & 0.642 & 0.675 & 0.677 & 0.661 & 0.665 & 0.670 & 0.675 & 0.676 & 0.713 & \cellcolor{gray!10}0.677 \\
& & wave     & 0.546 & 0.572 & 0.561 & 0.551 & 0.567 & 0.566 & 0.549 & 0.563 & 0.515 & \cellcolor{gray!10}0.573 \\
& & nonlin   & 0.146 & 0.142 & 0.145 & 0.159 & 0.146 & 0.145 & 0.141 & 0.139 & 0.196 & \cellcolor{gray!10}0.140 \\
& & freq     & 0.848 & 0.856 & 0.857 & 0.830 & 0.852 & 0.855 & 0.859 & 0.851 & 0.750 & \cellcolor{gray!10}0.851 \\
& & nonstat  & 0.674 & 0.695 & 0.685 & 0.686 & 0.687 & 0.671 & 0.683 & 0.685 & 0.629 & \cellcolor{gray!10}0.682 \\
\arrayrulecolor{black!20}\midrule
& \multirow{5}{*}{ankle+\,hand+\,HR} & miss     & 0.814 & 0.835 & 0.837 & 0.808 & 0.834 & 0.831 & 0.820 & 0.829 & 0.728 & \cellcolor{gray!10}0.841 \\
& & wave     & 0.789 & 0.807 & 0.813 & 0.781 & 0.810 & 0.800 & 0.798 & 0.813 & 0.660 & \cellcolor{gray!10}0.823 \\
& & nonlin   & 0.486 & 0.468 & 0.458 & 0.463 & 0.460 & 0.463 & 0.458 & 0.454 & 0.434 & \cellcolor{gray!10}0.457 \\
& & freq     & 0.856 & 0.874 & 0.868 & 0.845 & 0.865 & 0.873 & 0.869 & 0.866 & 0.771 & \cellcolor{gray!10}0.867 \\
& & nonstat  & 0.755 & 0.796 & 0.797 & 0.766 & 0.786 & 0.798 & 0.791 & 0.787 & 0.680 & \cellcolor{gray!10}0.828 \\
\arrayrulecolor{black}\midrule
\multirow{5}{*}{4-mod} & \multirow{5}{*}{all} & miss     & 0.635 & 0.669 & 0.665 & 0.650 & 0.669 & 0.662 & 0.657 & 0.664 & 0.655 & \cellcolor{gray!10}0.677 \\
& & wave     & 0.514 & 0.534 & 0.542 & 0.533 & 0.536 & 0.530 & 0.522 & 0.535 & 0.413 & \cellcolor{gray!10}0.545 \\
& & nonlin   & 0.142 & 0.142 & 0.140 & 0.157 & 0.144 & 0.131 & 0.142 & 0.145 & 0.203 & \cellcolor{gray!10}0.140 \\
& & freq     & 0.840 & 0.853 & 0.856 & 0.828 & 0.851 & 0.851 & 0.855 & 0.853 & 0.743 & \cellcolor{gray!10}0.852 \\
& & nonstat  & 0.605 & 0.664 & 0.644 & 0.647 & 0.646 & 0.641 & 0.644 & 0.654 & 0.580 & \cellcolor{gray!10}0.675 \\
\arrayrulecolor{black}\bottomrule

\end{tabular}
}
\end{table}

\section{Complete Results on WESAD dataset}\label{app:wesad_complete}

\noindent
For completeness, we provide the full WESAD results behind the aggregate numbers in the main text.
Tables~\ref{tab:appendix-wesad-maestro-fullgrid},
\ref{tab:appendix-wesad-shaspec-fullgrid}, and
\ref{tab:appendix-wesad-flexmoe-fullgrid} show the per-condition F1 scores for MAESTRO, ShaSpec, and Flex-MoE
across corrupted modality subsets and noise types.

\begin{table}[!htbp]
\centering
\caption{Full per-modality-subset, per-noise F1-score ($\uparrow$) on \textbf{WESAD} with the \textbf{MAESTRO} backbone. Each cell is the F1 for one specific (corrupted modality subset, noise type) combination. Noise abbreviations: \emph{miss}=discrete missingness, \emph{wave}=waveform anomaly, \emph{nonlin}=nonlinear degradation, \emph{freq}=frequency attenuation, \emph{nonstat}=nonstationary noise. Clean baseline: F1 = 0.666.}
\label{tab:appendix-wesad-maestro-fullgrid}
\resizebox{1.0\textwidth}{!}{%
\begin{tabular}{@{}lllcccccccccc@{}}
\toprule
\textbf{Card.} & \textbf{Subset} & \textbf{Noise}
& Static & TENT & EATA & READ & SuMi & SoTTA & AeO & PTA & PSDNorm & \cellcolor{gray!10}\textbf{Ours} \\
\midrule

\multirow{5}{*}{1-mod} & \multirow{5}{*}{chest\_EDA} & miss     & 0.610 & 0.615 & 0.618 & 0.615 & 0.617 & 0.606 & 0.611 & 0.605 & 0.640 & \cellcolor{gray!10}0.627 \\
& & wave     & 0.634 & 0.629 & 0.627 & 0.626 & 0.635 & 0.627 & 0.641 & 0.630 & 0.638 & \cellcolor{gray!10}0.628 \\
& & nonlin   & 0.660 & 0.656 & 0.656 & 0.662 & 0.653 & 0.659 & 0.666 & 0.644 & 0.652 & \cellcolor{gray!10}0.632 \\
& & freq     & 0.657 & 0.658 & 0.655 & 0.654 & 0.657 & 0.661 & 0.662 & 0.645 & 0.652 & \cellcolor{gray!10}0.624 \\
& & nonstat  & 0.604 & 0.602 & 0.594 & 0.590 & 0.599 & 0.602 & 0.596 & 0.594 & 0.637 & \cellcolor{gray!10}0.627 \\
\arrayrulecolor{black!20}\midrule
& \multirow{5}{*}{chest\_ACC} & miss     & 0.658 & 0.656 & 0.656 & 0.655 & 0.655 & 0.653 & 0.663 & 0.646 & 0.665 & \cellcolor{gray!10}0.637 \\
& & wave     & 0.666 & 0.656 & 0.655 & 0.664 & 0.660 & 0.663 & 0.671 & 0.643 & 0.664 & \cellcolor{gray!10}0.645 \\
& & nonlin   & 0.660 & 0.657 & 0.658 & 0.663 & 0.662 & 0.651 & 0.668 & 0.643 & 0.665 & \cellcolor{gray!10}0.642 \\
& & freq     & 0.655 & 0.664 & 0.658 & 0.658 & 0.656 & 0.661 & 0.666 & 0.648 & 0.665 & \cellcolor{gray!10}0.649 \\
& & nonstat  & 0.664 & 0.663 & 0.661 & 0.657 & 0.655 & 0.660 & 0.669 & 0.640 & 0.665 & \cellcolor{gray!10}0.637 \\
\arrayrulecolor{black!20}\midrule
& \multirow{5}{*}{wrist\_ACC} & miss     & 0.638 & 0.637 & 0.636 & 0.636 & 0.637 & 0.648 & 0.632 & 0.626 & 0.601 & \cellcolor{gray!10}0.632 \\
& & wave     & 0.644 & 0.645 & 0.642 & 0.645 & 0.638 & 0.629 & 0.647 & 0.628 & 0.637 & \cellcolor{gray!10}0.629 \\
& & nonlin   & 0.691 & 0.688 & 0.687 & 0.700 & 0.702 & 0.689 & 0.694 & 0.700 & 0.689 & \cellcolor{gray!10}0.680 \\
& & freq     & 0.654 & 0.653 & 0.653 & 0.643 & 0.650 & 0.649 & 0.649 & 0.645 & 0.625 & \cellcolor{gray!10}0.639 \\
& & nonstat  & 0.644 & 0.631 & 0.635 & 0.633 & 0.636 & 0.639 & 0.631 & 0.638 & 0.602 & \cellcolor{gray!10}0.638 \\
\arrayrulecolor{black!20}\midrule
& \multirow{5}{*}{wrist\_EDA} & miss     & 0.630 & 0.640 & 0.646 & 0.630 & 0.629 & 0.639 & 0.625 & 0.616 & 0.654 & \cellcolor{gray!10}0.635 \\
& & wave     & 0.638 & 0.636 & 0.645 & 0.634 & 0.634 & 0.637 & 0.626 & 0.620 & 0.643 & \cellcolor{gray!10}0.630 \\
& & nonlin   & 0.663 & 0.662 & 0.664 & 0.660 & 0.659 & 0.666 & 0.658 & 0.644 & 0.652 & \cellcolor{gray!10}0.638 \\
& & freq     & 0.664 & 0.660 & 0.659 & 0.655 & 0.665 & 0.663 & 0.667 & 0.644 & 0.663 & \cellcolor{gray!10}0.630 \\
& & nonstat  & 0.635 & 0.640 & 0.643 & 0.626 & 0.634 & 0.646 & 0.630 & 0.621 & 0.650 & \cellcolor{gray!10}0.633 \\
\arrayrulecolor{black!20}\midrule
& \multirow{5}{*}{chest\_ECG} & miss     & 0.658 & 0.662 & 0.657 & 0.657 & 0.660 & 0.661 & 0.657 & 0.647 & 0.659 & \cellcolor{gray!10}0.643 \\
& & wave     & 0.659 & 0.659 & 0.665 & 0.663 & 0.659 & 0.659 & 0.670 & 0.645 & 0.658 & \cellcolor{gray!10}0.642 \\
& & nonlin   & 0.663 & 0.653 & 0.664 & 0.653 & 0.653 & 0.659 & 0.661 & 0.644 & 0.663 & \cellcolor{gray!10}0.640 \\
& & freq     & 0.663 & 0.656 & 0.661 & 0.664 & 0.660 & 0.654 & 0.664 & 0.644 & 0.657 & \cellcolor{gray!10}0.643 \\
& & nonstat  & 0.661 & 0.663 & 0.660 & 0.658 & 0.659 & 0.658 & 0.661 & 0.646 & 0.660 & \cellcolor{gray!10}0.640 \\
\arrayrulecolor{black!20}\midrule
& \multirow{5}{*}{chest\_EMG} & miss     & 0.664 & 0.659 & 0.664 & 0.657 & 0.657 & 0.655 & 0.662 & 0.642 & 0.666 & \cellcolor{gray!10}0.647 \\
& & wave     & 0.662 & 0.658 & 0.650 & 0.663 & 0.663 & 0.663 & 0.663 & 0.650 & 0.660 & \cellcolor{gray!10}0.639 \\
& & nonlin   & 0.667 & 0.660 & 0.660 & 0.659 & 0.661 & 0.662 & 0.660 & 0.642 & 0.659 & \cellcolor{gray!10}0.644 \\
& & freq     & 0.658 & 0.667 & 0.659 & 0.661 & 0.653 & 0.665 & 0.658 & 0.646 & 0.663 & \cellcolor{gray!10}0.639 \\
& & nonstat  & 0.661 & 0.659 & 0.658 & 0.662 & 0.667 & 0.663 & 0.659 & 0.648 & 0.660 & \cellcolor{gray!10}0.636 \\
\arrayrulecolor{black!20}\midrule
& \multirow{5}{*}{chest\_RESP} & miss     & 0.660 & 0.653 & 0.662 & 0.655 & 0.661 & 0.655 & 0.654 & 0.643 & 0.661 & \cellcolor{gray!10}0.640 \\
& & wave     & 0.653 & 0.656 & 0.662 & 0.652 & 0.654 & 0.655 & 0.662 & 0.639 & 0.661 & \cellcolor{gray!10}0.641 \\
& & nonlin   & 0.633 & 0.641 & 0.635 & 0.646 & 0.630 & 0.628 & 0.624 & 0.631 & 0.633 & \cellcolor{gray!10}0.620 \\
& & freq     & 0.662 & 0.663 & 0.655 & 0.664 & 0.666 & 0.653 & 0.663 & 0.646 & 0.661 & \cellcolor{gray!10}0.628 \\
& & nonstat  & 0.656 & 0.652 & 0.656 & 0.650 & 0.653 & 0.652 & 0.655 & 0.639 & 0.660 & \cellcolor{gray!10}0.627 \\
\arrayrulecolor{black!20}\midrule
& \multirow{5}{*}{wrist\_TEMP} & miss     & 0.660 & 0.662 & 0.662 & 0.654 & 0.659 & 0.657 & 0.663 & 0.645 & 0.658 & \cellcolor{gray!10}0.639 \\
& & wave     & 0.660 & 0.652 & 0.661 & 0.658 & 0.663 & 0.664 & 0.667 & 0.646 & 0.660 & \cellcolor{gray!10}0.639 \\
& & nonlin   & 0.664 & 0.659 & 0.663 & 0.657 & 0.667 & 0.666 & 0.661 & 0.647 & 0.663 & \cellcolor{gray!10}0.637 \\
& & freq     & 0.657 & 0.662 & 0.668 & 0.653 & 0.664 & 0.660 & 0.671 & 0.648 & 0.664 & \cellcolor{gray!10}0.647 \\
& & nonstat  & 0.664 & 0.659 & 0.662 & 0.657 & 0.661 & 0.660 & 0.657 & 0.641 & 0.660 & \cellcolor{gray!10}0.646 \\
\arrayrulecolor{black!20}\midrule
& \multirow{5}{*}{wrist\_BVP} & miss     & 0.658 & 0.661 & 0.656 & 0.655 & 0.661 & 0.666 & 0.671 & 0.654 & 0.661 & \cellcolor{gray!10}0.656 \\
& & wave     & 0.665 & 0.662 & 0.666 & 0.665 & 0.662 & 0.666 & 0.661 & 0.650 & 0.660 & \cellcolor{gray!10}0.660 \\
& & nonlin   & 0.669 & 0.675 & 0.662 & 0.674 & 0.664 & 0.662 & 0.669 & 0.661 & 0.646 & \cellcolor{gray!10}0.665 \\
& & freq     & 0.660 & 0.664 & 0.666 & 0.660 & 0.666 & 0.666 & 0.666 & 0.649 & 0.663 & \cellcolor{gray!10}0.641 \\
& & nonstat  & 0.661 & 0.657 & 0.661 & 0.661 & 0.659 & 0.660 & 0.671 & 0.646 & 0.658 & \cellcolor{gray!10}0.655 \\
\arrayrulecolor{black}\midrule
\multirow{5}{*}{2-mod} & \multirow{5}{*}{C\_EDA+\,C\_ACC} & miss     & 0.621 & 0.616 & 0.616 & 0.612 & 0.617 & 0.615 & 0.609 & 0.611 & 0.644 & \cellcolor{gray!10}0.618 \\
& & wave     & 0.633 & 0.621 & 0.626 & 0.624 & 0.633 & 0.640 & 0.630 & 0.619 & 0.645 & \cellcolor{gray!10}0.630 \\
& & nonlin   & 0.657 & 0.663 & 0.665 & 0.660 & 0.658 & 0.655 & 0.656 & 0.644 & 0.651 & \cellcolor{gray!10}0.634 \\
& & freq     & 0.654 & 0.653 & 0.647 & 0.652 & 0.658 & 0.661 & 0.657 & 0.649 & 0.650 & \cellcolor{gray!10}0.632 \\
& & nonstat  & 0.603 & 0.598 & 0.602 & 0.591 & 0.598 & 0.605 & 0.588 & 0.594 & 0.636 & \cellcolor{gray!10}0.614 \\
\arrayrulecolor{black!20}\midrule
& \multirow{5}{*}{W\_EDA+\,W\_ACC} & miss     & 0.612 & 0.620 & 0.620 & 0.609 & 0.609 & 0.627 & 0.592 & 0.610 & 0.592 & \cellcolor{gray!10}0.641 \\
& & wave     & 0.601 & 0.608 & 0.618 & 0.599 & 0.602 & 0.598 & 0.582 & 0.585 & 0.609 & \cellcolor{gray!10}0.613 \\
& & nonlin   & 0.689 & 0.683 & 0.680 & 0.700 & 0.687 & 0.694 & 0.691 & 0.706 & 0.680 & \cellcolor{gray!10}0.680 \\
& & freq     & 0.651 & 0.647 & 0.651 & 0.655 & 0.651 & 0.656 & 0.647 & 0.642 & 0.621 & \cellcolor{gray!10}0.647 \\
& & nonstat  & 0.623 & 0.634 & 0.633 & 0.617 & 0.618 & 0.632 & 0.616 & 0.613 & 0.591 & \cellcolor{gray!10}0.637 \\
\arrayrulecolor{black!20}\midrule
& \multirow{5}{*}{C\_ACC+\,W\_ACC} & miss     & 0.637 & 0.642 & 0.644 & 0.643 & 0.651 & 0.639 & 0.630 & 0.638 & 0.596 & \cellcolor{gray!10}0.647 \\
& & wave     & 0.638 & 0.641 & 0.648 & 0.634 & 0.636 & 0.648 & 0.649 & 0.619 & 0.633 & \cellcolor{gray!10}0.634 \\
& & nonlin   & 0.696 & 0.681 & 0.681 & 0.693 & 0.694 & 0.691 & 0.685 & 0.696 & 0.686 & \cellcolor{gray!10}0.689 \\
& & freq     & 0.655 & 0.650 & 0.653 & 0.651 & 0.656 & 0.641 & 0.647 & 0.635 & 0.614 & \cellcolor{gray!10}0.641 \\
& & nonstat  & 0.640 & 0.636 & 0.631 & 0.635 & 0.638 & 0.637 & 0.641 & 0.635 & 0.598 & \cellcolor{gray!10}0.653 \\
\arrayrulecolor{black!20}\midrule
& \multirow{5}{*}{C\_ECG+\,C\_EMG} & miss     & 0.661 & 0.655 & 0.659 & 0.661 & 0.661 & 0.657 & 0.665 & 0.647 & 0.657 & \cellcolor{gray!10}0.643 \\
& & wave     & 0.658 & 0.658 & 0.661 & 0.660 & 0.661 & 0.659 & 0.665 & 0.649 & 0.663 & \cellcolor{gray!10}0.644 \\
& & nonlin   & 0.662 & 0.654 & 0.659 & 0.659 & 0.660 & 0.667 & 0.660 & 0.643 & 0.663 & \cellcolor{gray!10}0.640 \\
& & freq     & 0.661 & 0.656 & 0.665 & 0.654 & 0.657 & 0.661 & 0.665 & 0.649 & 0.662 & \cellcolor{gray!10}0.643 \\
& & nonstat  & 0.661 & 0.660 & 0.659 & 0.665 & 0.656 & 0.658 & 0.665 & 0.643 & 0.661 & \cellcolor{gray!10}0.642 \\
\arrayrulecolor{black}\midrule

\end{tabular}
}
\end{table}
\clearpage
\begin{table}[!htbp]\ContinuedFloat
\centering
% no repeated label here
\resizebox{1.0\textwidth}{!}{%
\begin{tabular}{@{}lllcccccccccc@{}}
\toprule
\textbf{Card.} & \textbf{Subset} & \textbf{Noise}
& Static & TENT & EATA & READ & SuMi & SoTTA & AeO & PTA & PSDNorm & \cellcolor{gray!10}\textbf{Ours} \\
\midrule

\multirow{5}{*}{3-mod} & \multirow{5}{2.6cm}{\raggedright C\_ACC+\allowbreak \,W\_ACC+\allowbreak \newline \,W\_EDA} & miss     & 0.609 & 0.621 & 0.622 & 0.611 & 0.598 & 0.625 & 0.593 & 0.603 & 0.589 & \cellcolor{gray!10}0.617 \\
& & wave     & 0.604 & 0.616 & 0.617 & 0.605 & 0.606 & 0.598 & 0.585 & 0.595 & 0.613 & \cellcolor{gray!10}0.592 \\
& & nonlin   & 0.695 & 0.681 & 0.687 & 0.690 & 0.694 & 0.686 & 0.692 & 0.701 & 0.676 & \cellcolor{gray!10}0.687 \\
& & freq     & 0.653 & 0.652 & 0.652 & 0.651 & 0.651 & 0.643 & 0.642 & 0.642 & 0.620 & \cellcolor{gray!10}0.653 \\
& & nonstat  & 0.615 & 0.636 & 0.628 & 0.617 & 0.623 & 0.628 & 0.600 & 0.616 & 0.591 & \cellcolor{gray!10}0.635 \\
\arrayrulecolor{black!20}\midrule
& \multirow{5}{2.6cm}{\raggedright C\_ECG+\allowbreak \,C\_EMG+\allowbreak \newline \,C\_RESP} & miss     & 0.658 & 0.657 & 0.659 & 0.658 & 0.660 & 0.652 & 0.647 & 0.653 & 0.664 & \cellcolor{gray!10}0.624 \\
& & wave     & 0.657 & 0.660 & 0.659 & 0.651 & 0.663 & 0.653 & 0.650 & 0.643 & 0.657 & \cellcolor{gray!10}0.618 \\
& & nonlin   & 0.634 & 0.636 & 0.636 & 0.636 & 0.635 & 0.636 & 0.618 & 0.626 & 0.625 & \cellcolor{gray!10}0.609 \\
& & freq     & 0.664 & 0.659 & 0.659 & 0.660 & 0.658 & 0.661 & 0.659 & 0.644 & 0.658 & \cellcolor{gray!10}0.607 \\
& & nonstat  & 0.657 & 0.655 & 0.655 & 0.655 & 0.653 & 0.664 & 0.647 & 0.642 & 0.662 & \cellcolor{gray!10}0.636 \\
\arrayrulecolor{black}\midrule
\multirow{5}{*}{4-mod} & \multirow{5}{2.6cm}{\raggedright C\_EDA+\allowbreak \,C\_ACC+\allowbreak \newline
\,W\_EDA+\allowbreak \,W\_ACC} & miss     & 0.482 & 0.462 & 0.467 & 0.481 & 0.479 & 0.474 & 0.490 & 0.483 & 0.561 & \cellcolor{gray!10}0.424 \\
& & wave     & 0.509 & 0.487 & 0.491 & 0.483 & 0.491 & 0.491 & 0.510 & 0.496 & 0.583 & \cellcolor{gray!10}0.491 \\
& & nonlin   & 0.692 & 0.683 & 0.682 & 0.684 & 0.692 & 0.696 & 0.670 & 0.692 & 0.677 & \cellcolor{gray!10}0.677 \\
& & freq     & 0.650 & 0.656 & 0.651 & 0.644 & 0.657 & 0.655 & 0.655 & 0.637 & 0.607 & \cellcolor{gray!10}0.657 \\
& & nonstat  & 0.494 & 0.467 & 0.478 & 0.489 & 0.475 & 0.475 & 0.496 & 0.486 & 0.561 & \cellcolor{gray!10}0.445 \\
\arrayrulecolor{black!20}\midrule
& \multirow{5}{2.6cm}{\raggedright W\_ACC+\allowbreak \,W\_EDA+\allowbreak \newline \,W\_TEMP+\allowbreak \,W\_BVP} & miss     & 0.612 & 0.623 & 0.626 & 0.614 & 0.619 & 0.632 & 0.585 & 0.607 & 0.586 & \cellcolor{gray!10}0.634 \\
& & wave     & 0.600 & 0.600 & 0.615 & 0.602 & 0.606 & 0.619 & 0.589 & 0.596 & 0.617 & \cellcolor{gray!10}0.612 \\
& & nonlin   & 0.704 & 0.707 & 0.706 & 0.699 & 0.697 & 0.698 & 0.681 & 0.685 & 0.683 & \cellcolor{gray!10}0.718 \\
& & freq     & 0.656 & 0.660 & 0.656 & 0.646 & 0.650 & 0.658 & 0.647 & 0.641 & 0.625 & \cellcolor{gray!10}0.644 \\
& & nonstat  & 0.625 & 0.637 & 0.637 & 0.618 & 0.620 & 0.632 & 0.599 & 0.613 & 0.584 & \cellcolor{gray!10}0.638 \\
\arrayrulecolor{black!20}\midrule
& \multirow{5}{2.6cm}{\raggedright C\_ECG+\allowbreak \,C\_EMG+\allowbreak  \,C\_RESP+\allowbreak \,W\_TEMP} & miss     & 0.661 & 0.658 & 0.658 & 0.652 & 0.653 & 0.663 & 0.665 & 0.644 & 0.655 & \cellcolor{gray!10}0.623 \\
& & wave     & 0.660 & 0.658 & 0.658 & 0.657 & 0.652 & 0.659 & 0.655 & 0.644 & 0.657 & \cellcolor{gray!10}0.622 \\
& & nonlin   & 0.628 & 0.642 & 0.633 & 0.644 & 0.635 & 0.631 & 0.619 & 0.618 & 0.625 & \cellcolor{gray!10}0.610 \\
& & freq     & 0.664 & 0.659 & 0.659 & 0.663 & 0.661 & 0.666 & 0.660 & 0.646 & 0.662 & \cellcolor{gray!10}0.602 \\
& & nonstat  & 0.657 & 0.658 & 0.655 & 0.662 & 0.656 & 0.658 & 0.655 & 0.640 & 0.659 & \cellcolor{gray!10}0.632 \\
\arrayrulecolor{black}\midrule
\multirow{5}{*}{5-mod} & \multirow{5}{2.6cm}{all} & miss     & 0.660 & 0.662 & 0.661 & 0.653 & 0.658 & 0.659 & 0.659 & 0.651 & 0.660 & \cellcolor{gray!10}0.640 \\
& & wave     & 0.663 & 0.655 & 0.656 & 0.662 & 0.659 & 0.657 & 0.654 & 0.654 & 0.651 & \cellcolor{gray!10}0.625 \\
& & nonlin   & 0.634 & 0.652 & 0.645 & 0.640 & 0.622 & 0.647 & 0.622 & 0.635 & 0.608 & \cellcolor{gray!10}0.622 \\
& & freq     & 0.658 & 0.665 & 0.664 & 0.659 & 0.653 & 0.651 & 0.675 & 0.647 & 0.663 & \cellcolor{gray!10}0.606 \\
& & nonstat  & 0.653 & 0.655 & 0.652 & 0.652 & 0.651 & 0.655 & 0.654 & 0.644 & 0.661 & \cellcolor{gray!10}0.633 \\
\arrayrulecolor{black}\bottomrule

\end{tabular}
}
\end{table}

\begin{table}[!htbp]
\centering
\caption{Full per-modality-subset, per-noise F1-score ($\uparrow$) on \textbf{WESAD} with the \textbf{ShaSpec} backbone. Each cell is the F1 for one specific (corrupted modality subset, noise type) combination. Noise abbreviations: \emph{miss}=discrete missingness, \emph{wave}=waveform anomaly, \emph{nonlin}=nonlinear degradation, \emph{freq}=frequency attenuation, \emph{nonstat}=nonstationary noise. Clean baseline: F1 = 0.554.}
\label{tab:appendix-wesad-shaspec-fullgrid}
\resizebox{1.0\textwidth}{!}{%
\begin{tabular}{@{}lllcccccccccc@{}}
\toprule
\textbf{Card.} & \textbf{Subset} & \textbf{Noise}
& Static & TENT & EATA & READ & SuMi & SoTTA & AeO & PTA & PSDNorm & \cellcolor{gray!10}\textbf{Ours} \\
\midrule

\multirow{5}{*}{1-mod} & \multirow{5}{*}{chest\_EDA} & miss     & 0.506 & 0.505 & 0.511 & 0.505 & 0.507 & 0.514 & 0.497 & 0.496 & 0.454 & \cellcolor{gray!10}0.523 \\
& & wave     & 0.493 & 0.497 & 0.496 & 0.497 & 0.493 & 0.501 & 0.484 & 0.492 & 0.465 & \cellcolor{gray!10}0.520 \\
& & nonlin   & 0.523 & 0.521 & 0.527 & 0.514 & 0.520 & 0.524 & 0.515 & 0.506 & 0.463 & \cellcolor{gray!10}0.545 \\
& & freq     & 0.532 & 0.530 & 0.539 & 0.526 & 0.527 & 0.531 & 0.515 & 0.521 & 0.478 & \cellcolor{gray!10}0.548 \\
& & nonstat  & 0.511 & 0.506 & 0.512 & 0.508 & 0.511 & 0.513 & 0.498 & 0.501 & 0.464 & \cellcolor{gray!10}0.531 \\
\arrayrulecolor{black!20}\midrule
& \multirow{5}{*}{chest\_ACC} & miss     & 0.547 & 0.547 & 0.550 & 0.555 & 0.542 & 0.547 & 0.542 & 0.551 & 0.493 & \cellcolor{gray!10}0.570 \\
& & wave     & 0.558 & 0.559 & 0.557 & 0.564 & 0.555 & 0.552 & 0.552 & 0.561 & 0.507 & \cellcolor{gray!10}0.573 \\
& & nonlin   & 0.543 & 0.543 & 0.548 & 0.537 & 0.539 & 0.545 & 0.537 & 0.524 & 0.498 & \cellcolor{gray!10}0.571 \\
& & freq     & 0.553 & 0.558 & 0.555 & 0.539 & 0.550 & 0.557 & 0.542 & 0.532 & 0.506 & \cellcolor{gray!10}0.581 \\
& & nonstat  & 0.560 & 0.559 & 0.564 & 0.566 & 0.558 & 0.564 & 0.550 & 0.554 & 0.511 & \cellcolor{gray!10}0.578 \\
\arrayrulecolor{black!20}\midrule
& \multirow{5}{*}{wrist\_ACC} & miss     & 0.549 & 0.552 & 0.556 & 0.546 & 0.545 & 0.556 & 0.533 & 0.538 & 0.452 & \cellcolor{gray!10}0.558 \\
& & wave     & 0.540 & 0.540 & 0.544 & 0.533 & 0.535 & 0.548 & 0.527 & 0.535 & 0.484 & \cellcolor{gray!10}0.547 \\
& & nonlin   & 0.559 & 0.554 & 0.566 & 0.552 & 0.555 & 0.558 & 0.546 & 0.543 & 0.494 & \cellcolor{gray!10}0.568 \\
& & freq     & 0.536 & 0.541 & 0.546 & 0.530 & 0.531 & 0.540 & 0.524 & 0.520 & 0.476 & \cellcolor{gray!10}0.548 \\
& & nonstat  & 0.511 & 0.520 & 0.519 & 0.519 & 0.512 & 0.516 & 0.501 & 0.510 & 0.453 & \cellcolor{gray!10}0.532 \\
\arrayrulecolor{black!20}\midrule
& \multirow{5}{*}{wrist\_EDA} & miss     & 0.542 & 0.534 & 0.544 & 0.526 & 0.538 & 0.541 & 0.525 & 0.514 & 0.502 & \cellcolor{gray!10}0.559 \\
& & wave     & 0.539 & 0.538 & 0.539 & 0.532 & 0.532 & 0.541 & 0.519 & 0.521 & 0.499 & \cellcolor{gray!10}0.557 \\
& & nonlin   & 0.545 & 0.543 & 0.548 & 0.539 & 0.542 & 0.551 & 0.530 & 0.524 & 0.506 & \cellcolor{gray!10}0.565 \\
& & freq     & 0.553 & 0.550 & 0.555 & 0.543 & 0.550 & 0.553 & 0.544 & 0.536 & 0.507 & \cellcolor{gray!10}0.579 \\
& & nonstat  & 0.539 & 0.538 & 0.541 & 0.531 & 0.534 & 0.544 & 0.525 & 0.517 & 0.500 & \cellcolor{gray!10}0.560 \\
\arrayrulecolor{black!20}\midrule
& \multirow{5}{*}{chest\_ECG} & miss     & 0.553 & 0.556 & 0.559 & 0.541 & 0.552 & 0.558 & 0.539 & 0.526 & 0.494 & \cellcolor{gray!10}0.569 \\
& & wave     & 0.500 & 0.485 & 0.498 & 0.497 & 0.493 & 0.501 & 0.495 & 0.480 & 0.509 & \cellcolor{gray!10}0.516 \\
& & nonlin   & 0.470 & 0.490 & 0.478 & 0.490 & 0.484 & 0.485 & 0.482 & 0.485 & 0.471 & \cellcolor{gray!10}0.506 \\
& & freq     & 0.433 & 0.440 & 0.439 & 0.446 & 0.444 & 0.443 & 0.457 & 0.449 & 0.474 & \cellcolor{gray!10}0.476 \\
& & nonstat  & 0.567 & 0.576 & 0.579 & 0.566 & 0.573 & 0.584 & 0.564 & 0.562 & 0.517 & \cellcolor{gray!10}0.582 \\
\arrayrulecolor{black!20}\midrule
& \multirow{5}{*}{chest\_EMG} & miss     & 0.554 & 0.544 & 0.557 & 0.543 & 0.545 & 0.552 & 0.543 & 0.531 & 0.504 & \cellcolor{gray!10}0.577 \\
& & wave     & 0.550 & 0.548 & 0.555 & 0.544 & 0.548 & 0.562 & 0.533 & 0.537 & 0.503 & \cellcolor{gray!10}0.575 \\
& & nonlin   & 0.554 & 0.552 & 0.556 & 0.542 & 0.552 & 0.549 & 0.542 & 0.536 & 0.507 & \cellcolor{gray!10}0.570 \\
& & freq     & 0.554 & 0.548 & 0.559 & 0.545 & 0.549 & 0.555 & 0.543 & 0.536 & 0.504 & \cellcolor{gray!10}0.581 \\
& & nonstat  & 0.554 & 0.549 & 0.558 & 0.536 & 0.549 & 0.556 & 0.537 & 0.532 & 0.504 & \cellcolor{gray!10}0.572 \\
\arrayrulecolor{black!20}\midrule
& \multirow{5}{*}{chest\_RESP} & miss     & 0.555 & 0.553 & 0.559 & 0.547 & 0.556 & 0.559 & 0.543 & 0.539 & 0.509 & \cellcolor{gray!10}0.579 \\
& & wave     & 0.553 & 0.550 & 0.558 & 0.540 & 0.543 & 0.550 & 0.537 & 0.539 & 0.509 & \cellcolor{gray!10}0.571 \\
& & nonlin   & 0.519 & 0.519 & 0.520 & 0.512 & 0.519 & 0.529 & 0.506 & 0.508 & 0.474 & \cellcolor{gray!10}0.523 \\
& & freq     & 0.553 & 0.550 & 0.554 & 0.544 & 0.554 & 0.548 & 0.535 & 0.535 & 0.504 & \cellcolor{gray!10}0.578 \\
& & nonstat  & 0.555 & 0.552 & 0.560 & 0.547 & 0.548 & 0.554 & 0.542 & 0.534 & 0.505 & \cellcolor{gray!10}0.574 \\
\arrayrulecolor{black!20}\midrule
& \multirow{5}{*}{wrist\_TEMP} & miss     & 0.555 & 0.554 & 0.561 & 0.548 & 0.548 & 0.559 & 0.538 & 0.537 & 0.509 & \cellcolor{gray!10}0.577 \\
& & wave     & 0.556 & 0.555 & 0.562 & 0.551 & 0.548 & 0.550 & 0.543 & 0.538 & 0.510 & \cellcolor{gray!10}0.582 \\
& & nonlin   & 0.558 & 0.557 & 0.561 & 0.550 & 0.551 & 0.559 & 0.538 & 0.536 & 0.505 & \cellcolor{gray!10}0.573 \\
& & freq     & 0.554 & 0.555 & 0.558 & 0.545 & 0.550 & 0.553 & 0.535 & 0.537 & 0.505 & \cellcolor{gray!10}0.574 \\
& & nonstat  & 0.559 & 0.556 & 0.561 & 0.551 & 0.551 & 0.553 & 0.539 & 0.534 & 0.507 & \cellcolor{gray!10}0.575 \\
\arrayrulecolor{black!20}\midrule
& \multirow{5}{*}{wrist\_BVP} & miss     & 0.557 & 0.558 & 0.558 & 0.542 & 0.552 & 0.551 & 0.540 & 0.538 & 0.506 & \cellcolor{gray!10}0.569 \\
& & wave     & 0.557 & 0.553 & 0.559 & 0.545 & 0.551 & 0.561 & 0.535 & 0.541 & 0.504 & \cellcolor{gray!10}0.566 \\
& & nonlin   & 0.522 & 0.519 & 0.522 & 0.509 & 0.511 & 0.527 & 0.507 & 0.503 & 0.500 & \cellcolor{gray!10}0.545 \\
& & freq     & 0.553 & 0.550 & 0.557 & 0.544 & 0.547 & 0.552 & 0.539 & 0.534 & 0.503 & \cellcolor{gray!10}0.570 \\
& & nonstat  & 0.553 & 0.551 & 0.556 & 0.548 & 0.550 & 0.555 & 0.536 & 0.533 & 0.507 & \cellcolor{gray!10}0.567 \\
\arrayrulecolor{black}\midrule
\multirow{5}{*}{2-mod} & \multirow{5}{*}{C\_EDA+\,C\_ACC} & miss     & 0.514 & 0.509 & 0.518 & 0.507 & 0.511 & 0.519 & 0.513 & 0.506 & 0.440 & \cellcolor{gray!10}0.532 \\
& & wave     & 0.539 & 0.544 & 0.542 & 0.535 & 0.538 & 0.535 & 0.526 & 0.537 & 0.479 & \cellcolor{gray!10}0.553 \\
& & nonlin   & 0.514 & 0.512 & 0.516 & 0.506 & 0.505 & 0.523 & 0.503 & 0.497 & 0.454 & \cellcolor{gray!10}0.534 \\
& & freq     & 0.537 & 0.531 & 0.542 & 0.532 & 0.531 & 0.534 & 0.531 & 0.512 & 0.480 & \cellcolor{gray!10}0.554 \\
& & nonstat  & 0.517 & 0.517 & 0.518 & 0.520 & 0.513 & 0.519 & 0.509 & 0.508 & 0.462 & \cellcolor{gray!10}0.535 \\
\arrayrulecolor{black!20}\midrule
& \multirow{5}{*}{W\_EDA+\,W\_ACC} & miss     & 0.542 & 0.541 & 0.546 & 0.536 & 0.537 & 0.543 & 0.522 & 0.527 & 0.454 & \cellcolor{gray!10}0.552 \\
& & wave     & 0.538 & 0.542 & 0.540 & 0.535 & 0.536 & 0.546 & 0.517 & 0.518 & 0.495 & \cellcolor{gray!10}0.547 \\
& & nonlin   & 0.554 & 0.554 & 0.556 & 0.540 & 0.547 & 0.556 & 0.532 & 0.528 & 0.495 & \cellcolor{gray!10}0.559 \\
& & freq     & 0.535 & 0.537 & 0.541 & 0.530 & 0.532 & 0.539 & 0.524 & 0.520 & 0.475 & \cellcolor{gray!10}0.539 \\
& & nonstat  & 0.516 & 0.522 & 0.523 & 0.519 & 0.513 & 0.529 & 0.504 & 0.509 & 0.455 & \cellcolor{gray!10}0.534 \\
\arrayrulecolor{black!20}\midrule
& \multirow{5}{*}{C\_ACC+\,W\_ACC} & miss     & 0.529 & 0.541 & 0.533 & 0.552 & 0.537 & 0.537 & 0.528 & 0.537 & 0.438 & \cellcolor{gray!10}0.547 \\
& & wave     & 0.508 & 0.521 & 0.512 & 0.537 & 0.513 & 0.509 & 0.497 & 0.513 & 0.474 & \cellcolor{gray!10}0.534 \\
& & nonlin   & 0.551 & 0.548 & 0.555 & 0.539 & 0.547 & 0.556 & 0.539 & 0.537 & 0.489 & \cellcolor{gray!10}0.567 \\
& & freq     & 0.537 & 0.536 & 0.545 & 0.533 & 0.533 & 0.540 & 0.523 & 0.521 & 0.473 & \cellcolor{gray!10}0.548 \\
& & nonstat  & 0.499 & 0.504 & 0.504 & 0.506 & 0.500 & 0.498 & 0.491 & 0.507 & 0.444 & \cellcolor{gray!10}0.522 \\
\arrayrulecolor{black!20}\midrule
& \multirow{5}{*}{C\_ECG+\,C\_EMG} & miss     & 0.549 & 0.550 & 0.554 & 0.541 & 0.547 & 0.552 & 0.530 & 0.527 & 0.499 & \cellcolor{gray!10}0.567 \\
& & wave     & 0.491 & 0.481 & 0.493 & 0.491 & 0.487 & 0.500 & 0.498 & 0.475 & 0.507 & \cellcolor{gray!10}0.519 \\
& & nonlin   & 0.470 & 0.493 & 0.478 & 0.486 & 0.483 & 0.488 & 0.477 & 0.484 & 0.472 & \cellcolor{gray!10}0.505 \\
& & freq     & 0.433 & 0.441 & 0.437 & 0.449 & 0.447 & 0.446 & 0.461 & 0.453 & 0.474 & \cellcolor{gray!10}0.482 \\
& & nonstat  & 0.567 & 0.576 & 0.576 & 0.572 & 0.566 & 0.576 & 0.560 & 0.560 & 0.507 & \cellcolor{gray!10}0.581 \\
\arrayrulecolor{black}\midrule

\end{tabular}
}
\end{table}
\clearpage
\clearpage
\begin{table}[!htbp]\ContinuedFloat
\centering
% no repeated label here
\resizebox{1.0\textwidth}{!}{%
\begin{tabular}{@{}lllcccccccccc@{}}
\toprule
\textbf{Card.} & \textbf{Subset} & \textbf{Noise}
& Static & TENT & EATA & READ & SuMi & SoTTA & AeO & PTA & PSDNorm & \cellcolor{gray!10}\textbf{Ours} \\
\midrule
\multirow{5}{*}{3-mod} & \multirow{5}{2.6cm}{\raggedright C\_ACC+\allowbreak \,W\_ACC+\allowbreak \newline \,W\_EDA} & miss     & 0.533 & 0.534 & 0.538 & 0.526 & 0.537 & 0.542 & 0.523 & 0.526 & 0.435 & \cellcolor{gray!10}0.553 \\
& & wave     & 0.513 & 0.519 & 0.518 & 0.524 & 0.524 & 0.523 & 0.516 & 0.528 & 0.477 & \cellcolor{gray!10}0.540 \\
& & nonlin   & 0.544 & 0.537 & 0.544 & 0.539 & 0.541 & 0.544 & 0.527 & 0.526 & 0.488 & \cellcolor{gray!10}0.563 \\
& & freq     & 0.537 & 0.538 & 0.545 & 0.531 & 0.534 & 0.541 & 0.517 & 0.522 & 0.472 & \cellcolor{gray!10}0.546 \\
& & nonstat  & 0.505 & 0.511 & 0.510 & 0.514 & 0.505 & 0.509 & 0.494 & 0.508 & 0.445 & \cellcolor{gray!10}0.530 \\
\arrayrulecolor{black!20}\midrule
& \multirow{5}{2.6cm}{\raggedright C\_ECG+\allowbreak \,C\_EMG+\allowbreak \newline \,C\_RESP} & miss     & 0.555 & 0.547 & 0.560 & 0.544 & 0.553 & 0.557 & 0.551 & 0.536 & 0.486 & \cellcolor{gray!10}0.570 \\
& & wave     & 0.504 & 0.493 & 0.504 & 0.491 & 0.495 & 0.504 & 0.493 & 0.479 & 0.514 & \cellcolor{gray!10}0.518 \\
& & nonlin   & 0.518 & 0.542 & 0.525 & 0.537 & 0.527 & 0.530 & 0.496 & 0.528 & 0.502 & \cellcolor{gray!10}0.535 \\
& & freq     & 0.432 & 0.443 & 0.438 & 0.452 & 0.449 & 0.440 & 0.461 & 0.449 & 0.474 & \cellcolor{gray!10}0.469 \\
& & nonstat  & 0.568 & 0.574 & 0.580 & 0.575 & 0.575 & 0.583 & 0.561 & 0.568 & 0.503 & \cellcolor{gray!10}0.593 \\
\arrayrulecolor{black}\midrule

\multirow{5}{*}{4-mod} & \multirow{5}{2.6cm}{\raggedright C\_EDA+\allowbreak \,C\_ACC+\allowbreak \newline \,W\_EDA+\allowbreak \,W\_ACC} & miss     & 0.502 & 0.499 & 0.509 & 0.498 & 0.505 & 0.511 & 0.492 & 0.499 & 0.399 & \cellcolor{gray!10}0.514 \\
& & wave     & 0.523 & 0.525 & 0.525 & 0.509 & 0.522 & 0.530 & 0.501 & 0.517 & 0.457 & \cellcolor{gray!10}0.526 \\
& & nonlin   & 0.506 & 0.503 & 0.506 & 0.507 & 0.504 & 0.511 & 0.490 & 0.500 & 0.447 & \cellcolor{gray!10}0.520 \\
& & freq     & 0.515 & 0.518 & 0.520 & 0.513 & 0.516 & 0.519 & 0.504 & 0.503 & 0.445 & \cellcolor{gray!10}0.530 \\
& & nonstat  & 0.483 & 0.491 & 0.487 & 0.482 & 0.480 & 0.487 & 0.461 & 0.480 & 0.413 & \cellcolor{gray!10}0.511 \\
\arrayrulecolor{black!20}\midrule
& \multirow{5}{2.6cm}{\raggedright W\_ACC+\allowbreak \,W\_EDA+\allowbreak \newline \,W\_TEMP+\allowbreak \,W\_BVP} & miss     & 0.538 & 0.544 & 0.548 & 0.536 & 0.536 & 0.546 & 0.524 & 0.521 & 0.452 & \cellcolor{gray!10}0.547 \\
& & wave     & 0.542 & 0.542 & 0.543 & 0.535 & 0.540 & 0.544 & 0.515 & 0.530 & 0.490 & \cellcolor{gray!10}0.539 \\
& & nonlin   & 0.511 & 0.491 & 0.501 & 0.492 & 0.490 & 0.497 & 0.485 & 0.489 & 0.483 & \cellcolor{gray!10}0.522 \\
& & freq     & 0.535 & 0.541 & 0.540 & 0.533 & 0.533 & 0.544 & 0.523 & 0.525 & 0.474 & \cellcolor{gray!10}0.548 \\
& & nonstat  & 0.515 & 0.523 & 0.523 & 0.519 & 0.515 & 0.525 & 0.500 & 0.508 & 0.450 & \cellcolor{gray!10}0.532 \\
\arrayrulecolor{black!20}\midrule
& \multirow{5}{2.6cm}{\raggedright C\_ECG+\allowbreak \,C\_EMG+\allowbreak \newline \,C\_RESP+\allowbreak \,W\_TEMP} & miss     & 0.553 & 0.555 & 0.559 & 0.554 & 0.552 & 0.559 & 0.551 & 0.542 & 0.499 & \cellcolor{gray!10}0.582 \\
& & wave     & 0.501 & 0.503 & 0.499 & 0.503 & 0.500 & 0.512 & 0.500 & 0.482 & 0.501 & \cellcolor{gray!10}0.528 \\
& & nonlin   & 0.512 & 0.537 & 0.520 & 0.536 & 0.518 & 0.529 & 0.486 & 0.532 & 0.500 & \cellcolor{gray!10}0.533 \\
& & freq     & 0.433 & 0.440 & 0.438 & 0.452 & 0.449 & 0.443 & 0.460 & 0.452 & 0.473 & \cellcolor{gray!10}0.469 \\
& & nonstat  & 0.563 & 0.569 & 0.570 & 0.563 & 0.569 & 0.571 & 0.553 & 0.570 & 0.503 & \cellcolor{gray!10}0.575 \\
\arrayrulecolor{black}\midrule
\multirow{5}{*}{5-mod} & \multirow{5}{*}{all} & miss     & 0.563 & 0.554 & 0.563 & 0.555 & 0.556 & 0.564 & 0.548 & 0.547 & 0.509 & \cellcolor{gray!10}0.578 \\
& & wave     & 0.506 & 0.494 & 0.504 & 0.490 & 0.495 & 0.507 & 0.501 & 0.480 & 0.498 & \cellcolor{gray!10}0.516 \\
& & nonlin   & 0.529 & 0.541 & 0.538 & 0.536 & 0.525 & 0.536 & 0.496 & 0.530 & 0.522 & \cellcolor{gray!10}0.533 \\
& & freq     & 0.434 & 0.442 & 0.440 & 0.452 & 0.446 & 0.442 & 0.457 & 0.451 & 0.472 & \cellcolor{gray!10}0.465 \\
& & nonstat  & 0.570 & 0.578 & 0.573 & 0.570 & 0.571 & 0.581 & 0.560 & 0.571 & 0.501 & \cellcolor{gray!10}0.586 \\
\arrayrulecolor{black}\bottomrule

\end{tabular}
}
\end{table}

\begin{table}[!htbp]
\centering
\caption{Full per-modality-subset, per-noise F1-score ($\uparrow$) on \textbf{WESAD} with the \textbf{Flex-MoE} backbone. Each cell is the F1 for one specific (corrupted modality subset, noise type) combination. Noise abbreviations: \emph{miss}=discrete missingness, \emph{wave}=waveform anomaly, \emph{nonlin}=nonlinear degradation, \emph{freq}=frequency attenuation, \emph{nonstat}=nonstationary noise. Clean baseline: F1 = 0.554.}
\label{tab:appendix-wesad-flexmoe-fullgrid}
\resizebox{1.0\textwidth}{!}{%
\begin{tabular}{@{}lllcccccccccc@{}}
\toprule
\textbf{Card.} & \textbf{Subset} & \textbf{Noise}
& Static & TENT & EATA & READ & SuMi & SoTTA & AeO & PTA & PSDNorm & \cellcolor{gray!10}\textbf{Ours} \\
\midrule

\multirow{5}{*}{1-mod} & \multirow{5}{*}{chest\_EDA} & miss     & 0.519 & 0.536 & 0.535 & 0.515 & 0.534 & 0.556 & 0.516 & 0.509 & 0.492 & \cellcolor{gray!10}0.520 \\
& & wave     & 0.515 & 0.534 & 0.540 & 0.511 & 0.530 & 0.549 & 0.501 & 0.512 & 0.506 & \cellcolor{gray!10}0.522 \\
& & nonlin   & 0.505 & 0.530 & 0.524 & 0.544 & 0.504 & 0.517 & 0.500 & 0.489 & 0.484 & \cellcolor{gray!10}0.519 \\
& & freq     & 0.532 & 0.539 & 0.554 & 0.561 & 0.525 & 0.553 & 0.492 & 0.502 & 0.505 & \cellcolor{gray!10}0.543 \\
& & nonstat  & 0.515 & 0.538 & 0.538 & 0.526 & 0.523 & 0.538 & 0.482 & 0.509 & 0.495 & \cellcolor{gray!10}0.524 \\
\arrayrulecolor{black!20}\midrule
& \multirow{5}{*}{chest\_ACC} & miss     & 0.526 & 0.543 & 0.555 & 0.550 & 0.501 & 0.499 & 0.506 & 0.525 & 0.540 & \cellcolor{gray!10}0.550 \\
& & wave     & 0.545 & 0.559 & 0.547 & 0.549 & 0.524 & 0.522 & 0.521 & 0.530 & 0.539 & \cellcolor{gray!10}0.558 \\
& & nonlin   & 0.554 & 0.578 & 0.567 & 0.543 & 0.517 & 0.524 & 0.500 & 0.514 & 0.544 & \cellcolor{gray!10}0.557 \\
& & freq     & 0.544 & 0.570 & 0.569 & 0.573 & 0.540 & 0.538 & 0.529 & 0.529 & 0.537 & \cellcolor{gray!10}0.559 \\
& & nonstat  & 0.535 & 0.552 & 0.552 & 0.539 & 0.519 & 0.521 & 0.508 & 0.504 & 0.535 & \cellcolor{gray!10}0.556 \\
\arrayrulecolor{black!20}\midrule
& \multirow{5}{*}{wrist\_ACC} & miss     & 0.534 & 0.544 & 0.537 & 0.552 & 0.502 & 0.511 & 0.488 & 0.515 & 0.443 & \cellcolor{gray!10}0.531 \\
& & wave     & 0.529 & 0.535 & 0.536 & 0.534 & 0.510 & 0.527 & 0.483 & 0.509 & 0.514 & \cellcolor{gray!10}0.534 \\
& & nonlin   & 0.528 & 0.540 & 0.547 & 0.544 & 0.554 & 0.570 & 0.532 & 0.519 & 0.519 & \cellcolor{gray!10}0.540 \\
& & freq     & 0.531 & 0.543 & 0.542 & 0.538 & 0.482 & 0.499 & 0.482 & 0.496 & 0.508 & \cellcolor{gray!10}0.544 \\
& & nonstat  & 0.462 & 0.464 & 0.474 & 0.474 & 0.447 & 0.449 & 0.456 & 0.437 & 0.441 & \cellcolor{gray!10}0.469 \\
\arrayrulecolor{black!20}\midrule
& \multirow{5}{*}{wrist\_EDA} & miss     & 0.557 & 0.575 & 0.577 & 0.600 & 0.516 & 0.515 & 0.486 & 0.516 & 0.532 & \cellcolor{gray!10}0.571 \\
& & wave     & 0.543 & 0.550 & 0.555 & 0.585 & 0.474 & 0.498 & 0.462 & 0.491 & 0.541 & \cellcolor{gray!10}0.558 \\
& & nonlin   & 0.544 & 0.541 & 0.542 & 0.561 & 0.499 & 0.516 & 0.469 & 0.500 & 0.526 & \cellcolor{gray!10}0.544 \\
& & freq     & 0.554 & 0.568 & 0.566 & 0.555 & 0.524 & 0.534 & 0.504 & 0.519 & 0.537 & \cellcolor{gray!10}0.567 \\
& & nonstat  & 0.555 & 0.566 & 0.570 & 0.581 & 0.493 & 0.498 & 0.457 & 0.491 & 0.535 & \cellcolor{gray!10}0.568 \\
\arrayrulecolor{black!20}\midrule
& \multirow{5}{*}{chest\_ECG} & miss     & 0.552 & 0.582 & 0.576 & 0.543 & 0.545 & 0.559 & 0.523 & 0.526 & 0.529 & \cellcolor{gray!10}0.561 \\
& & wave     & 0.543 & 0.559 & 0.569 & 0.565 & 0.504 & 0.521 & 0.480 & 0.493 & 0.548 & \cellcolor{gray!10}0.558 \\
& & nonlin   & 0.528 & 0.531 & 0.530 & 0.524 & 0.542 & 0.541 & 0.538 & 0.510 & 0.501 & \cellcolor{gray!10}0.521 \\
& & freq     & 0.476 & 0.470 & 0.457 & 0.479 & 0.516 & 0.498 & 0.499 & 0.512 & 0.519 & \cellcolor{gray!10}0.469 \\
& & nonstat  & 0.571 & 0.602 & 0.598 & 0.563 & 0.578 & 0.607 & 0.546 & 0.550 & 0.529 & \cellcolor{gray!10}0.584 \\
\arrayrulecolor{black!20}\midrule
& \multirow{5}{*}{chest\_EMG} & miss     & 0.554 & 0.576 & 0.570 & 0.552 & 0.525 & 0.519 & 0.508 & 0.526 & 0.539 & \cellcolor{gray!10}0.558 \\
& & wave     & 0.533 & 0.548 & 0.554 & 0.534 & 0.504 & 0.530 & 0.489 & 0.495 & 0.508 & \cellcolor{gray!10}0.548 \\
& & nonlin   & 0.550 & 0.567 & 0.574 & 0.565 & 0.515 & 0.525 & 0.495 & 0.517 & 0.533 & \cellcolor{gray!10}0.561 \\
& & freq     & 0.546 & 0.564 & 0.561 & 0.570 & 0.512 & 0.509 & 0.485 & 0.519 & 0.536 & \cellcolor{gray!10}0.551 \\
& & nonstat  & 0.551 & 0.578 & 0.566 & 0.560 & 0.520 & 0.517 & 0.512 & 0.517 & 0.536 & \cellcolor{gray!10}0.563 \\
\arrayrulecolor{black!20}\midrule
& \multirow{5}{*}{chest\_RESP} & miss     & 0.554 & 0.576 & 0.582 & 0.547 & 0.522 & 0.532 & 0.500 & 0.514 & 0.535 & \cellcolor{gray!10}0.568 \\
& & wave     & 0.554 & 0.569 & 0.565 & 0.554 & 0.526 & 0.530 & 0.499 & 0.511 & 0.533 & \cellcolor{gray!10}0.560 \\
& & nonlin   & 0.539 & 0.541 & 0.550 & 0.534 & 0.564 & 0.575 & 0.535 & 0.528 & 0.519 & \cellcolor{gray!10}0.532 \\
& & freq     & 0.552 & 0.572 & 0.560 & 0.557 & 0.519 & 0.536 & 0.496 & 0.514 & 0.534 & \cellcolor{gray!10}0.559 \\
& & nonstat  & 0.550 & 0.574 & 0.570 & 0.563 & 0.527 & 0.526 & 0.491 & 0.518 & 0.533 & \cellcolor{gray!10}0.562 \\
\arrayrulecolor{black!20}\midrule
& \multirow{5}{*}{wrist\_TEMP} & miss     & 0.551 & 0.580 & 0.588 & 0.537 & 0.534 & 0.543 & 0.488 & 0.524 & 0.528 & \cellcolor{gray!10}0.564 \\
& & wave     & 0.554 & 0.573 & 0.591 & 0.545 & 0.531 & 0.544 & 0.475 & 0.520 & 0.529 & \cellcolor{gray!10}0.572 \\
& & nonlin   & 0.550 & 0.571 & 0.578 & 0.554 & 0.523 & 0.539 & 0.492 & 0.523 & 0.535 & \cellcolor{gray!10}0.567 \\
& & freq     & 0.554 & 0.573 & 0.571 & 0.543 & 0.524 & 0.543 & 0.506 & 0.519 & 0.537 & \cellcolor{gray!10}0.562 \\
& & nonstat  & 0.553 & 0.582 & 0.567 & 0.559 & 0.521 & 0.536 & 0.502 & 0.514 & 0.533 & \cellcolor{gray!10}0.569 \\
\arrayrulecolor{black!20}\midrule
& \multirow{5}{*}{wrist\_BVP} & miss     & 0.554 & 0.569 & 0.566 & 0.527 & 0.528 & 0.524 & 0.512 & 0.525 & 0.530 & \cellcolor{gray!10}0.562 \\
& & wave     & 0.557 & 0.585 & 0.573 & 0.581 & 0.524 & 0.531 & 0.491 & 0.516 & 0.531 & \cellcolor{gray!10}0.566 \\
& & nonlin   & 0.477 & 0.494 & 0.487 & 0.503 & 0.503 & 0.527 & 0.472 & 0.488 & 0.498 & \cellcolor{gray!10}0.488 \\
& & freq     & 0.551 & 0.563 & 0.573 & 0.555 & 0.523 & 0.547 & 0.494 & 0.519 & 0.535 & \cellcolor{gray!10}0.559 \\
& & nonstat  & 0.549 & 0.565 & 0.561 & 0.550 & 0.515 & 0.533 & 0.504 & 0.507 & 0.528 & \cellcolor{gray!10}0.554 \\
\arrayrulecolor{black}\midrule
\multirow{5}{*}{2-mod} & \multirow{5}{*}{C\_EDA+\,C\_ACC} & miss     & 0.519 & 0.519 & 0.520 & 0.545 & 0.531 & 0.549 & 0.507 & 0.529 & 0.508 & \cellcolor{gray!10}0.522 \\
& & wave     & 0.524 & 0.549 & 0.545 & 0.520 & 0.512 & 0.509 & 0.492 & 0.483 & 0.513 & \cellcolor{gray!10}0.538 \\
& & nonlin   & 0.516 & 0.547 & 0.550 & 0.529 & 0.518 & 0.533 & 0.503 & 0.498 & 0.507 & \cellcolor{gray!10}0.541 \\
& & freq     & 0.542 & 0.561 & 0.563 & 0.568 & 0.552 & 0.553 & 0.546 & 0.523 & 0.518 & \cellcolor{gray!10}0.561 \\
& & nonstat  & 0.496 & 0.535 & 0.534 & 0.494 & 0.486 & 0.500 & 0.459 & 0.483 & 0.510 & \cellcolor{gray!10}0.526 \\
\arrayrulecolor{black!20}\midrule
& \multirow{5}{*}{W\_EDA+\,W\_ACC} & miss     & 0.537 & 0.532 & 0.533 & 0.601 & 0.504 & 0.499 & 0.472 & 0.511 & 0.446 & \cellcolor{gray!10}0.535 \\
& & wave     & 0.531 & 0.529 & 0.531 & 0.565 & 0.484 & 0.500 & 0.450 & 0.491 & 0.516 & \cellcolor{gray!10}0.525 \\
& & nonlin   & 0.531 & 0.561 & 0.565 & 0.558 & 0.566 & 0.572 & 0.546 & 0.536 & 0.510 & \cellcolor{gray!10}0.548 \\
& & freq     & 0.529 & 0.549 & 0.545 & 0.552 & 0.488 & 0.500 & 0.491 & 0.493 & 0.505 & \cellcolor{gray!10}0.542 \\
& & nonstat  & 0.476 & 0.485 & 0.485 & 0.464 & 0.448 & 0.441 & 0.437 & 0.445 & 0.449 & \cellcolor{gray!10}0.484 \\
\arrayrulecolor{black!20}\midrule
& \multirow{5}{*}{C\_ACC+\,W\_ACC} & miss     & 0.501 & 0.482 & 0.478 & 0.508 & 0.485 & 0.468 & 0.494 & 0.516 & 0.455 & \cellcolor{gray!10}0.489 \\
& & wave     & 0.507 & 0.504 & 0.505 & 0.504 & 0.515 & 0.519 & 0.498 & 0.506 & 0.494 & \cellcolor{gray!10}0.512 \\
& & nonlin   & 0.531 & 0.554 & 0.563 & 0.525 & 0.567 & 0.582 & 0.535 & 0.538 & 0.518 & \cellcolor{gray!10}0.547 \\
& & freq     & 0.526 & 0.536 & 0.536 & 0.539 & 0.547 & 0.563 & 0.538 & 0.502 & 0.512 & \cellcolor{gray!10}0.525 \\
& & nonstat  & 0.443 & 0.455 & 0.461 & 0.461 & 0.442 & 0.439 & 0.444 & 0.432 & 0.446 & \cellcolor{gray!10}0.448 \\
\arrayrulecolor{black!20}\midrule
& \multirow{5}{*}{C\_ECG+\,C\_EMG} & miss     & 0.558 & 0.589 & 0.586 & 0.552 & 0.531 & 0.547 & 0.515 & 0.518 & 0.537 & \cellcolor{gray!10}0.562 \\
& & wave     & 0.515 & 0.522 & 0.520 & 0.481 & 0.509 & 0.530 & 0.506 & 0.472 & 0.486 & \cellcolor{gray!10}0.524 \\
& & nonlin   & 0.534 & 0.538 & 0.539 & 0.536 & 0.549 & 0.546 & 0.546 & 0.521 & 0.504 & \cellcolor{gray!10}0.539 \\
& & freq     & 0.479 & 0.475 & 0.472 & 0.492 & 0.506 & 0.503 & 0.493 & 0.487 & 0.516 & \cellcolor{gray!10}0.476 \\
& & nonstat  & 0.580 & 0.600 & 0.603 & 0.589 & 0.573 & 0.601 & 0.525 & 0.545 & 0.551 & \cellcolor{gray!10}0.591 \\
\arrayrulecolor{black}\bottomrule

\end{tabular}
}
\end{table}

\clearpage
\begin{table}[!t]\ContinuedFloat
\centering
% \caption{Full per-modality-subset, per-noise F1-score ($\uparrow$) on \textbf{WESAD} with the \textbf{FlexMoE} backbone. Each cell is the F1 for one specific (corrupted modality subset, noise type) combination. Noise abbreviations: \emph{miss}=discrete missingness, \emph{wave}=waveform anomaly, \emph{nonlin}=nonlinear degradation, \emph{freq}=frequency attenuation, \emph{nonstat}=nonstationary noise. Clean baseline: F1 = 0.5542.}
\label{tab:appendix-wesad-flexmoe-fullgrid-cont}
\resizebox{1.0\textwidth}{!}{%
\begin{tabular}{@{}lllcccccccccc@{}}
\toprule
\textbf{Card.} & \textbf{Subset} & \textbf{Noise}
& Static & TENT & EATA & READ & SuMi & SoTTA & AeO & PTA & PSDNorm & \cellcolor{gray!10}\textbf{Ours} \\
\midrule

\multirow{5}{*}{3-mod} & \multirow{5}{2.6cm}{\raggedright C\_ACC+\allowbreak \,W\_ACC+\allowbreak \newline \,W\_EDA} & miss     & 0.510 & 0.502 & 0.493 & 0.503 & 0.489 & 0.474 & 0.493 & 0.516 & 0.455 & \cellcolor{gray!10}0.501 \\
& & wave     & 0.522 & 0.516 & 0.522 & 0.508 & 0.516 & 0.504 & 0.530 & 0.520 & 0.511 & \cellcolor{gray!10}0.522 \\
& & nonlin   & 0.539 & 0.564 & 0.553 & 0.518 & 0.490 & 0.501 & 0.465 & 0.489 & 0.504 & \cellcolor{gray!10}0.552 \\
& & freq     & 0.526 & 0.530 & 0.527 & 0.549 & 0.547 & 0.568 & 0.537 & 0.516 & 0.510 & \cellcolor{gray!10}0.526 \\
& & nonstat  & 0.464 & 0.473 & 0.477 & 0.455 & 0.448 & 0.448 & 0.441 & 0.427 & 0.445 & \cellcolor{gray!10}0.476 \\
\arrayrulecolor{black!20}\midrule
& \multirow{5}{2.6cm}{\raggedright C\_ECG+\allowbreak \,C\_EMG+\allowbreak \newline \,C\_RESP} & miss     & 0.548 & 0.574 & 0.589 & 0.555 & 0.523 & 0.544 & 0.480 & 0.508 & 0.527 & \cellcolor{gray!10}0.571 \\
& & wave     & 0.514 & 0.525 & 0.539 & 0.509 & 0.489 & 0.511 & 0.474 & 0.464 & 0.485 & \cellcolor{gray!10}0.522 \\
& & nonlin   & 0.525 & 0.526 & 0.527 & 0.525 & 0.524 & 0.524 & 0.546 & 0.513 & 0.489 & \cellcolor{gray!10}0.526 \\
& & freq     & 0.477 & 0.468 & 0.473 & 0.479 & 0.511 & 0.501 & 0.494 & 0.490 & 0.518 & \cellcolor{gray!10}0.481 \\
& & nonstat  & 0.566 & 0.599 & 0.597 & 0.573 & 0.575 & 0.587 & 0.533 & 0.541 & 0.528 & \cellcolor{gray!10}0.577 \\
\arrayrulecolor{black}\midrule
\multirow{5}{*}{4-mod} & \multirow{5}{2.6cm}{\raggedright C\_EDA+\allowbreak \,C\_ACC+\allowbreak \newline \,W\_EDA+\allowbreak \,W\_ACC} & miss     & 0.483 & 0.493 & 0.486 & 0.477 & 0.410 & 0.376 & 0.413 & 0.458 & 0.410 & \cellcolor{gray!10}0.483 \\
& & wave     & 0.499 & 0.500 & 0.495 & 0.489 & 0.416 & 0.394 & 0.428 & 0.452 & 0.500 & \cellcolor{gray!10}0.498 \\
& & nonlin   & 0.500 & 0.530 & 0.526 & 0.498 & 0.521 & 0.515 & 0.473 & 0.507 & 0.478 & \cellcolor{gray!10}0.523 \\
& & freq     & 0.519 & 0.537 & 0.534 & 0.535 & 0.527 & 0.532 & 0.520 & 0.507 & 0.492 & \cellcolor{gray!10}0.538 \\
& & nonstat  & 0.423 & 0.448 & 0.438 & 0.457 & 0.377 & 0.356 & 0.397 & 0.383 & 0.410 & \cellcolor{gray!10}0.444 \\
\arrayrulecolor{black!20}\midrule
& \multirow{5}{2.6cm}{\raggedright W\_ACC+\allowbreak \,W\_EDA+\allowbreak \newline \,W\_TEMP+\allowbreak \,W\_BVP} & miss     & 0.546 & 0.571 & 0.568 & 0.593 & 0.504 & 0.540 & 0.453 & 0.476 & 0.447 & \cellcolor{gray!10}0.569 \\
& & wave     & 0.537 & 0.564 & 0.558 & 0.560 & 0.482 & 0.527 & 0.458 & 0.468 & 0.518 & \cellcolor{gray!10}0.557 \\
& & nonlin   & 0.454 & 0.465 & 0.459 & 0.439 & 0.472 & 0.469 & 0.454 & 0.472 & 0.483 & \cellcolor{gray!10}0.459 \\
& & freq     & 0.534 & 0.536 & 0.537 & 0.548 & 0.490 & 0.483 & 0.473 & 0.499 & 0.500 & \cellcolor{gray!10}0.532 \\
& & nonstat  & 0.477 & 0.487 & 0.490 & 0.471 & 0.448 & 0.451 & 0.457 & 0.438 & 0.441 & \cellcolor{gray!10}0.483 \\
\arrayrulecolor{black!20}\midrule
& \multirow{5}{2.6cm}{\raggedright C\_ECG+\allowbreak \,C\_EMG+\allowbreak \newline \,C\_RESP+\allowbreak \,W\_TEMP} & miss     & 0.552 & 0.583 & 0.584 & 0.518 & 0.521 & 0.538 & 0.467 & 0.508 & 0.530 & \cellcolor{gray!10}0.573 \\
& & wave     & 0.508 & 0.522 & 0.538 & 0.499 & 0.537 & 0.550 & 0.504 & 0.497 & 0.464 & \cellcolor{gray!10}0.518 \\
& & nonlin   & 0.514 & 0.516 & 0.517 & 0.505 & 0.525 & 0.523 & 0.533 & 0.489 & 0.484 & \cellcolor{gray!10}0.520 \\
& & freq     & 0.478 & 0.470 & 0.474 & 0.481 & 0.508 & 0.507 & 0.492 & 0.493 & 0.520 & \cellcolor{gray!10}0.480 \\
& & nonstat  & 0.575 & 0.603 & 0.603 & 0.586 & 0.573 & 0.582 & 0.534 & 0.532 & 0.544 & \cellcolor{gray!10}0.588 \\
\arrayrulecolor{black}\midrule
\multirow{5}{*}{5-mod} & \multirow{5}{*}{all} & miss     & 0.543 & 0.582 & 0.581 & 0.497 & 0.517 & 0.537 & 0.473 & 0.505 & 0.527 & \cellcolor{gray!10}0.565 \\
& & wave     & 0.503 & 0.516 & 0.530 & 0.483 & 0.508 & 0.523 & 0.471 & 0.470 & 0.449 & \cellcolor{gray!10}0.518 \\
& & nonlin   & 0.543 & 0.541 & 0.545 & 0.519 & 0.470 & 0.455 & 0.496 & 0.480 & 0.501 & \cellcolor{gray!10}0.539 \\
& & freq     & 0.480 & 0.474 & 0.469 & 0.474 & 0.507 & 0.497 & 0.493 & 0.493 & 0.518 & \cellcolor{gray!10}0.478 \\
& & nonstat  & 0.566 & 0.572 & 0.579 & 0.572 & 0.546 & 0.580 & 0.511 & 0.528 & 0.522 & \cellcolor{gray!10}0.569 \\
\arrayrulecolor{black}\bottomrule

\end{tabular}
}
\end{table}

\section{Complete Efficiency Metrics}\label{app:efficiency}

% \yue{Add all the efficiency numbers here. We can just put it like a table, no need to put performance again. Just write a few lines on how you measure each metric, the scenario for each dataset and the numbers.}

We report the complete efficiency measurements in Table~\ref{tab:efficiency_dsads} and Table~\ref{tab:efficiency_dalia_pamap2}. All measurements are conducted on corrupted test batches with batch size $64$. Latency is the average wall-clock time per batch after warm-up, including test-time adaptation and inference. For detector-based methods, the detector is executed as part of the method, so its overhead is included. Energy is measured as the average CPU energy per batch in Joules. GFLOPs are measured using the PyTorch profiler and averaged over test batches. Trainable parameters denote the number of parameters updated during test-time adaptation; non-parametric methods have zero trainable parameters.

For DSADS, the efficiency benchmark corrupts the torso, right-arm, and left-arm modalities with \texttt{waveform\_anomaly} noise. For DaliaHAR, the benchmark corrupts all wrist modalities. For PAMAP2, the benchmark corrupts the ankle IMU, hand IMU, and heart-rate modalities.

\begin{table*}[t]
\centering
\small
\caption{Efficiency metrics on DSADS.}
\label{tab:efficiency_dsads}
\begin{tabular}{llrrrr}
\toprule
Backbone & Method & Latency (ms) & Energy (J) & GFLOPs & Trainable Params \\
\midrule
\multicolumn{6}{l}{\textbf{MAESTRO}} \\
 & Static & 435.0 & 381.1 & 5.56 & 0 \\
 & TENT & 1200.2 & 1254.5 & 53.63 & 3584 \\
 & EATA & 1038.6 & 1108.4 & 53.63 & 3584 \\
 & READ & 554.8 & 528.6 & 12.59 & 49920 \\
 & SuMi & 1238.4 & 1319.8 & 53.36 & 3584 \\
 & SoTTA & 1349.9 & 1465.7 & 45.55 & 3584 \\
 & AeO & 4878.5 & 5651.6 & 299.49 & 3584 \\
 & PTA & 5190.5 & 5991.9 & 293.93 & 3584 \\
 & PSDNorm & 1073.1 & 915.1 & 5.99 & 0 \\
 & \method{} & 1713.2 & 1692.7 & 48.54 & 1539.0 \\
\midrule
\multicolumn{6}{l}{\textbf{ShaSpec}} \\
 & Static & 464.1 & 414.0 & 52.03 & 0 \\
 & TENT & 1799.2 & 2098.0 & 826.82 & 8960 \\
 & EATA & 1124.4 & 1373.1 & 826.82 & 8960 \\
 & READ & 494.4 & 478.1 & 144.18 & 119192 \\
 & SuMi & 764.0 & 820.2 & 454.49 & 8960 \\
 & SoTTA & 565.5 & 552.1 & 147.48 & 8960 \\
 & AeO & 5987.8 & 7936.3 & 4672.61 & 8960 \\
 & PTA & 5312.6 & 7142.5 & 4600.54 & 8960 \\
 & PSDNorm & 1026.3 & 863.8 & 52.47 & 0 \\
 & \method{} & 1382.9 & 1434.9 & 563.22 & 4355.4 \\
\midrule
\multicolumn{6}{l}{\textbf{Flex-MoE}} \\
 & Static & 396.5 & 336.6 & 27.68 & 0 \\
 & TENT & 1312.3 & 1348.7 & 431.79 & 6912 \\
 & EATA & 1357.5 & 1386.2 & 431.79 & 6912 \\
 & READ & 565.8 & 517.7 & 77.10 & 49152 \\
 & SuMi & 1080.2 & 1065.5 & 299.81 & 6912 \\
 & SoTTA & 2164.2 & 2122.8 & 225.17 & 6912 \\
 & AeO & 8049.2 & 8497.9 & 2440.03 & 6912 \\
 & PTA & 9764.2 & 10194.9 & 2402.35 & 6912 \\
 & PSDNorm & 1081.8 & 904.5 & 28.12 & 0 \\
 & \method{} & 2191.3 & 2147.9 & 365.35 & 4867.4 \\
\bottomrule
\end{tabular}
\end{table*}

\begin{table*}[t]
\centering
\small
\caption{Efficiency metrics on DaliaHAR and PAMAP2 with MAESTRO.}
\label{tab:efficiency_dalia_pamap2}
\begin{tabular}{llrrrr}
\toprule
Dataset & Method & Latency (ms) & Energy (J) & GFLOPs & Trainable Params \\
\midrule
\multicolumn{6}{l}{\textbf{DaliaHAR / MAESTRO}} \\
 & Static & 141.2 & 147.8 & 1.44 & 0 \\
 & TENT & 508.3 & 597.9 & 13.84 & 1536 \\
 & EATA & 480.7 & 554.4 & 13.84 & 1536 \\
 & READ & 177.4 & 184.8 & 3.20 & 6336 \\
 & SuMi & 521.4 & 560.2 & 12.22 & 1536 \\
 & SoTTA & 662.0 & 685.3 & 6.79 & 1536 \\
 & AeO & 2392.8 & 2911.9 & 77.26 & 1536 \\
 & PTA & 3061.1 & 3627.8 & 75.81 & 1536 \\
 & PSDNorm & 264.4 & 243.0 & 1.63 & 0 \\
 & \method{} & 693.3 & 761.6 & 13.11 & 971.8 \\
\midrule
\multicolumn{6}{l}{\textbf{PAMAP2 / MAESTRO}} \\
 & Static & 915.1 & 856.7 & 19.39 & 0 \\
 & TENT & 3161.8 & 3447.7 & 185.67 & 3072 \\
 & EATA & 2262.1 & 2521.3 & 185.67 & 3072 \\
 & READ & 1223.6 & 1228.7 & 43.92 & 49920 \\
 & SuMi & 3016.0 & 3257.3 & 147.53 & 3072 \\
 & SoTTA & 3895.2 & 4343.8 & 242.04 & 3072 \\
 & AeO & 9600.4 & 11220.5 & 870.11 & 3072 \\
 & PTA & 7524.0 & 9341.5 & 850.69 & 3072 \\
 & PSDNorm & 1640.1 & 1443.1 & 21.08 & 0 \\
 & \method{} & 3116.1 & 3250.2 & 177.50 & 1603.1 \\
\bottomrule
\end{tabular}
\end{table*}

\subsection{Detector Training and Deployment Footprint}\label{app:detector-footprint}

Table~\ref{tab:detector-footprint} reports the detector footprint for DaliaHAR--MAESTRO. This separate profiling study uses an NVIDIA L40. Detector inference is measured with batch size 64, reporting median latency over 50 synchronized runs after 10 warm-up iterations. Detector training is performed offline; deployment retains the trained detector weights, feature statistics, and frozen reliance profile. These detector-only measurements complement the end-to-end profiles above and should not be directly combined with measurements from different hardware.

\begin{table}[!htbp]
\centering
\caption{\textbf{Detector footprint on DaliaHAR--MAESTRO.} Offline training cost and detector-only inference cost on an NVIDIA L40. Inference measurements are per batch of 64.}
\label{tab:detector-footprint}
\small
\setlength{\tabcolsep}{5pt}
\renewcommand{\arraystretch}{1.05}
\begin{tabular}{@{}lrrrrr@{}}
\toprule
& & \multicolumn{2}{c}{Offline training} & \multicolumn{2}{c}{Detector inference} \\
\cmidrule(lr){3-4}\cmidrule(l){5-6}
Modality & Parameters & Time (s) & GFLOPs & Latency (ms) & GFLOPs \\
\midrule
Wrist acceleration & 8,512 & 35.3 & 97.7 & 33.9 & 0.0032 \\
Wrist BVP          & 8,384 & 42.2 & 32.6 & 12.2 & 0.0011 \\
Wrist EDA          & 8,384 & 42.1 & 32.6 & 12.2 & 0.0011 \\
Wrist temperature  & 8,384 & 42.3 & 32.6 & 12.1 & 0.0011 \\
Chest acceleration & 8,512 & 42.6 & 97.7 & 33.9 & 0.0032 \\
\midrule
\textbf{Reported total} & \textbf{42,176} & \textbf{204.6} & \textbf{293.0} & \textbf{104.5} & \textbf{0.0095} \\
\bottomrule
\end{tabular}

\smallskip
\begin{minipage}{\linewidth}
\footnotesize
Totals retain the reported aggregate measurements; sums of rounded modality-level entries may differ slightly.
\end{minipage}
\end{table}

\textbf{Memory and stored artifacts.} The reported peak CUDA memory is 4.52 MiB per modality when detectors are evaluated sequentially with buffer reuse. Stored artifacts comprise 164.8 KB of detector weights, 16.3 KB of feature statistics, and a 40-byte reliance profile. The memory figure describes the detector computation rather than the complete adaptation pipeline. The following subsection evaluates an implementation optimization separately.

\subsection{Detector Implementation Optimization}
\label{app:detector-optimization}

We assess detector implementation overhead in a separate controlled DSADS--MAESTRO rerun. Reusing the full-channel FITS reconstruction within each batch and batching spectral computations reduces detector latency from 467 to 16\,ms and end-to-end latency from 1,713 to 1,262\,ms, a reduction of approximately 26\%. Both implementations produce identical detector features, suspect flags, and predictions. Table~\ref{tab:detector-optimization} reports this comparison separately from the five reference profiles in Tables~\ref{tab:efficiency_dsads} and~\ref{tab:efficiency_dalia_pamap2}.

\begin{table}[htbp]
\centering
\small
\setlength{\tabcolsep}{5pt}
\renewcommand{\arraystretch}{0.95}
\caption{\textbf{DSADS--MAESTRO detector optimization.} Latency in milliseconds per batch from a controlled rerun. End-to-end latency includes detection, adaptation, and inference.}
\label{tab:detector-optimization}
\begin{tabular}{@{}lrr@{}}
\toprule
Implementation & Detector & End-to-end \\
\midrule
Reference & 467 & 1,713 \\
Optimized & 16 & 1,262 \\
\bottomrule
\end{tabular}
\end{table}

\section{Additional Results}\label{app:res_additional}

\begin{figure}[htbp]
    \centering
    % \captionsetup{font=scriptsize}
    \vspace{-8pt}
    \includegraphics[width=0.9\textwidth]{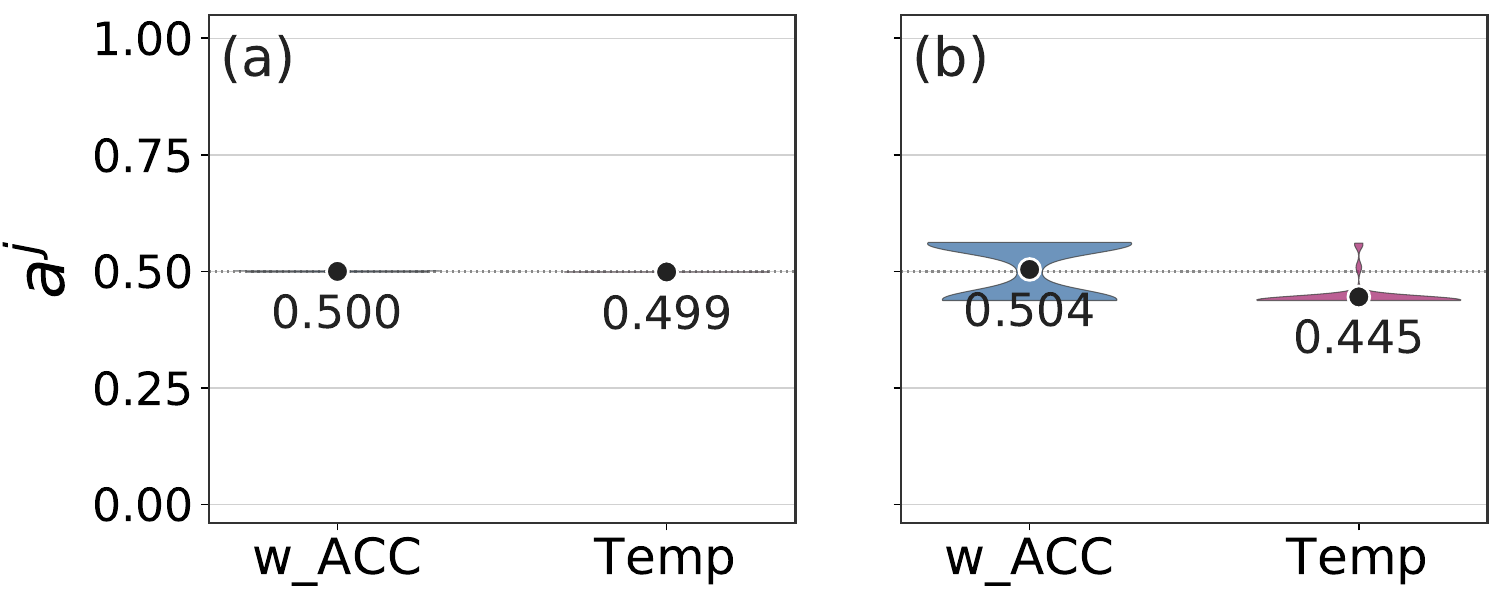}
    \caption{Gate lr sensitivity on DaliaHAR, 
    $\mathcal{S}_i = \{\text{wrist\_ACC}, \text{wrist\_Temp}\}$. 
    (a) $\eta_\alpha = 10^{-3}$ (F1\,$=$\,0.65): gates pinned 
    at $\sim 0.5$. (b) $\eta_\alpha = 5 \times 10^{-2}$ 
    (F1\,$=$\,0.67): gates move within the episodic budget.}
    \vspace{-8pt}
    \label{fig:sensitivity_lr}
\end{figure}

Additional results shown in Figure~\ref{fig:sensitivity_lr}.
% \begin{figure}{r}{0.4\textwidth}
%     \centering
%     \captionsetup{font=scriptsize}
%     \vspace{-10pt}
%     \includegraphics[width=\linewidth]{figures/sensitivity_raw_param_lr.pdf}
%     \caption{\scriptsize Gate lr sensitivity on DaliaHAR, 
%     $\mathcal{S}_i = \{\text{wrist\_ACC}, \text{wrist\_Temp}\}$. 
%     (a) $\eta_\alpha = $ 1e-3 (F1\,$=$\,0.65): gates pinned 
%     at $\sim 0.5$. (b) $\eta_\alpha = $ 5e-2 
%     (F1\,$=$\,0.67): gates move within the episodic budget.}
%     \vspace{-10pt}
%     \label{fig:sensitivity_lr}
% \end{figure}

\begin{table}[!htbp]
\centering
\caption{\small \textbf{Unconditional drop vs.\ \method{} on MAESTRO.} F1-score ($\uparrow$) averaged across 10 corrupted-modality scenarios $\times$ 5 noise types per dataset. Mean$_{\pm\text{std}}$ across all (subset, noise) combinations. \textcolor{burgundy}{\textbf{Bold}}: higher of \{Unconditional drop, \method{}\} per row. We restrict this analysis to MAESTRO, the most stable missingness-resilient backbone in our benchmark, where the routing-vs-dropping mechanism is evaluated without confounding from backbone-specific fragility under modality dropout.}
\label{apptab:detect_drop_full}
\setlength{\tabcolsep}{6pt}
\renewcommand{\arraystretch}{1.2}
\centering
\begin{tabular}{@{}lccc@{}}
\toprule
\textbf{Dataset} & \textbf{Static} & \textbf{Unconditional drop} & \textbf{\method{}} \\
\midrule
DSADS    & 0.69 {\scriptsize $\pm$ 0.22} & 0.67 {\scriptsize $\pm$ 0.22} & \textcolor{burgundy}{\textbf{0.72}} {\scriptsize $\pm$ 0.20} \\
DaliaHAR & 0.74 {\scriptsize $\pm$ 0.07} & 0.67 {\scriptsize $\pm$ 0.11} & \textcolor{burgundy}{\textbf{0.76}} {\scriptsize $\pm$ 0.07} \\
PAMAP2   & 0.92 {\scriptsize $\pm$ 0.04} & 0.68 {\scriptsize $\pm$ 0.36} & \textcolor{burgundy}{\textbf{0.92}} {\scriptsize $\pm$ 0.04} \\
WESAD    & 0.65 {\scriptsize $\pm$ 0.04} & 0.62 {\scriptsize $\pm$ 0.03} & \textcolor{burgundy}{\textbf{0.63}} {\scriptsize $\pm$ 0.04} \\
\bottomrule
\end{tabular}
\end{table}

\section{Statistical and Condition-Level Analyses}
\label{app:effectiveness_stats}

We complement the setting-level averages in Table~\ref{tab:overall-robustness} with paired comparisons, a repeated-corruption experiment, and condition-level rank and dispersion analyses. The benchmark contains 720 matched conditions: 50 per backbone on DSADS and DaliaHAR, 45 on PAMAP2, and 95 on WESAD. Each condition specifies a dataset, backbone, corrupted-modality subset, and corruption family. Methods are paired on the same corrupted data and source checkpoint within each setting. Unless stated otherwise, $\Delta$ denotes \method{} minus the named comparator. Effect sizes and interval bounds below are expressed in \emph{macro-F1 points} (100 times the difference on the 0--1 scale), preserving the original reported precision while displaying two decimals.

\subsection{Paired comparisons over matched corruption conditions}
\label{app:paired-condition-analysis}
% Source: reviewer ed5e response, W1+Q1(3).
% AUTHOR CHECK: the source calls these estimates medians; verify the estimator,
% CI construction and multiplicity convention against the original analysis script.

Table~\ref{tab:paired-condition-tests} reproduces the paired Wilcoxon signed-rank summaries against the baseline with the highest mean F1 in each setting. This setting-specific reference may select different methods in different rows; it complements the fixed-method comparisons in Table~\ref{tab:overall-robustness}. The W/T/L column instead counts the reported outcomes against \emph{all nine alternatives} (Static and eight TTA baselines), summing to nine in each row. Its T category denotes no declared difference in the original analysis, not statistical equivalence or a practical tie band. Across the 108 comparisons, the reported counts are 68 wins, 28 non-rejections, and 12 losses.

Because corruption conditions share examples, checkpoints, and overlapping modality subsets, these condition-level tests are exploratory summaries of the evaluated suites, rather than 720 independent replications. They do not quantify training-seed uncertainty or establish superiority on a population of unseen datasets. The intervals and $p$-values below retain the values reported in the original analysis.

\begin{table}[!htbp]
\centering
\caption{\small \textbf{Paired condition-level comparisons.} $n$ is the number of matched corruption conditions. The reported paired effect and 95\% interval compare \method{} with the named setting-specific comparator. Estimates, intervals, and $p$-values are retained from the original paired analysis. W/T/L counts the reported test outcomes across nine alternatives, not wins across the $n$ conditions. Effect sizes and intervals are in macro-F1 points; positive values favor \method{}.}
\label{tab:paired-condition-tests}
\setlength{\tabcolsep}{3pt}
\renewcommand{\arraystretch}{1.12}
\small
\begin{tabular*}{\textwidth}{@{\extracolsep{\fill}}lrcrcrc@{}}
\toprule
\textbf{Setting} & $n$ & \textbf{Comparator} & \textbf{Reported $\Delta$} & \textbf{Reported 95\% CI} & \textbf{Reported $p$} & \textbf{W/T/L} \\
\midrule
DSADS / MAESTRO & 50 & PTA & $+0.20$ & $[-0.60,\,+1.90]$ & $1.00$ & 5/4/0 \\
DSADS / ShaSpec & 50 & PSDNorm & $+7.00$ & $[+5.30,\,+9.10]$ & $6\!\times\!10^{-8}$ & 9/0/0 \\
DSADS / Flex-MoE & 50 & PTA & $+0.00$ & $[-0.40,\,+0.70]$ & $1.00$ & 2/7/0 \\
\midrule
DaliaHAR / MAESTRO & 50 & EATA & $-0.20$ & $[-0.40,\,-0.10]$ & $0.008$ & 7/0/2 \\
DaliaHAR / ShaSpec & 50 & TENT & $+1.80$ & $[+0.90,\,+3.70]$ & $6\!\times\!10^{-4}$ & 9/0/0 \\
DaliaHAR / Flex-MoE & 50 & EATA & $-0.20$ & $[-0.40,\,+0.00]$ & $0.19$ & 7/2/0 \\
\midrule
PAMAP2 / MAESTRO & 45 & EATA & $-0.20$ & $[-0.40,\,-0.10]$ & $0.14$ & 4/5/0 \\
PAMAP2 / ShaSpec & 45 & PSDNorm & $+0.60$ & $[-0.50,\,+1.70]$ & $0.90$ & 2/7/0 \\
PAMAP2 / Flex-MoE & 45 & TENT & $+0.20$ & $[-0.00,\,+0.50]$ & $0.30$ & 8/1/0 \\
\midrule
WESAD / MAESTRO & 95 & Static & $-1.40$ & $[-1.80,\,-1.10]$ & $6\!\times\!10^{-9}$ & 0/1/8 \\
WESAD / ShaSpec & 95 & SoTTA & $+1.50$ & $[+1.30,\,+1.70]$ & $2\!\times\!10^{-15}$ & 9/0/0 \\
WESAD / Flex-MoE & 95 & EATA & $-0.60$ & $[-0.70,\,-0.40]$ & $5\!\times\!10^{-7}$ & 6/1/2 \\
\bottomrule
\end{tabular*}
\end{table}

The reported positive paired effects are largest on DSADS--ShaSpec ($+7.00$ points), DaliaHAR--ShaSpec ($+1.80$), and WESAD--ShaSpec ($+1.50$). Negative effects remain visible on DaliaHAR--MAESTRO, WESAD--MAESTRO, and WESAD--Flex-MoE. This breakdown is consistent with the concentration of gains by backbone in the main results.
\clearpage
\subsection{Experiment R3.1: repeated corruption realizations}
\label{app:repeated-corruption-r31}
% Source: reviewer ed5e response, W1+Q1(4), New Experiment R3.1.
% AUTHOR CHECK: recover how the five trials were aggregated before calculating
% condition-level differences, confidence intervals and p-values.

We repeated DSADS--MAESTRO over five trials with redrawn corruption realizations for each of the 50 matched conditions, keeping the source checkpoint fixed. The reported mean macro-F1 is 73.80 for \method{} on the 0--100 scale. Table~\ref{tab:repeated-corruption-r31} lists the eight reported comparator summaries; PSDNorm was not included in the reported repeated-trial table. All eight reported paired intervals exclude zero in favor of \method{}. The reported average per-condition standard deviation across trials is 0.60 macro-F1 points. These results are a separate repeated-corruption evaluation and do not replace the single-evaluation means in Table~\ref{tab:overall-robustness}. The intervals concern the reported paired comparison, not uncertainty from five independent source-training runs.

\begin{table}[!htbp]
\centering
\caption{\small \textbf{Experiment R3.1: DSADS--MAESTRO under five corruption redraws.} Mean F1, paired differences and interval bounds use the 0--100 scale. \method{} achieves 73.80 mean F1. W/T/L sums to 50 matched conditions; the estimates, intervals and $p$-values retain the original repeated-evaluation reporting.}
\label{tab:repeated-corruption-r31}
\setlength{\tabcolsep}{4pt}
\renewcommand{\arraystretch}{1.12}
\small
\begin{tabular*}{\textwidth}{@{\extracolsep{\fill}}lcccrc@{}}
\toprule
\textbf{Comparator} & \textbf{Mean F1} & \textbf{W/T/L} & \textbf{Median $\Delta$} & \textbf{Reported 95\% CI} & \textbf{Reported $p$} \\
\midrule
Static & $69.30$ & 36/0/14 & $+2.70$ & $[+1.30,\,+4.80]$ & $6\!\times\!10^{-5}$ \\
TENT & $69.30$ & 36/1/13 & $+2.40$ & $[+1.20,\,+4.40]$ & $9\!\times\!10^{-5}$ \\
EATA & $69.30$ & 34/4/12 & $+2.40$ & $[+1.10,\,+4.50]$ & $9\!\times\!10^{-5}$ \\
READ & $69.00$ & 38/0/12 & $+3.10$ & $[+1.60,\,+5.60]$ & $5\!\times\!10^{-5}$ \\
SuMi & $69.10$ & 37/1/12 & $+3.00$ & $[+1.60,\,+5.30]$ & $6\!\times\!10^{-5}$ \\
SoTTA & $69.10$ & 35/1/14 & $+2.70$ & $[+1.30,\,+4.90]$ & $7\!\times\!10^{-5}$ \\
AeO & $69.10$ & 36/1/13 & $+3.00$ & $[+1.50,\,+5.60]$ & $6\!\times\!10^{-5}$ \\
PTA & $69.30$ & 35/0/15 & $+2.80$ & $[+1.40,\,+5.10]$ & $7\!\times\!10^{-5}$ \\
\bottomrule
\end{tabular*}
\end{table}

\subsection{Condition-level ranks across corruptions, backbones, and datasets}
\label{app:condition-level-ranks}
% Source: reviewer ed5e response Q4 and the 2 August fixed-baseline clarification.

For this analysis, the ten methods are ranked \emph{within each matched corruption condition}, and those ranks are averaged over the relevant group. Average ranks are assigned to tied condition scores; smaller ranks are better. This differs from Table~\ref{tab:overall-robustness}, which first averages F1 within each dataset--backbone setting and then ranks the resulting means. The pooled condition-level average also weights settings by their number of conditions, so WESAD contributes more observations than the other datasets. These rank summaries are descriptive.

\begin{table}[!htbp]
\centering
\caption{\small \textbf{Condition-level rank breakdowns.} Left: \method{} across the five corruption families. Right: \method{} and the lowest-rank alternative within each backbone. The reported means pool the conditions in the indicated group.}
\label{tab:condition-rank-breakdowns}
\small
\setlength{\tabcolsep}{4pt}
\renewcommand{\arraystretch}{1.13}
\begin{minipage}[t]{0.46\textwidth}
\vspace{0pt}\centering
\begin{tabular*}{\linewidth}{@{\extracolsep{\fill}}lr@{}}
\toprule
\textbf{Corruption} & \textbf{Rank} $\downarrow$ \\
\midrule
Waveform anomaly & 3.19 \\
Discrete missingness & 3.25 \\
Nonstationary noise & 3.49 \\
Nonlinear degradation & 4.03 \\
Frequency attenuation & 4.62 \\
\bottomrule
\end{tabular*}
\end{minipage}\hfill
\begin{minipage}[t]{0.51\textwidth}
\vspace{0pt}\centering
\begin{tabular*}{\linewidth}{@{\extracolsep{\fill}}lrl@{}}
\toprule
\textbf{Backbone} & \textbf{Rank} $\downarrow$ & \textbf{Best alternative} \\
\midrule
ShaSpec & 2.18 & Static (4.59) \\
Flex-MoE & 3.59 & TENT (3.42) \\
MAESTRO & 5.38 & EATA (3.85) \\
\bottomrule
\end{tabular*}
\end{minipage}
\end{table}

\method{} has its lowest average rank under waveform anomalies and its highest under frequency attenuation. The backbone breakdown is more uneven: \method{} ranks 2.18 on ShaSpec, 3.59 on Flex-MoE, and 5.38 on MAESTRO. Thus, the condition-level analysis also exposes the backbone dependence visible in the setting-level averages.

\newpage
Table~\ref{tab:dataset-condition-ranks} retains the dataset-level and pooled ranks from the same analysis. \method{} has the lowest average rank on DSADS, DaliaHAR, and PAMAP2; EATA has the lowest on WESAD. The difference between PTA's DSADS and WESAD ranks illustrates variation across sensing setups. Because the datasets also differ in task, sensor count, class distribution, and corruption difficulty, this comparison does not identify modality heterogeneity as the sole cause.

\begin{table}[!htbp]
\centering
\caption{\small \textbf{Mean condition-level ranks by dataset.} Smaller is better. Pooled averages all 720 condition ranks; Worst is the largest of the four dataset-level ranks. These are not the ranks of dataset--backbone means in Table~\ref{tab:overall-robustness}. Bold and underline mark the lowest and second-lowest displayed ranks per column.}
\label{tab:dataset-condition-ranks}
\setlength{\tabcolsep}{5pt}
\renewcommand{\arraystretch}{1.12}
\small
\begin{tabular*}{\textwidth}{@{\extracolsep{\fill}}lrrrrrr@{}}
\toprule
\textbf{Method} & \textbf{DSADS} & \textbf{DaliaHAR} & \textbf{PAMAP2} & \textbf{WESAD} & \textbf{Pooled} & \textbf{Worst} \\
\midrule
\textbf{\method{}} & \textcolor{burgundy}{\textbf{3.56}} & \textcolor{burgundy}{\textbf{2.39}} & \textcolor{burgundy}{\textbf{4.09}} & 4.32 & \textcolor{burgundy}{\textbf{3.72}} & \textcolor{burgundy}{\textbf{4.32}} \\
EATA & 5.30 & 2.68 & 4.47 & \textcolor{burgundy}{\textbf{3.70}} & \underline{3.97} & \underline{5.30} \\
TENT & 5.53 & \underline{2.46} & 4.65 & \underline{4.25} & 4.22 & 5.53 \\
SoTTA & 5.00 & 4.57 & 5.41 & 4.49 & 4.79 & 5.41 \\
Static & 5.51 & 5.96 & 4.60 & 4.89 & 5.19 & 5.96 \\
SuMi & 6.02 & 6.92 & 5.56 & 5.71 & 6.00 & 6.92 \\
PTA & \underline{3.63} & 6.70 & \underline{4.32} & 7.97 & 6.12 & 7.97 \\
READ & 6.66 & 5.74 & 8.21 & 5.29 & 6.22 & 8.21 \\
AeO & 5.61 & 8.77 & 6.30 & 7.12 & 7.00 & 8.77 \\
PSDNorm & 8.17 & 8.80 & 7.40 & 7.25 & 7.79 & 8.80 \\
\bottomrule
\end{tabular*}
\end{table}

\subsection{Dispersion across corruption families}
\label{app:corruption-dispersion}
% Source: reviewer ed5e response, W1+Q1(2).

To separate variation across corruption families from variation across corrupted-modality subsets, we compute the sample standard deviation across the five corruption-family scores within each fixed dataset--backbone--subset scenario, then average these standard deviations. Across the 144 scenarios, the reported average is 6.20 macro-F1 points for \method{}, compared with 7.20--7.70 points for the seven parameter-updating baselines (TENT, EATA, READ, SuMi, SoTTA, AeO, and PTA). The paired Wilcoxon comparisons of these scenario-level deviations reported $p$-values between $3\times10^{-11}$ and $3\times10^{-4}$. The dependence qualifications of Appendix~\ref{app:paired-condition-analysis} also apply to these scenario-level comparisons. PSDNorm has a smaller reported deviation, 5.60 points, alongside a poorer pooled mean rank of 7.79; low dispersion alone therefore does not establish better predictive performance. Table~\ref{tab:corruption-dispersion} lists the four settings highlighted in the analysis. This dispersion describes differences in corruption difficulty and sensitivity, not uncertainty across training seeds.

\begin{table}[!htbp]
\centering
\caption{\small \textbf{Within-subset dispersion across corruption families.} Each entry averages the sample standard deviation over the five corruption scores within a fixed modality-subset scenario. Values are in macro-F1 points. The baseline column gives the reported range across the seven parameter-updating baselines.}
\label{tab:corruption-dispersion}
\setlength{\tabcolsep}{5pt}
\renewcommand{\arraystretch}{1.12}
\small
\begin{tabular*}{\textwidth}{@{\extracolsep{\fill}}lcc@{}}
\toprule
\textbf{Setting} & \textbf{Baseline range} & \textbf{\method{}} \\
\midrule
DSADS / ShaSpec & $18.80$--$21.60$ & $12.30$ \\
PAMAP2 / ShaSpec & $20.20$--$21.90$ & $16.20$ \\
DSADS / MAESTRO & $12.80$--$13.50$ & $10.20$ \\
DSADS / Flex-MoE & $9.90$--$10.40$ & $9.20$ \\
\bottomrule
\end{tabular*}
\end{table}

% APPENDIX: new source-training comparisons and design controls supporting Section 4.2.
% Append after the existing appendix material (and the Section 4.1 statistical appendix).
% Do not duplicate the original all-dataset source-training table or sensitivity figures.
\section{Source-Training Replacements and Additional Design Analyses}\label{app:curriculum_ablation}

\subsection{Alternative source-training regimes}\label{app:curriculum_replacement}

We replace source training for DaliaHAR--ShaSpec with OGM, OPM, or MIAM and evaluate missingness robustness (Table~\ref{tab:curriculum-source-full}) and subsequent TTA (Table~\ref{tab:curriculum-tta-full}). Uniform/asymmetric source results reproduce the F1 entries of Table~\ref{tab:rec_ours}, whose full dataset--backbone comparison remains in Appendix~\ref{app:rec_vs_ours_results}. Asymmetric training achieves the highest F1 at 40\% missingness (0.61); uniform training has the highest complete-input F1 (0.77).

\begin{table}[htbp]
\centering
\captionsetup{font=small,skip=4pt}
\caption{\textbf{Source-training replacement on DaliaHAR--ShaSpec.} Macro-F1 ($\uparrow$), with the reported dispersions where available. Uniform/asymmetric entries use the original three-seed source-training evaluation. MIAM reports means only at 20\%/40\% missingness; -- denotes an unreported result. \textcolor{burgundy}{\textbf{Bold}}/\underline{underline}: best/second distinct displayed mean per column.}
\label{tab:curriculum-source-full}
\begingroup
\fontsize{9}{9.5}\selectfont
\setlength{\tabcolsep}{7pt}
\renewcommand{\arraystretch}{1.08}
\begin{tabular}{@{}lccc@{}}
\toprule
Source training & 0\% missing & 20\% missing & 40\% missing \\
\midrule
Uniform (Original) & \textcolor{burgundy}{\textbf{0.77}}$_{\pm0.02}$ & \underline{0.57}$_{\pm0.02}$ & 0.39$_{\pm0.01}$ \\
OGM & 0.73$_{\pm0.00}$ & 0.50$_{\pm0.01}$ & 0.35$_{\pm0.02}$ \\
OPM & \underline{0.76}$_{\pm0.00}$ & \underline{0.57}$_{\pm0.02}$ & 0.45$_{\pm0.01}$ \\
MIAM & -- & \textcolor{burgundy}{\textbf{0.68}} & \underline{0.59} \\
\midrule
\rowcolor{tolblue!10}
Asymmetric (ours) & 0.74$_{\pm0.00}$ & \textcolor{burgundy}{\textbf{0.68}}$_{\pm0.02}$ & \textcolor{burgundy}{\textbf{0.61}}$_{\pm0.02}$ \\
\bottomrule
\end{tabular}
\endgroup
\end{table}

\method{} leads all OGM/OPM columns in Table~\ref{tab:curriculum-tta-full}, and the MIAM \texttt{all\_wrist}/Div. columns; on MIAM \texttt{w\_acc}, TENT/EATA lead (0.64 vs.\ 0.63). MIAM varies masking rates by batch, so we estimate its prior as the expectation of the checkpoint-specific masking distribution. Our curriculum supplies the terminal dropout profile directly. Across the three curricula, \method{} achieves the highest Div. F1, exceeding the strongest TTA baseline by 1, 2, and 2 points for OGM, OPM, and MIAM, respectively.

\begin{table}[htbp]
\centering
\captionsetup{font=small,skip=4pt}
\caption{\textbf{Complete TTA comparison on replacement checkpoints.} DaliaHAR--ShaSpec macro-F1 ($\uparrow$), reported mean$_{\pm\mathrm{SD}}$ across the evaluated corruption conditions. \texttt{all\_w}: \texttt{all\_wrist}; Div.: pooled diverse setting. The spreads describe variation across conditions, not confidence intervals over independent training runs. Styling is applied within each column to displayed means; rounded ties are not statistical equivalence.}
\label{tab:curriculum-tta-full}
\begingroup
\fontsize{8}{9.5}\selectfont
\setlength{\tabcolsep}{2.4pt}
\renewcommand{\arraystretch}{1.08}
\begin{tabular}{@{}lccc|ccc|ccc@{}}
\toprule
& \multicolumn{3}{c}{OGM source training} & \multicolumn{3}{c}{OPM source training} & \multicolumn{3}{c}{MIAM source training} \\
\cmidrule(lr){2-4}\cmidrule(lr){5-7}\cmidrule(lr){8-10}
Method & \texttt{w\_acc} & \texttt{all\_w} & Div. & \texttt{w\_acc} & \texttt{all\_w} & Div. & \texttt{w\_acc} & \texttt{all\_w} & Div. \\
\midrule
Static & 0.55$_{\pm0.10}$ & 0.53$_{\pm0.13}$ & 0.57$_{\pm0.08}$ & 0.58$_{\pm0.10}$ & 0.55$_{\pm0.15}$ & 0.59$_{\pm0.09}$ & 0.60$_{\pm0.08}$ & 0.55$_{\pm0.13}$ & 0.58$_{\pm0.08}$ \\
TENT & 0.60$_{\pm0.15}$ & 0.57$_{\pm0.18}$ & \underline{0.62}$_{\pm0.10}$ & \underline{0.63}$_{\pm0.16}$ & 0.57$_{\pm0.21}$ & \underline{0.63}$_{\pm0.12}$ & \textcolor{burgundy}{\textbf{0.64}}$_{\pm0.15}$ & \underline{0.57}$_{\pm0.21}$ & \underline{0.61}$_{\pm0.11}$ \\
EATA & \underline{0.61}$_{\pm0.14}$ & 0.57$_{\pm0.18}$ & \underline{0.62}$_{\pm0.10}$ & \underline{0.63}$_{\pm0.13}$ & \underline{0.58}$_{\pm0.19}$ & \underline{0.63}$_{\pm0.10}$ & \textcolor{burgundy}{\textbf{0.64}}$_{\pm0.11}$ & \underline{0.57}$_{\pm0.17}$ & \underline{0.61}$_{\pm0.09}$ \\
READ & 0.60$_{\pm0.16}$ & \underline{0.58}$_{\pm0.17}$ & \underline{0.62}$_{\pm0.10}$ & 0.60$_{\pm0.18}$ & 0.57$_{\pm0.18}$ & 0.62$_{\pm0.11}$ & \underline{0.63}$_{\pm0.12}$ & 0.55$_{\pm0.19}$ & 0.60$_{\pm0.10}$ \\
SuMi & 0.56$_{\pm0.11}$ & 0.54$_{\pm0.14}$ & 0.58$_{\pm0.08}$ & 0.58$_{\pm0.13}$ & 0.54$_{\pm0.16}$ & 0.59$_{\pm0.10}$ & 0.59$_{\pm0.11}$ & 0.53$_{\pm0.16}$ & 0.57$_{\pm0.09}$ \\
SoTTA & 0.60$_{\pm0.13}$ & 0.56$_{\pm0.15}$ & 0.61$_{\pm0.09}$ & 0.60$_{\pm0.13}$ & 0.55$_{\pm0.17}$ & 0.61$_{\pm0.10}$ & 0.60$_{\pm0.12}$ & 0.54$_{\pm0.17}$ & 0.58$_{\pm0.09}$ \\
AeO & 0.45$_{\pm0.04}$ & 0.42$_{\pm0.06}$ & 0.45$_{\pm0.04}$ & 0.43$_{\pm0.05}$ & 0.41$_{\pm0.05}$ & 0.44$_{\pm0.04}$ & 0.43$_{\pm0.04}$ & 0.41$_{\pm0.04}$ & 0.42$_{\pm0.04}$ \\
PTA & 0.54$_{\pm0.11}$ & 0.54$_{\pm0.14}$ & 0.58$_{\pm0.08}$ & 0.59$_{\pm0.13}$ & 0.55$_{\pm0.17}$ & 0.60$_{\pm0.10}$ & 0.61$_{\pm0.10}$ & 0.56$_{\pm0.17}$ & 0.59$_{\pm0.09}$ \\
PSDNorm & 0.49$_{\pm0.06}$ & 0.48$_{\pm0.07}$ & 0.53$_{\pm0.05}$ & 0.53$_{\pm0.07}$ & 0.51$_{\pm0.09}$ & 0.56$_{\pm0.06}$ & 0.56$_{\pm0.06}$ & 0.53$_{\pm0.08}$ & 0.56$_{\pm0.06}$ \\
\midrule
\rowcolor{tolblue!10}
\method{} & \textcolor{burgundy}{\textbf{0.62}}$_{\pm0.16}$ & \textcolor{burgundy}{\textbf{0.60}}$_{\pm0.18}$ & \textcolor{burgundy}{\textbf{0.63}}$_{\pm0.09}$ & \textcolor{burgundy}{\textbf{0.66}}$_{\pm0.15}$ & \textcolor{burgundy}{\textbf{0.63}}$_{\pm0.19}$ & \textcolor{burgundy}{\textbf{0.65}}$_{\pm0.10}$ & \underline{0.63}$_{\pm0.15}$ & \textcolor{burgundy}{\textbf{0.62}}$_{\pm0.16}$ & \textcolor{burgundy}{\textbf{0.63}}$_{\pm0.09}$ \\
\bottomrule
\end{tabular}
\endgroup
\end{table}

\subsection{Additional component ablations}\label{app:additional_components}

In Table~\ref{tab:ablation-components}, R8 retains detection and routing, fixes $a^j=\rho^j$ for suspect modalities $j\in\mathcal{S}_i$, and disables the optimizer. Comparison with full \method{} measures the joint benefit of normalization and attenuation updates. R9 applies \method{} unchanged to a uniform-trained ShaSpec checkpoint with a uniform prior; source weights and the prior both change. It measures their combined contribution, while Table~\ref{tab:ablation-prior} provides the complementary intervention on the routing prior itself. These ablations report point estimates.

\clearpage
\subsection{Uninformative-modality control}\label{app:white_noise_control}

On DaliaHAR--MAESTRO, we replace one, two, or three input modalities with variance-matched white noise before SAX tokenization and keep the replacement fixed across training epochs. This tests source training in the presence of persistently uninformative streams. Table~\ref{tab:white-noise-control} reports the terminal dropout profile $\rho^j$ used by \method{}. Replaced streams receive values of 0.00--0.01, while the remaining wrist/chest accelerometers retain values of 0.80. Classification F1 is 80.40--82.50 points in the replacement settings, compared with 80.20$\pm$1.40 for the three-seed control. The available means show no observed F1 reduction, but replacement-run dispersions were not reported, so this comparison does not establish statistical equivalence. It also does not test whether a prior learned before a sensor failure remains appropriate after a persistent deployment shift.

\begin{table}[htbp]
\centering
\captionsetup{font=small,skip=4pt}
\caption{\textbf{Variance-matched white-noise control on DaliaHAR--MAESTRO.} Bold profile entries identify the replaced streams, not the best values. $\rho^j$ is the terminal dropout probability, rather than a profile normalized to sum to one. F1 is displayed on a 0--100 scale. Only the unmodified control includes a reported three-seed spread.}
\label{tab:white-noise-control}
\begingroup
\fontsize{8}{9.5}\selectfont
\setlength{\tabcolsep}{3pt}
\renewcommand{\arraystretch}{1.08}
\begin{tabular}{@{}lccccc|c@{}}
\toprule
& \multicolumn{5}{c}{Terminal dropout profile $\rho^j$} & \\
\cmidrule(lr){2-6}
White-noise modalities & \texttt{w\_acc} & \texttt{w\_bvp} & \texttt{w\_eda} & \texttt{w\_temp} & \texttt{c\_acc} & F1 points \\
\midrule
None & 0.75 & 0.80 & 0.66 & 0.40 & 0.80 & 80.20$_{\pm1.40}$ \\
\texttt{w\_temp} & 0.80 & 0.80 & 0.44 & \textbf{0.01} & 0.80 & 81.40 \\
\texttt{w\_temp, w\_eda} & 0.80 & 0.80 & \textbf{0.00} & \textbf{0.00} & 0.80 & 82.50 \\
\texttt{w\_temp, w\_eda, w\_bvp} & 0.80 & \textbf{0.00} & \textbf{0.00} & \textbf{0.00} & 0.80 & 80.40 \\
\bottomrule
\end{tabular}
\endgroup
\end{table}

\subsection{Sensitivity to input-projection width}\label{app:width_control}

We separately widen one modality's input projection eightfold and retrain DaliaHAR--MAESTRO with the remaining setup fixed: chest acceleration from hidden width 96 to 768, or wrist temperature from 32 to 256. Table~\ref{tab:width-control} reports the normalized gradient-based reliance ratios $r_{e^\star}^j$, which precede the dropout transformation and should not be confused with the terminal dropout probabilities $\rho^j$ in Table~\ref{tab:white-noise-control}. The widened modality's ratio increases by 2.90 percentage points for chest acceleration and 4.80 points for wrist temperature. Mean classification F1 changes from 80.30 to 80.20 and 79.50 points, respectively. These limited retraining controls do not show reliance increasing in proportion to the eightfold width change, but several other modalities' ratios also change; they therefore do not establish parameterization invariance. \method{} uses the reliance profile with the checkpoint from which it was estimated.

\begin{table}[htbp]
\centering
\captionsetup{font=small,skip=4pt}
\caption{\textbf{Input-projection widening on DaliaHAR--MAESTRO.} Reported mean$_{\pm\mathrm{SD}}$; bold identifies the widened modality. Reliance ratios are percentages; F1 uses a 0--100 scale. These are separate retraining controls from the white-noise experiment.}
\label{tab:width-control}
\begingroup
\fontsize{8}{9.5}\selectfont
\setlength{\tabcolsep}{3pt}
\renewcommand{\arraystretch}{1.08}
\begin{tabular}{@{}lccccc|c@{}}
\toprule
& \multicolumn{5}{c}{Normalized reliance $100\,r_{e^\star}^j$ (\%)} & \\
\cmidrule(lr){2-6}
Input projection & \texttt{w\_acc} & \texttt{w\_bvp} & \texttt{w\_eda} & \texttt{w\_temp} & \texttt{c\_acc} & F1 points \\
\midrule
Original widths & 29.50$_{\pm4.40}$ & 17.80$_{\pm1.30}$ & 19.30$_{\pm3.10}$ & 9.80$_{\pm4.10}$ & 23.70$_{\pm13.00}$ & 80.30$_{\pm0.70}$ \\
\texttt{c\_acc} $\times8$ & 22.20$_{\pm2.90}$ & 17.80$_{\pm1.70}$ & 12.10$_{\pm9.00}$ & 21.30$_{\pm6.20}$ & \textbf{26.60}$_{\pm11.60}$ & 80.20$_{\pm0.70}$ \\
\texttt{w\_temp} $\times8$ & 19.30$_{\pm4.30}$ & 28.20$_{\pm7.80}$ & 14.60$_{\pm5.70}$ & \textbf{14.60}$_{\pm11.50}$ & 23.40$_{\pm3.80}$ & 79.50$_{\pm2.10}$ \\
\bottomrule
\end{tabular}
\endgroup
\end{table}

%% file: iclr2027_conference.bib
@article{lubana2020gradient,
  title={A gradient flow framework for analyzing network pruning},
  author={Lubana, Ekdeep Singh and Dick, Robert P},
  journal={arXiv preprint arXiv:2009.11839},
  year={2020}
}

@inproceedings{NEURIPS2024_b2f2af54,
 author = {Yun, Sukwon and Choi, Inyoung and Peng, Jie and Wu, Yangfan and Bao, Jingxuan and Zhang, Qiyiwen and Xin, Jiayi and Long, Qi and Chen, Tianlong},
 booktitle = {Advances in Neural Information Processing Systems},
 pages = {98782--98805},
 publisher = {Curran Associates, Inc.},
 title = {Flex-MoE: Modeling Arbitrary Modality Combination via the Flexible Mixture-of-Experts},
 volume = {37},
 year = {2024}
}

@inproceedings{schmidt2018introducing,
  title={Introducing wesad, a multimodal dataset for wearable stress and affect detection},
  author={Schmidt, Philip and Reiss, Attila and Duerichen, Robert and Marberger, Claus and Van Laerhoven, Kristof},
  booktitle={Proceedings of the 20th ACM international conference on multimodal interaction},
  pages={400--408},
  year={2018}
}

@article{altun2010comparative,
  title={Comparative study on classifying human activities with miniature inertial and magnetic sensors},
  author={Altun, Kerem and Barshan, Billur and Tun{\c{c}}el, Orkun},
  journal={Pattern Recognition},
  volume={43},
  number={10},
  pages={3605--3620},
  year={2010},
  publisher={Elsevier}
}

@inproceedings{sharma2007prevalence,
  title={On the prevalence of sensor faults in real-world deployments},
  author={Sharma, Abhishek and Golubchik, Leana and Govindan, Ramesh},
  booktitle={2007 4th Annual IEEE Communications Society Conference on Sensor, Mesh and Ad Hoc Communications and Networks},
  pages={213--222},
  year={2007},
  organization={IEEE}
}

@article{reiss2019deep,
  title={Deep PPG: Large-scale heart rate estimation with convolutional neural networks},
  author={Reiss, Attila and Indlekofer, Ina and Schmidt, Philip and Van Laerhoven, Kristof},
  journal={Sensors},
  volume={19},
  number={14},
  pages={3079},
  year={2019},
  publisher={MDPI}
}

@inproceedings{wang2023multi,
  title={Multi-modal learning with missing modality via shared-specific feature modelling},
  author={Wang, Hu and Chen, Yuanhong and Ma, Congbo and Avery, Jodie and Hull, Louise and Carneiro, Gustavo},
  booktitle={Proceedings of the IEEE/CVF Conference on Computer Vision and Pattern Recognition},
  pages={15878--15887},
  year={2023}
}

@article{mohapatra2024wearable,
  title={Wearable network for multilevel physical fatigue prediction in manufacturing workers},
  author={Mohapatra, Payal and Aravind, Vasudev and Bisram, Marisa and Lee, Young-Joong and Jeong, Hyoyoung and Jinkins, Katherine and Gardner, Richard and Streamer, Jill and Bowers, Brent and Cavuoto, Lora and others},
  journal={PNAS nexus},
  volume={3},
  number={10},
  pages={pgae421},
  year={2024},
  publisher={Oxford University Press US}
}

@inproceedings{mohapatra24_interspeech,
  title     = {{Missingness-resilient Video-enhanced Multimodal Disfluency Detection}},
  author    = {Payal Mohapatra and Shamika Likhite and Subrata Biswas and Bashima Islam and Qi Zhu},
  year      = {2024},
  booktitle = {{Interspeech 2024}},
  pages     = {5093--5097},
  doi       = {10.21437/Interspeech.2024-1458},
  issn      = {2958-1796},
}

@inproceedings{mohapatra2023person,
  title={Person identification with wearable sensing using missing feature encoding and multi-stage modality fusion},
  author={Mohapatra, Payal and Pandey, Akash and Keten, Sinan and Chen, Wei and Zhu, Qi},
  booktitle={ICASSP 2023-2023 IEEE International Conference on Acoustics, Speech and Signal Processing (ICASSP)},
  pages={1--2},
  year={2023},
  organization={IEEE}
}

@article{mohapatra2025maestro,
  title={MAESTRO: Adaptive Sparse Attention and Robust Learning for Multimodal Dynamic Time Series},
  author={Mohapatra, Payal and Sui, Yueyuan and Pandey, Akash and Xia, Stephen and Zhu, Qi},
  journal={arXiv preprint arXiv:2509.25278},
  year={2025}
}

@inproceedings{peng2022balanced,
  title={Balanced multimodal learning via on-the-fly gradient modulation},
  author={Peng, Xiaokang and Wei, Yake and Deng, Andong and Wang, Dong and Hu, Di},
  booktitle={Proceedings of the IEEE/CVF conference on computer vision and pattern recognition},
  pages={8238--8247},
  year={2022}
}

@inproceedings{wang2020makes,
  title={What makes training multi-modal classification networks hard?},
  author={Wang, Weiyao and Tran, Du and Feiszli, Matt},
  booktitle={Proceedings of the IEEE/CVF conference on computer vision and pattern recognition},
  pages={12695--12705},
  year={2020}
}

@article{xu2023fits,
  title={FITS: Modeling time series with $10 k $ parameters},
  author={Xu, Zhijian and Zeng, Ailing and Xu, Qiang},
  journal={arXiv preprint arXiv:2307.03756},
  year={2023}
}

@inproceedings{sunmultivariate,
  title={Multivariate Time Series Anomaly Detection with Idempotent Reconstruction},
  author={Sun, Xin and Zhou, Heng and Li, Chao},
  booktitle={The Thirty-ninth Annual Conference on Neural Information Processing Systems}
}

@article{xu2021anomaly,
  title={Anomaly transformer: Time series anomaly detection with association discrepancy},
  author={Xu, Jiehui and Wu, Haixu and Wang, Jianmin and Long, Mingsheng},
  journal={arXiv preprint arXiv:2110.02642},
  year={2021}
}

@inproceedings{kim2025battling,
  title={Battling the non-stationarity in time series forecasting via test-time adaptation},
  author={Kim, HyunGi and Kim, Siwon and Mok, Jisoo and Yoon, Sungroh},
  booktitle={Proceedings of the AAAI Conference on Artificial Intelligence},
  volume={39},
  number={17},
  pages={17868--17876},
  year={2025}
}

@inproceedings{grover2025shift,
  title={Shift-aware test time adaptation and benchmarking for time-series forecasting},
  author={Grover, Shivam and Etemad, Ali},
  booktitle={Second Workshop on Test-Time Adaptation: Putting Updates to the Test! at ICML 2025},
  year={2025}
}

@inproceedings{yang2024test,
  title={Test-time adaptation against multi-modal reliability bias},
  author={Yang, Mouxing and Li, Yunfan and Zhang, Changqing and Hu, Peng and Peng, Xi},
  booktitle={The twelfth international conference on learning representations},
  year={2024}
}

@inproceedings{chen2025test,
  title={Test-time selective adaptation for uni-modal distribution shift in multi-modal data},
  author={Chen, MingCai and Zhang, Baoming and Han, Zongbo and Jiang, Wenyu and Wang, Yanmeng and Feng, Shuai and Du, Yuntao and Bao, Bingkun},
  booktitle={Forty-second International Conference on Machine Learning},
  year={2025}
}

@article{guo2025smoothing,
  title={Smoothing the shift: Towards stable test-time adaptation under complex multimodal noises},
  author={Guo, Zirun and Jin, Tao},
  journal={arXiv preprint arXiv:2503.02616},
  year={2025}
}

@article{gnassounou2025psdnorm,
  title={Psdnorm: Test-time temporal normalization for deep learning in sleep staging},
  author={Gnassounou, Th{\'e}o and Collas, Antoine and Flamary, R{\'e}mi and Gramfort, Alexandre},
  journal={arXiv preprint arXiv:2503.04582},
  year={2025}
}

@article{gnassounou2024multi,
  title={Multi-source and test-time domain adaptation on multivariate signals using spatio-temporal monge alignment},
  author={Gnassounou, Th{\'e}o and Collas, Antoine and Flamary, R{\'e}mi and Lounici, Karim and Gramfort, Alexandre},
  journal={arXiv preprint arXiv:2407.14303},
  year={2024}
}

@article{gong2023sotta,
  title={Sotta: Robust test-time adaptation on noisy data streams},
  author={Gong, Taesik and Kim, Yewon and Lee, Taeckyung and Chottananurak, Sorn and Lee, Sung-Ju},
  journal={Advances in Neural Information Processing Systems},
  volume={36},
  pages={14070--14093},
  year={2023}
}

@inproceedings{wangpartition,
  title={Partition-Then-Adapt: Combating Prediction Bias for Reliable Multi-Modal Test-Time Adaptation},
  author={Wang, Guowei and Lyu, Fan and Ding, Changxing},
  booktitle={The Thirty-ninth Annual Conference on Neural Information Processing Systems}
}

@article{dong2025towards,
  title={Towards robust multimodal open-set test-time adaptation via adaptive entropy-aware optimization},
  author={Dong, Hao and Chatzi, Eleni and Fink, Olga},
  journal={arXiv preprint arXiv:2501.13924},
  year={2025}
}

@article{wang2020tent,
  title={Tent: Fully test-time adaptation by entropy minimization},
  author={Wang, Dequan and Shelhamer, Evan and Liu, Shaoteng and Olshausen, Bruno and Darrell, Trevor},
  journal={arXiv preprint arXiv:2006.10726},
  year={2020}
}

@inproceedings{niu2022efficient,
  title={Efficient test-time model adaptation without forgetting},
  author={Niu, Shuaicheng and Wu, Jiaxiang and Zhang, Yifan and Chen, Yaofo and Zheng, Shijian and Zhao, Peilin and Tan, Mingkui},
  booktitle={International conference on machine learning},
  pages={16888--16905},
  year={2022},
  organization={PMLR}
}

@misc{defrana_physical_activity,
  author       = {de Frana, Diego Silva},
  title        = {Physical Activity Prediction Dataset},
  year         = {n.d.},
  publisher    = {Kaggle},
  howpublished = {\url{https://www.kaggle.com/datasets/diegosilvadefrana/fisical-activity-dataset}},
  note         = {Accessed: 2026-05-05}
}

@book{little2019statistical,
  title={Statistical analysis with missing data},
  author={Little, Roderick JA and Rubin, Donald B},
  year={2019},
  publisher={John Wiley \& Sons}
}

@article{allen2007photoplethysmography,
  title={Photoplethysmography and its application in clinical physiological measurement},
  author={Allen, John},
  journal={Physiological measurement},
  volume={28},
  number={3},
  pages={R1--R39},
  year={2007}
}

@article{braei2020anomaly,
  title={Anomaly detection in univariate time-series: A survey on the state-of-the-art},
  author={Braei, Mohammad and Wagner, Sebastian},
  journal={arXiv preprint arXiv:2004.00433},
  year={2020}
}

@inproceedings{bashar2020tanogan,
  title={TAnoGAN: Time series anomaly detection with generative adversarial networks},
  author={Bashar, Md Abul and Nayak, Richi},
  booktitle={2020 IEEE Symposium Series on Computational Intelligence (SSCI)},
  pages={1778--1785},
  year={2020},
  organization={IEEE}
}

@book{oppenheim1999discrete,
  title={Discrete-time signal processing},
  author={Oppenheim, Alan V},
  year={1999},
  publisher={Pearson Education India}
}

@article{orphanidou2017signal,
  title={Signal quality assessment in physiological monitoring: state of the art and practical considerations},
  author={Orphanidou, Christina},
  year={2017},
  publisher={Springer}
}

@inproceedings{reiss2012pamap2,
author = {Reiss, Attila and Stricker, Didier},
title = {Creating and benchmarking a new dataset for physical activity monitoring},
year = {2012},
isbn = {9781450313001},
publisher = {Association for Computing Machinery},
address = {New York, NY, USA},
url = {https://doi.org/10.1145/2413097.2413148},
doi = {10.1145/2413097.2413148},
booktitle = {Proceedings of the 5th International Conference on PErvasive Technologies Related to Assistive Environments},
articleno = {40},
numpages = {8},
location = {Heraklion, Crete, Greece},
series = {PETRA '12}
}

@article{zhang2014troika,
  title={TROIKA: A general framework for heart rate monitoring using wrist-type photoplethysmographic signals during intensive physical exercise},
  author={Zhang, Zhilin and Pi, Zhouyue and Liu, Benyuan},
  journal={IEEE Transactions on biomedical engineering},
  volume={62},
  number={2},
  pages={522--531},
  year={2014},
  publisher={IEEE}
}

@article{wei2024fly,
  title={On-the-fly modulation for balanced multimodal learning},
  author={Wei, Yake and Hu, Di and Du, Henghui and Wen, Ji-Rong},
  journal={IEEE Transactions on Pattern Analysis and Machine Intelligence},
  volume={47},
  number={1},
  pages={469--485},
  year={2024},
  publisher={IEEE}
}

@inproceedings{zbinden2026miam,
  title={MIAM: Modality Imbalance-Aware Masking for Multimodal Ecological Applications},
  author={Zbinden, Robin and Monteith-Finas, Wesley and Sumbul, Gencer and van Tiel, Nina and Vanalli, Chiara and Tuia, Devis},
  booktitle={The Fourteenth International Conference on Learning Representations},
  year={2026}
}

@article{he2026decoupling,
  title={Decoupling stability and plasticity for multi-modal test-time adaptation},
  author={He, Yongbo and Guo, Zirun and Jin, Tao},
  journal={arXiv preprint arXiv:2603.00574},
  year={2026}
}
